\documentclass[runningheads]{llncs}
\usepackage{graphicx}
\usepackage{amsmath,amssymb} 
\usepackage{color}
\usepackage{rotating}
\newcommand{\myparagraph}[1]{\vspace{0.25em}\noindent\emph{#1}\hspace{0.3cm}}
\usepackage[width=122mm,left=12mm,paperwidth=146mm,height=193mm,top=12mm,paperheight=217mm]{geometry}
\begin{document}
\pagestyle{headings}
\mainmatter
\def\ECCV16SubNumber{1165}  

\title{A Multi-cut Formulation for Joint Segmentation and Tracking of Multiple Objects} 

\titlerunning{Joint Segmentation and Tracking}
\authorrunning{Keuper et al.}

\author{Margret Keuper$^1$, Siyu Tang$^2$, Yu Zhongjie$^2$, Bjoern Andres$^2$\\Thomas Brox$^1$, Bernt Schiele$^2$}
\institute{$^1$ Department for Computer Science, University of Freiburg, Germany\\
$^2$Max Planck Institute for Informatics, Saarbruecken, Germany}

\maketitle

\begin{abstract}
Recently, Minimum Cost Multicut Formulations have been proposed and proven to be successful in both motion trajectory segmentation and multi-target tracking scenarios. Both tasks benefit from decomposing a graphical model into an optimal number of connected components based on attractive and repulsive pairwise terms. 
The two tasks are formulated on different levels of granularity and, accordingly, leverage mostly local information for motion segmentation and mostly high-level information for multi-target tracking.
In this paper we argue that point trajectories and their local relationships can contribute to the high-level task of multi-target tracking and also argue that high-level cues from object detection and tracking are helpful to solve motion segmentation. 
We propose a joint graphical model for point trajectories and object detections whose Multicuts are solutions to motion segmentation {\it and} multi-target tracking problems at once. Results on the FBMS59 motion segmentation benchmark as well as on pedestrian tracking sequences from the 2D MOT 2015 benchmark demonstrate the promise of this joint approach.
 
\end{abstract}

\section{Introduction}
Several problems in computer vision, such as image segmentation or motion segmentation in video, are traditionally approached in a low-level, bottom-up way while other tasks like object detection, multi-target tracking, and action recognition often require previously learned model information and are therefore traditionally approached from a high-level perspective.

In this paper, we propose a joint formulation for one such classical high-level problem (multi-target tracking) and a low-level problem (moving object segmentation).

Multi-target tracking and motion segmentation are both active fields in computer vision \cite{Segal_2013_ICCV,Huang:2008:ROT,WojekECCV10,AndrilukaCVPR2010,FragkiadakiECCV12,Zamir:2012:GMC,WojekPAMI2013,Henschel:2014:EMP,tang14ijcv,shiICCV2013,Ochs14,Dragon2014,Ji2014,keuper15a}.
These two problems are clearly related in the sense that their goal is to determine those regions that belong to the same moving object in an image sequence.

We argue that these interrelated problems can and should be addressed jointly so as to leverage the advantages of both.
In particular, the low-level information contained in point trajectories and in their relation to one another form important cues for the high-level task of multi-target tracking. They carry the information where single, well localized points are moving and can thus help to disambiguate partial occlusions and motion speed changes, both of which are key challenges for multi-target tracking. For motion segmentation, challenges are presented by (1) articulated motion, where purely local cues lead to over-segmentation and (2) coherently moving objects where motion cues cannot tell the objects apart. High level information from an object detector or even an object tracking system is beneficial as it provides information about the rough object location, extent, and possibly re-identification after occlusion.

Ideally, employing such pairwise information between detections may replace higher-order terms on trajectories as proposed in \cite{Ochs12}. While it is impossible to tell two rotational or scaling motions apart from only pairs of trajectories, pairs of detection bounding boxes contain enough points to distinguish their motion. With sufficiently complex detection models, even articulated motion can be disambiguated.

To leverage high-level spatial information as well as low-level motion cues in both scenarios, we propose a unified graphical model in which multi-target tracking and motion segmentation are both cast in one graph partitioning problem.
As a result, the method provides consistent identity labels in conjunction with accurate segmentations of moving objects.

We show that this joint graphical model improves over the individual, task specific models. Our results improve over the state of the art in motion segmentation evaluated on the FBMS59 \cite{Ochs14} motion segmentation benchmark and are competitive on standard multi-target pedestrian tracking sequences \cite{Andriluka:2008:PTD,Zamir:2012:GMC} while additionally providing fine-grained motion segmentations.
\section{Related Work}
Combining high-level cues with low-level cues is an established idea in computer vision
and has been used successfully e.g.\ for image segmentation \cite{DBLP:journals/corr/BertasiusST15}. 
Similarly, motion trajectories have been used for tracking \cite{Fragkiadaki11,FragkiadakiECCV12} 
and object detections for segmenting moving objects \cite{Fragkiadaki_2015_CVPR}. 
However, our proposed method is substantially different in that we provide a unified graph structure whose partitioning both solves the low level problem, here, the motion segmentation task, and the high-level problem, i.e.\ the multi target tracking task, at the  same time. In that spirit, the most related previous work is \cite{FragkiadakiECCV12}, where detectlets, small tracks of detections, are classified in a graphical model that, at the same time, performs trajectory clustering. While we draw from the motivation provided in \cite{FragkiadakiECCV12}, the key difference to our approach is that we cast both, motion segmentation and multi-target tracking, as clustering problems, allowing for the direct optimization of the Minimum Cost Multicuts \cite{chopra-1993,deza-1997}. Thus, we perform bottom-up segmentation and tracking in a single step.

In \cite{milan15}, tracking and video segmentation are also approched as one problem. However, their appoach employs CRFs instead of Minimum Cost Multicuts, is built upon temporal superpixels \cite{tsp} instead of point trajectories and strongly relies on unary terms on these superpixels learned using support vector machines.

In computer vision, Minimum Cost Multicut Formulations have been mainly applied to image segmentation \cite{andres-2011,andres-2012-globally,kappes-2011,keupericcv}. Exceptiontions are \cite{keuper15a} applying this model to motion segmentation and \cite{tang15} applying it to pedestrian tracking. In \cite{tang15}, Minimum Cost Multicuts have shown to provide a suitable alternative to network flow approaches \cite{networkflow1,networkflow2}. The clustering nature of minimum cost multicuts avoids the explicit non-maximum suppression step which is a crucial ingredient in disjoint path formulations such as \cite{networkflow1,networkflow2}.

Different approaches towards solving this combinatorial problem of linking the right detection proposal over time use integer linear programming \cite{Shitrit:2011:TMP,wang-et-al-2014}, MAP estimation \cite{Pirsiavash:2011:GOG}, or continuous optimization \cite{Andriyenko2012CVPR}. In these approaches, the complexity usually needs to be reduced by either applying non-maximum suppression  or pre-grouping detections into tracklets \cite{Huang:2008:ROT,WojekECCV10,AndrilukaCVPR2010,FragkiadakiECCV12,Zamir:2012:GMC,WojekPAMI2013,Henschel:2014:EMP,tang14ijcv}.

As \cite{Bro10c,lezama11,Ochs12,Li2013_12,shiICCV2013,Ochs14,Dragon2014,Ji2014,keuper15a}, we cast motion segmentation as a problem of grouping dense point trajectories. Most of these related approaches employ the spectral clustering paradigm to generate segmentations, while recently \cite{keuper15a} have shown the advantages of casting the motion trajectory segmentation as a minimum cost multicut problem.\\
The approaches of \cite{Cheridata2009,Dragon2012,Bro10c,Ochs14,Li2013_12} base their segmentations on pairwise affinities while \cite{Ochs12,zografos14,Elhamifar2009_6} model higher order motions by different means. In our approach, we do not make use of any higher order motion models. In fact, much of the information these terms carry is already contained in the detections we are using, such that we can leverage this information with pairwise terms. 

\section{Joint Multicut Problem Formulation}
\label{sec:model}
Here, we describe the proposed joint high-level - low-level Minimum Cost Multicut Problem formulation which we want to jointly apply to  
multi-target tracking and moving object segmentation. Our aim is to build a graphical model representing detection and point trajectory nodes and their relationships between one another in a simple, unified way such that the Multicut Problem on this graph directly yields a joint clustering of these high-level and low-level nodes into an optimal number of motion segments and according object tracks. 

\begin{figure} [t]
\setlength{\abovecaptionskip}{0pt}
  \centering
      \includegraphics[width=0.95\linewidth]{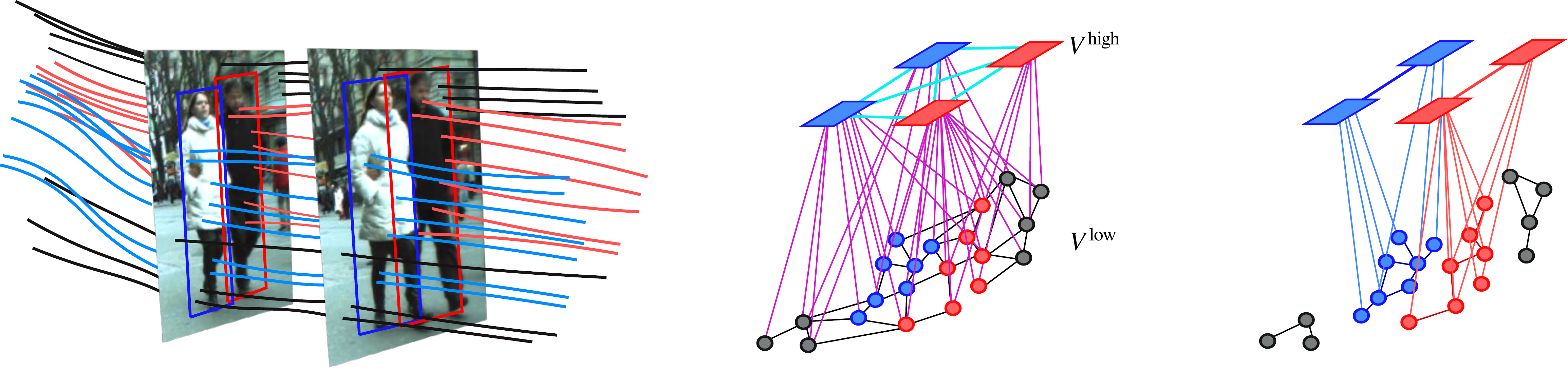}\\
\begin{minipage}{0.99\linewidth}
\begin{tabular}{ @{}c @{\hspace{0.5cm}}c @{\hspace{0.95cm}}c @{}}
(a) high-level and low-level entities&(b) proposed graph& (c) feasible Multicut 
\end{tabular}\vspace{0.1cm} 
\end{minipage}
  \caption{\label{fig:jointGraph}(a) While pedestrian detections, here drawn as bounding boxes, represent frame-wise high-level information, point trajectories computed on the same sequence represent spatio-temporal low-level cues. Both can be represented as vertices in a joint graphical model (b). The optimal decomposition of this graph into connected components yields both a motion trajectory segmentation of the sequence as well as the tracking of moving objects represented by the detections (c).}
\end{figure}

We define an undirected graph $G = (V,E)$, where $V = \{V^{\text{high}}, V^{\text{low}}\}$ is composed of nodes $v^{\text{high}}\in V^{\text{high}}$ representing high-level entities (detections), and nodes $v^{\text{low}}\in V^{\text{low}}$ representing fine-grained, low-level entities (point trajectories) as depicted in Fig.~\ref{fig:jointGraph}~(b). 

To represent the three different types of pairwise relations between these nodes, we define three different kinds of edges. The edge set $E =\{E^{\text{high}}, E^{\text{low}}, E^{\text{hl}}\} $ consists of edges $e^{\text{high}}\in E^{\text{high}}$ defining the pairwise relations between detections (depicted in cyan in Fig.~\ref{fig:jointGraph}~(b)). These can provide pairwise information computed from strong, very specific object features, reflected in the real-valued edge costs $c_{e^{\text{high}}}$. The edges $e^{\text{low}}\in E^{\text{low}}$ represent pairwise relations between point trajectories (depicted in black in  Fig.~\ref{fig:jointGraph}~(b)). The according costs $c_{e^{\text{low}}}$ are mostly based on local information. The edges $e^{\text{hl}}\in E^{\text{hl}}$ represent the pairwise relations between these two levels of granularity (depicted in magenta in Fig.~\ref{fig:jointGraph}~(b)). 
The Minimum Cost Multicut Problem on this graph defines a binary \emph{edge} labeling problem:

\begin{align}
\label{eq:JiontMCdef}
\min_{y \in \{0,1\}^E}&\sum_{e^{\text{high}} \in E^{\text{high}}} c_{e^{\text{high}}} y_{e^{\text{high}}} +\sum_{e^{\text{low}} \in E^{\text{low}}} c_{e^{\text{low}}} y_{e^{\text{low}}} + \sum_{e^{\text{hl}} \in E^{\text{hl}}} c_{e^{\text{hl}}} y_{e^{\text{hl}}}\\
\textnormal{subject to}&
	                              \quad y \in \text{MC},&\nonumber	
\end{align}
where $\text{MC}$ is the set of exactly all edge labelings $y \in\{0,1\}^E$ that decompose the graph into connected components. Thus, the feasible solutions to the optimization problem from Eq.~\ref{eq:JiontMCdef} are exactly all {\it partitionings} of the graph $G$. In the optimal case, each partition describes either the entire backgroud or exactly one object throughout the whole video at two levels of granularity:
the tracked bounding boxes of this object and the point trajectories of all points on the object. In Fig.~\ref{fig:jointGraph}~(c), the proposed solution to the Multicut problem on the graph in  Fig.~\ref{fig:jointGraph}~(b) contains four clusters: one for each pedestrian tracked over time, and two background clusters in which no detections are contained.

Formally, the feasible set of all multicuts of $G$ can be defined by the cycle inequalities \cite{chopra-1993}
$\displaystyle\textcolor{black}{\forall C\in\text{cycles}(G), \forall e \in C:} \quad y_e \leq \hspace{-2ex} \sum_{e' \in C \setminus \{e\}} \hspace{-2ex} y_{e'}$, 
making the optimization problem APX-hard \cite{demaine-2006}. Yet, the benefit of this formulation is that (1) it contains exactly the right set of feasible solutions, and (2) if $p_e$ denotes the probability of an edge $e\in E$ to be cut, then an optimal solution of the Minimum Cost Multicut Problem with the edge weights computed as $c_e=\text{logit}(p_e) = \log\frac{p_e}{1-p_e}$ is a maximally likely decomposition of $G$. Note that the \emph{logit} function generates real valued costs $c_e$ such that trivial solutions are avoided.

\subsection{Pairwise Potentials}
In this section, we describe the computation of the pairwise potentials $c_e$ we use in our model. Ideally, one would like to learn terms from training data. However, since the available training datasets for motion segmentation as well as for multi-target tracking are quite small , we choose to rather define intuitive pairwise terms whose parameters have been validated on training data.   

\subsubsection{Low-level Nodes and Edges}
\label{sec:trajectorypairwise}
In our problem setup, low-level information for motion segmentation and multi-target tracking is built upon point trajectory nodes $v^{\text{low}}$ over time and their respective pairwise relations are represented by edge costs $c_{e^{\text{low}}}$. 

\myparagraph{Low-level Nodes $v^{\text{low}}$: Motion Trajectory Computation} 
A motion trajectory is a spatio-temporal curve that describes the long-term motion of a single tracked point.
We compute the motion trajectories according to the method proposed in \cite{Ochs14}. For a given point sampling rate, all points in the first video frame having some underlying image structure are tracked based on large displacement optical flow \cite{ldof} until they are occluded or lost.

The decision about ending a trajectory is made by considering the consistency between forward and backward optical flow. In case of large inconsistencies, a point is assumed to be occluded in one of the two frames. Whenever trajectories end, new trajectories are inserted to maintain the desired sampling rate.

\myparagraph{Trajectory Edge Potentials $c_e^{\text{low}}$}
The edge potentials $c_{e^{\text{low}}}$ between point trajectories $v_i^{\text{low}}$ and $v_j^{\text{low}}$ are all computed from low-level image and motion information. Motion distances $d^{\text{m}}(v_i^{\text{low}},v_j^{\text{low}})$ are computed from the maximum motion difference between two trajectories during their common life-time as in \cite{Ochs14}.
Additionally, we compute color and spatial distances $d^{\text{c}}(v_i^{\text{low}},v_j^{\text{low}})$ and $d^{\text{sp}}(v_i^{\text{low}},v_j^{\text{low}})$ between each pair of trajectories with a common life-time and spatial distances for trajectories without temporal overlap as in \cite{keuper15a} and combine them non-linearly to $z:=c_e^{\text{low}}=\text{max}(\bar{\theta}_0+\theta_1d^{\text{m}}+\theta_2d^{\text{c}}+\theta_3d^{\text{sp}}, \theta_0+\theta_1d^{\text{m}})$. The model parameters $\theta$ are set as in \cite{keuper15a}. 
These costs can be mapped to cut probabilities $p_e$ by the logistic function
\begin{equation}
p_e = \frac{1}{1 + \exp(-z)} .
\label{eq:logFunc}
\end{equation}

\subsubsection{High-level Nodes and Edges}
The high-level nodes $v^{\text{high}}$ we consider represent object detections. Since these build upon strong underlying object models, the choice of the object detector is task dependent. While our experiments on the pedestrian tracking sequences make use the Deformable Part Model (DPM) person detector \cite{Felzenszwalb2010PAMI}, our experiments on the FBMS59 dataset \cite{Ochs14} employ a generic object detector (LSDA) \cite{Hoffman14Lsda} which is trained for a wide range of object classes as well as the more specific faster R-CNN \cite{renNIPS15fasterrcnn}. Details on the specific detectors and resulting vertex sets $V^{\text{high}}$ are given in the experimental section (Sec. \ref{sec:experimetns}).

\myparagraph{Detection Edge Potentials $c_e^{\text{high}}$}
Depending on the employed object detector and the specific task, a variety of different object features could potentially be used to compute high-level pairwise potentials. In our setup, we compute the high-level pairwise terms on simple features based on the intersection over union (IoU) of bounding boxes. On the pedestrian tracking sequences, the high-level part of our graph is built as in \cite{tang15}. More details for edges in $E^{\text{high}}$ will be specified in the experimental section (Sec. \ref{sec:experimetns}).


\subsubsection{Pairwise Potentials $c_e^{\text{hl}}$ between High-level and Low-level Nodes}
\label{sec:hlpairwise}
We assume, the safest information we can draw from any kind of object detection represented by a node $v^{\text{high}}_i$ is its spatio-temporal center position $\text{pos}_{v^{\text{high}}_i}=(x_{v^{\text{high}}_i},y_{v^{\text{high}}_i},t_{v^{\text{high}}_i})^\top$ and size $(w_{v^{\text{high}}_i},h_{v^{\text{high}}_i})^\top$. Ideally, the underlying object model allows to produce a tentative frame-wise object segmentation or template $T_{v^{\text{high}}_i}$ of the detected object. Such a segmentation template can provide far more information than the bounding box alone. Potentially, a template indicates uncertainties and enables to find regions within each bounding box, where points most likely belong to the detected object. For point trajectory nodes $v^{\text{low}}_j$, the spatio-temporal location $(x_{v^{\text{low}}_j}(t),y_{v^{\text{low}}_j}(t))^\top$ is the most reliable property.

Thus, it makes sense to compute pairwise relations between detections and trajectories according to their spatio-temporal relationship, computed from the normalized spatial distance 
\begin{align}
d^{\text{sp}}(v_i^{\text{high}},v_j^{\text{low}}) = 2\left\|
\begin{pmatrix}\frac{x_{v^{\text{high}}_i} - x_{v^{\text{low}}_j}(t)}{w_{v^{\text{high}}_i}}\\
\frac{y_{v^{\text{high}}_i} - y_{v^{\text{low}}_j}(t)}{h_{v^{\text{high}}_i}}
\end{pmatrix}\right\| \quad \text{for}\quad t=t_{v^{\text{high}}_i}
\label{eq:disthl} 
\end{align}
and the template value at the trajectory position $T_{v^{\text{high}}_i}(x_{v^{\text{low}}_j}(t),y_{v^{\text{low}}_j}(t))$. If a trajectory passes through a detected object in a frame $t$, it probably belongs to that object. If it passes far outside the objects bounding box in a certain frame, it is probably not part of this object. 

Thus, we compute edge cut probabilities $p_{e^{\text{hl}}}$ from the above described measures as
\begin{align}
p_{e^{\text{hl}}_{ij}}=
    \begin{cases}
      1-T_{v^{\text{high}}_i}(x_{v^{\text{low}}_j}(t),y_{v^{\text{low}}_j}(t)), & \text{if}\ T_{v^{\text{high}}_i}(x_{v^{\text{low}}_j}(t),y_{v^{\text{low}}_j}(t))>0.5 \\
      1, & \text{if}\ d^{\text{sp}}(v_i^{\text{high}},v_j^{\text{low}})>\sigma \\
      0.5, & \text{otherwise}
    \end{cases}
\label{eq:p_hl} 
\end{align}
using an application dependent threshold $\sigma$.

\subsection{Solving Minimum Cost Multicut Problems}
The Minimum Cost Multicut problem defined by the integer linear program in Eq.~\eqref{eq:JiontMCdef} is APX-hard \cite{demaine-2006}.
Still, instances of sizes relevant for computer vision can potentially be solved to optimallity or within tight bounds using branch
and cut \cite{andres-2012-globally}.
However, finding the optimal solution is not necessary for many real world  applications.
Recently, the primal heuristic proposed by Kernighan and Lin \cite{KL} has shown to
provide very reasonable results on image and motion segmentation tasks \cite{keupericcv,keuper15a}.
Alternative heuristics were in \cite{CGC,fusionMoves}. In our experiments, we employ \cite{KL} because of its computation speed and robust behavior.

\section{Experiments}
\label{sec:experimetns}
We evaluate the proposed Joint Multicut Formulation on both motion segmentation and multi-target tracking applications. First, we show our results on the FBMS59\cite{Ochs14} motion segmentation dataset containing sequences with various object categories and motion patterns. Then, tracking \emph{and} motion segmentation performance will be evaluated on three standard multi-target pedestrian tracking sequences. Last, we evaluate the tracking performance on the 2D MOT 2015 benchmark \cite{MOT15}. 

\subsection{Motion Segmentation Dataset}
The FBMS59\cite{Ochs14} motion segmentation dataset consists of 59 sequences split into a training set of 29 and a test set of 30 sequences. The videos are of varying length (19 to about 500 frames) and show diverse types of moving objects such as cars, persons and different types of animals.

To exploit the Joint Multicut model for this data, the very first question is how to obtain reliable detections in a video sequence without knowing the category of the object of interest. Here, we evaluate on detections from two different methods : Large Scale Detection through Adaptation (LSDA) \cite{Hoffman14Lsda} and the Faster R-CNN \cite{renNIPS15fasterrcnn}.
\begin{figure} [tb]
\centering
\rotatebox{0}{LSDA + selective search}
\begin{tabular}{@{}c@{}c@{}c@{}c@{}}
\includegraphics[height=15mm]{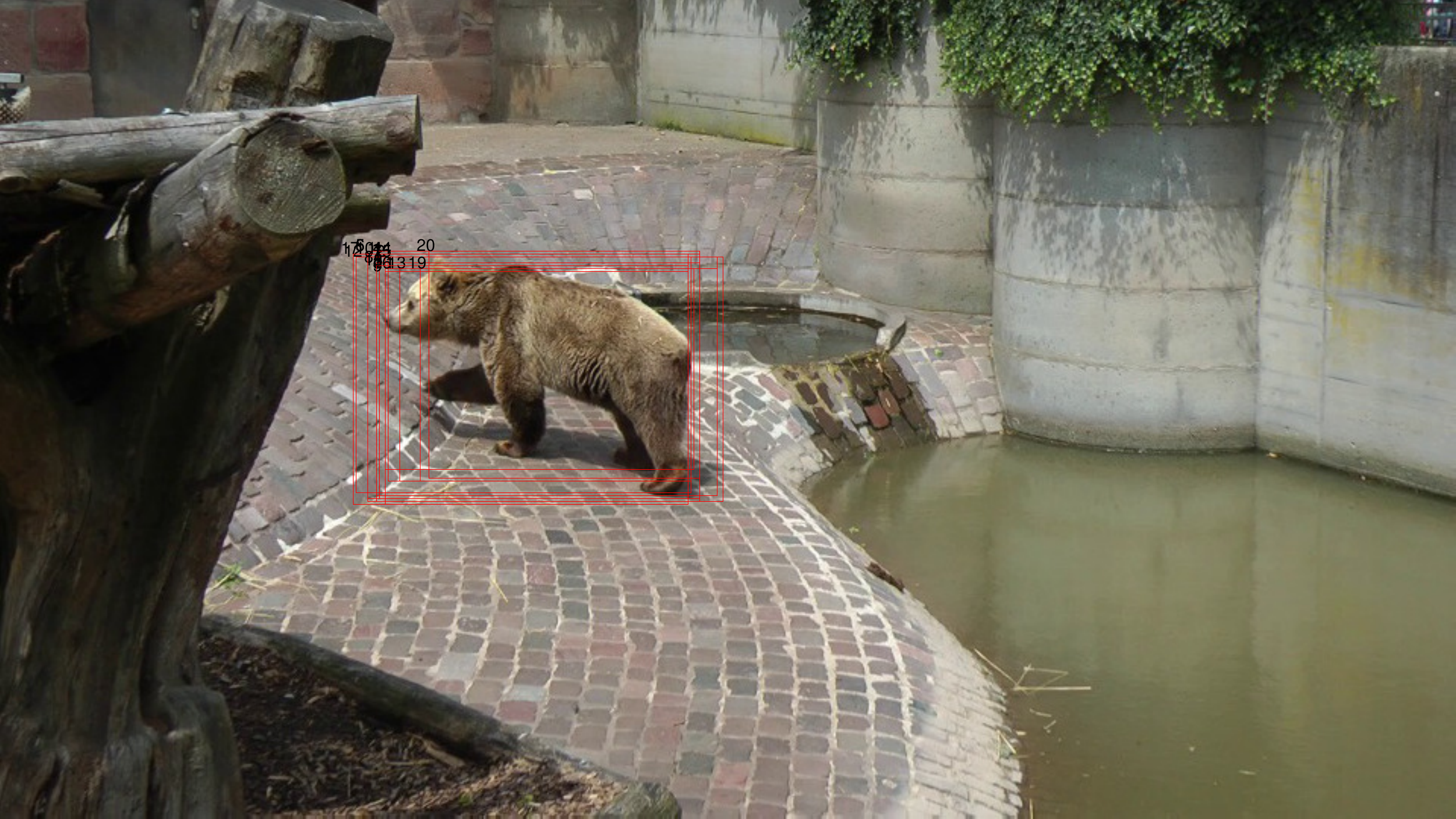}&
\includegraphics[height=15mm]{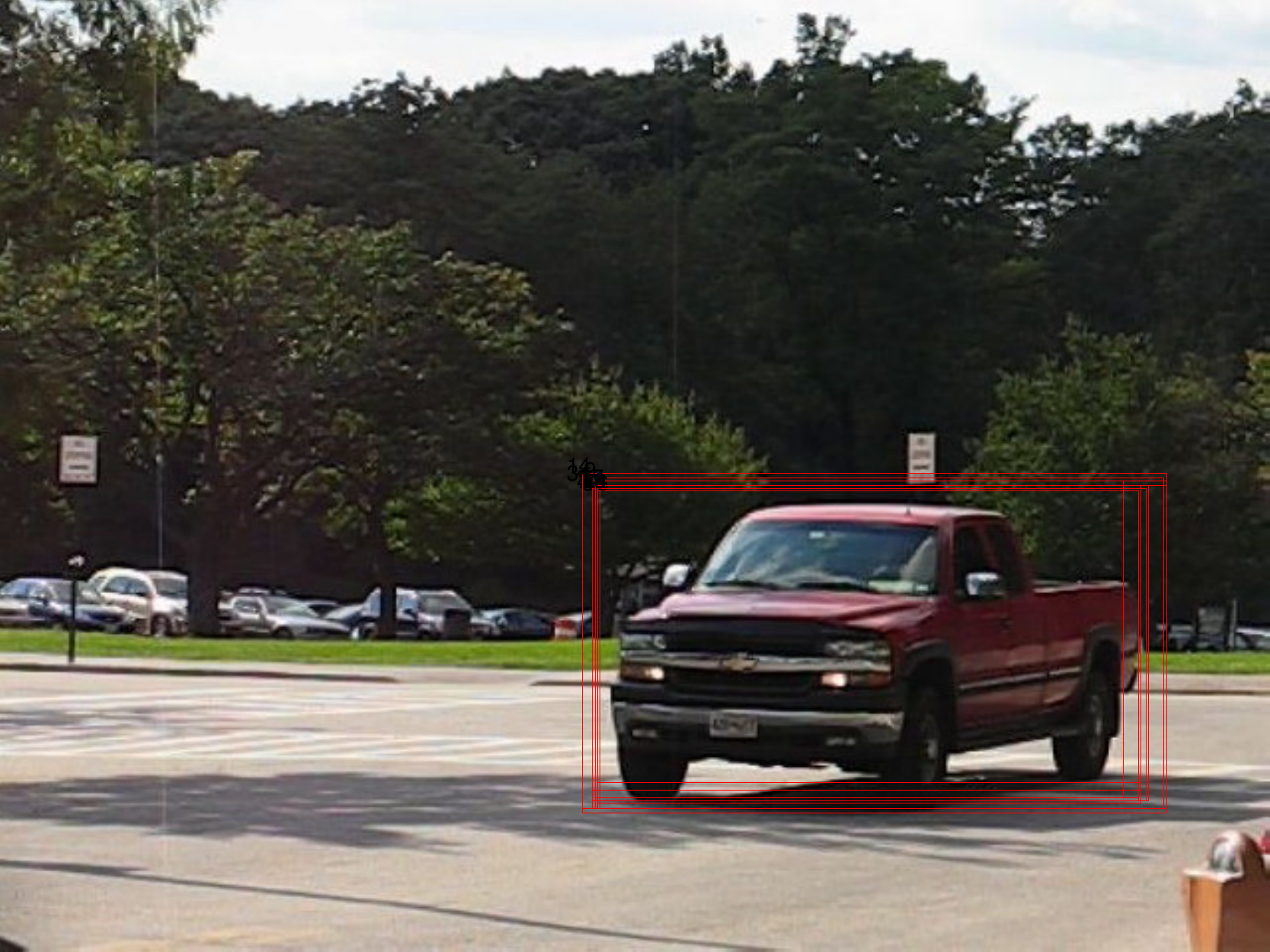}&
\includegraphics[height=15mm]{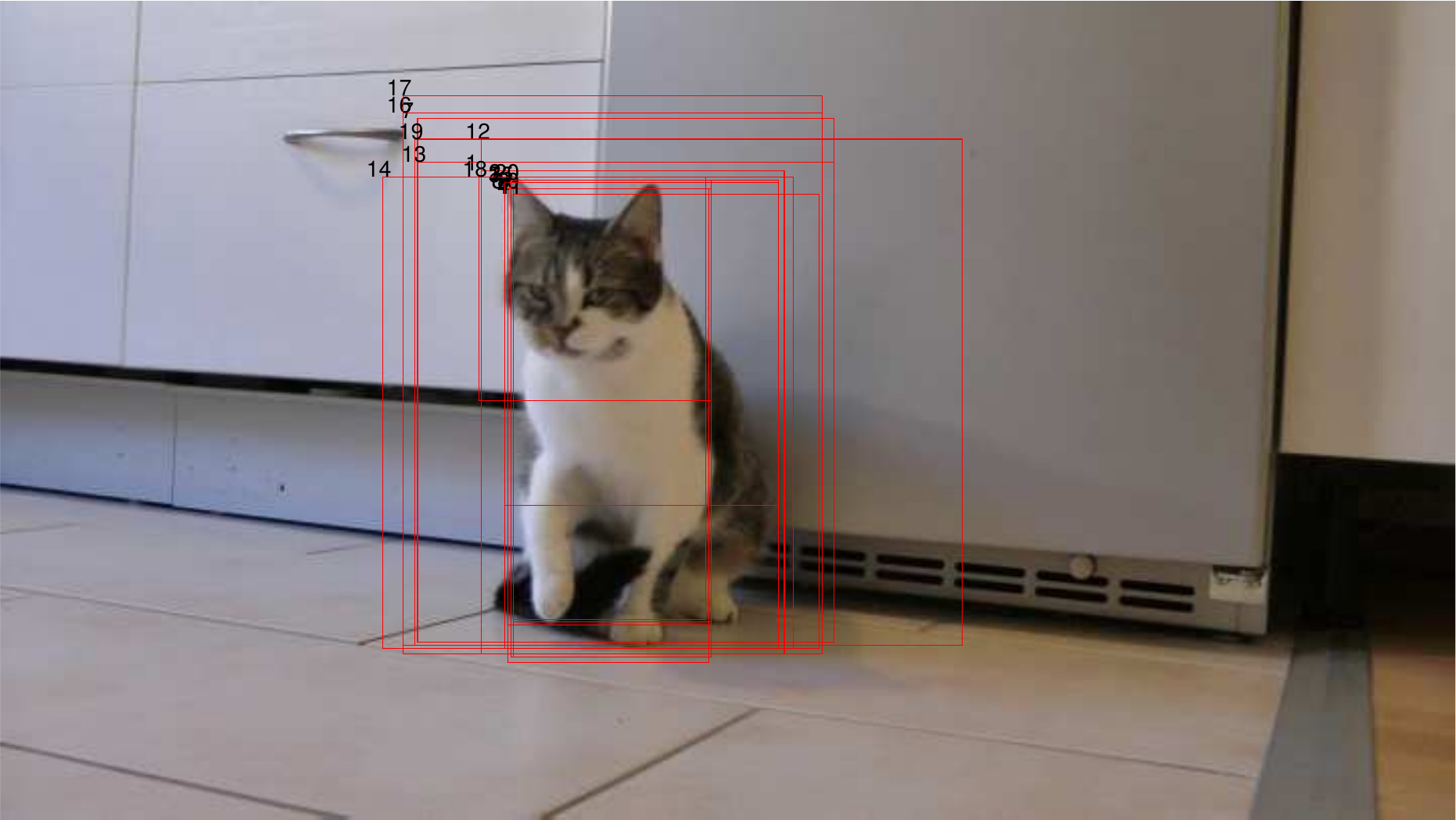}&
\includegraphics[height=15mm]{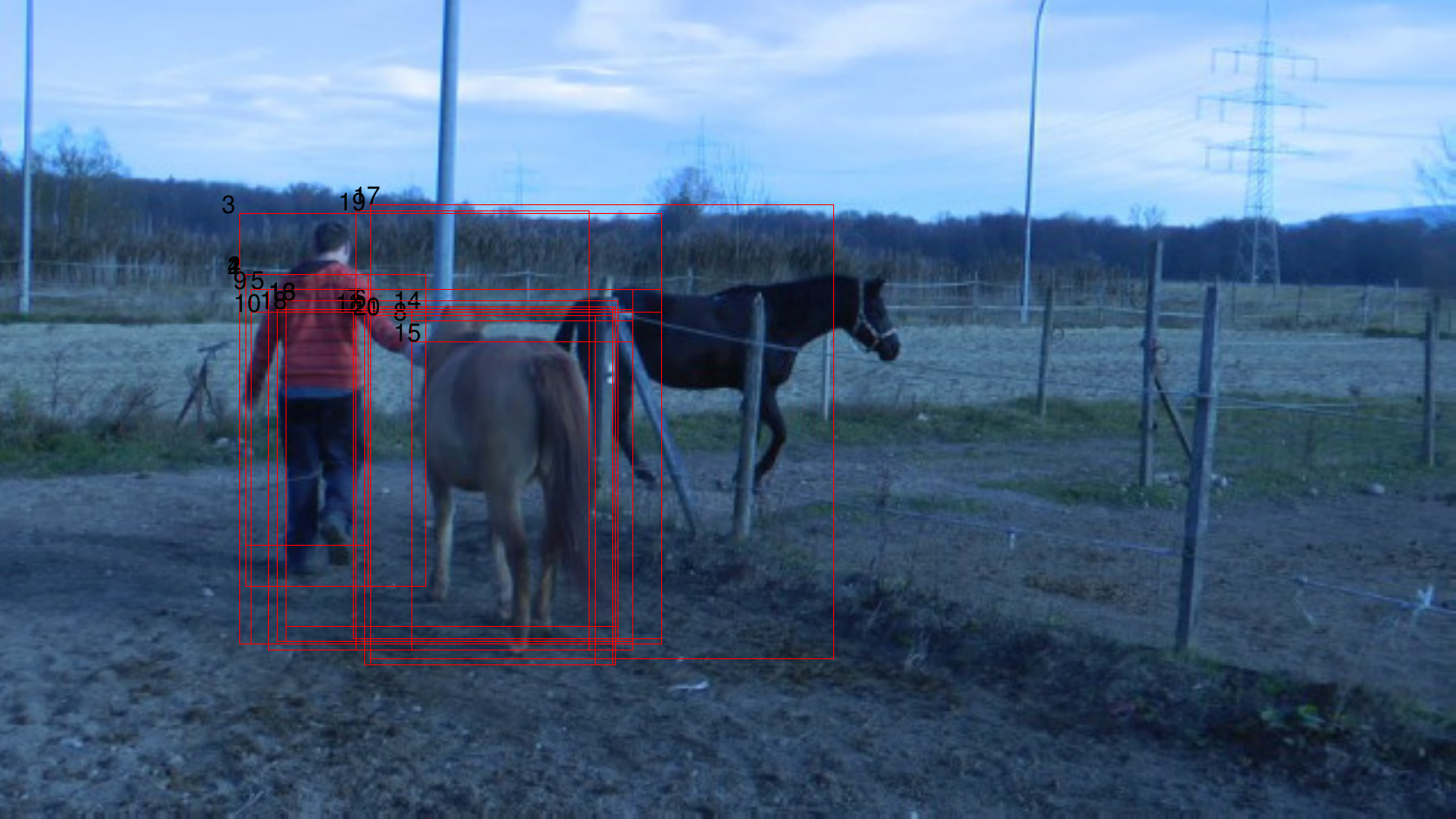}\\
\end{tabular}
\begin{tabular}{@{}c@{}c@{}c@{}c@{}c@{}c@{}c@{}c@{}c@{}c@{}c@{}c@{}c@{}c@{}c@{}}
\includegraphics[height=5mm]{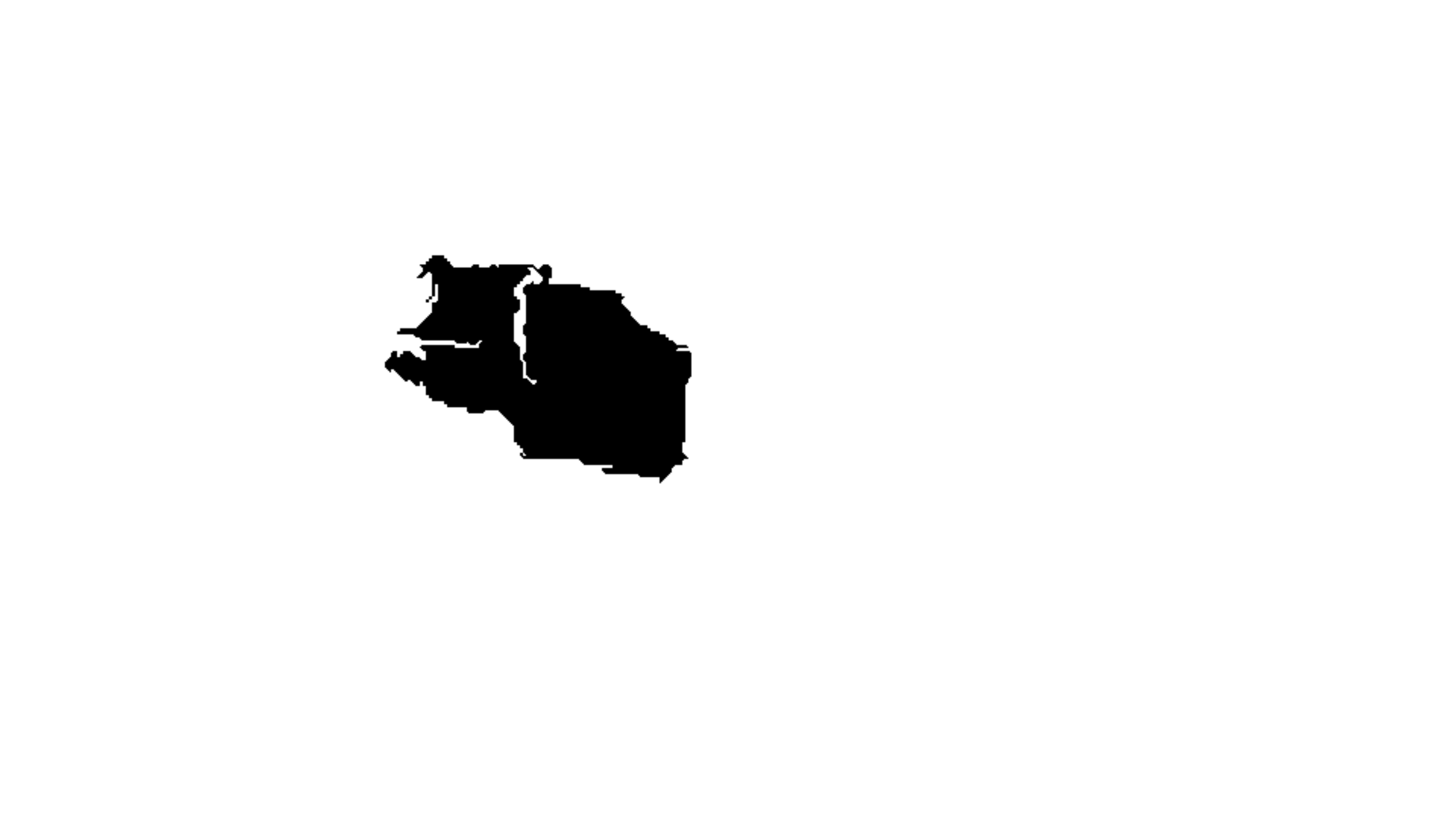}&
\includegraphics[height=5mm]{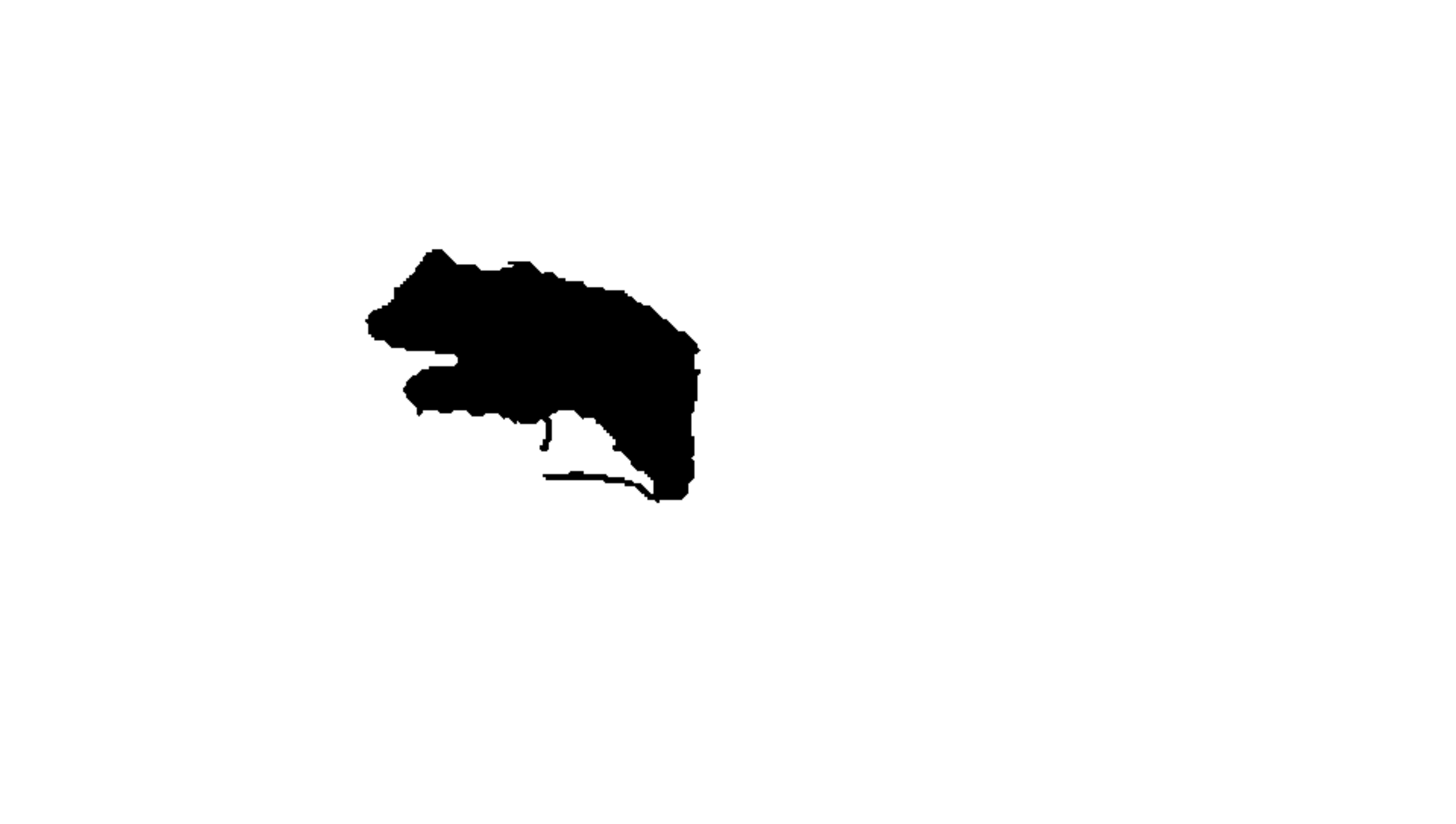}&
\includegraphics[height=5mm]{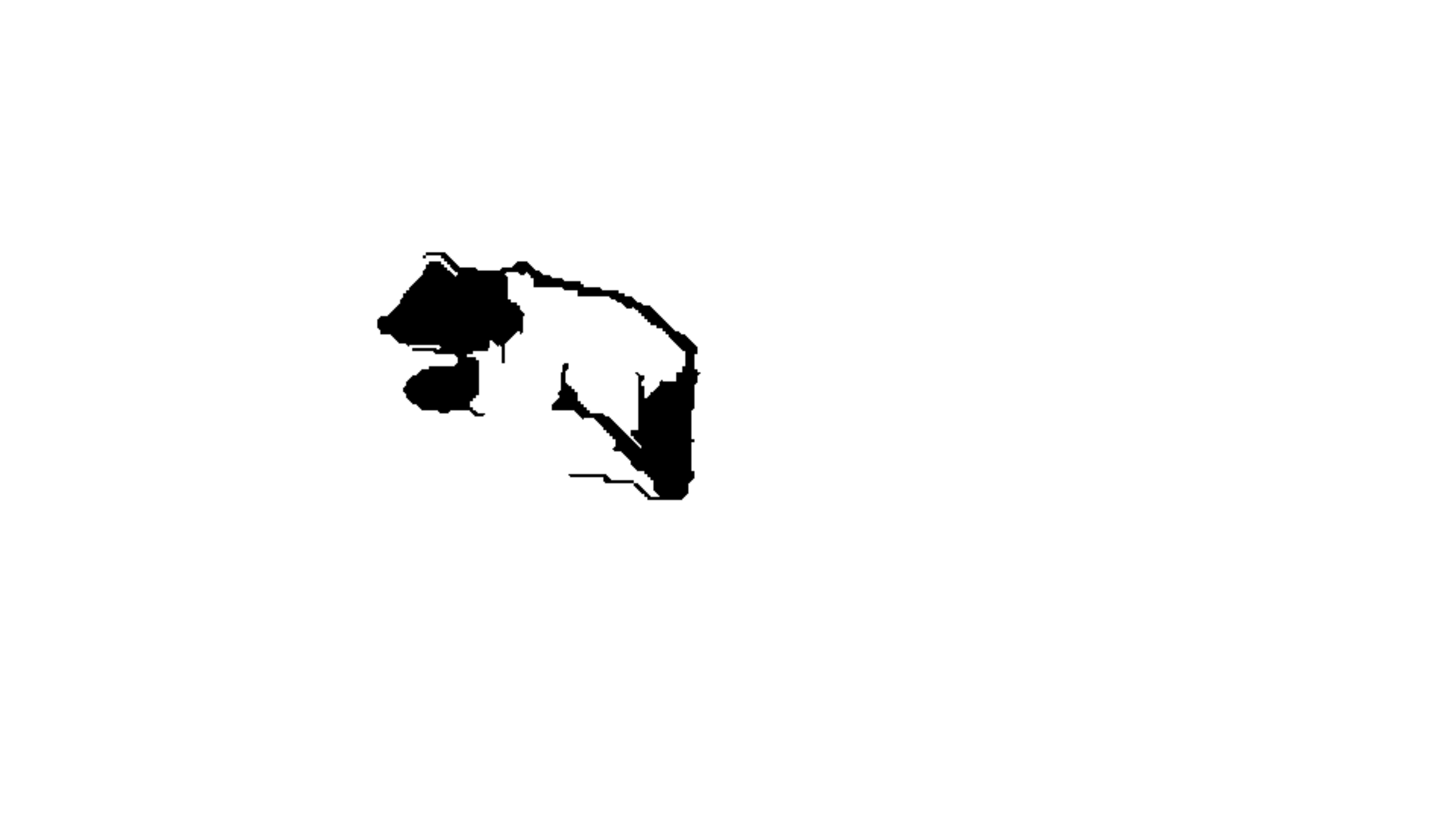}&
\includegraphics[height=5mm]{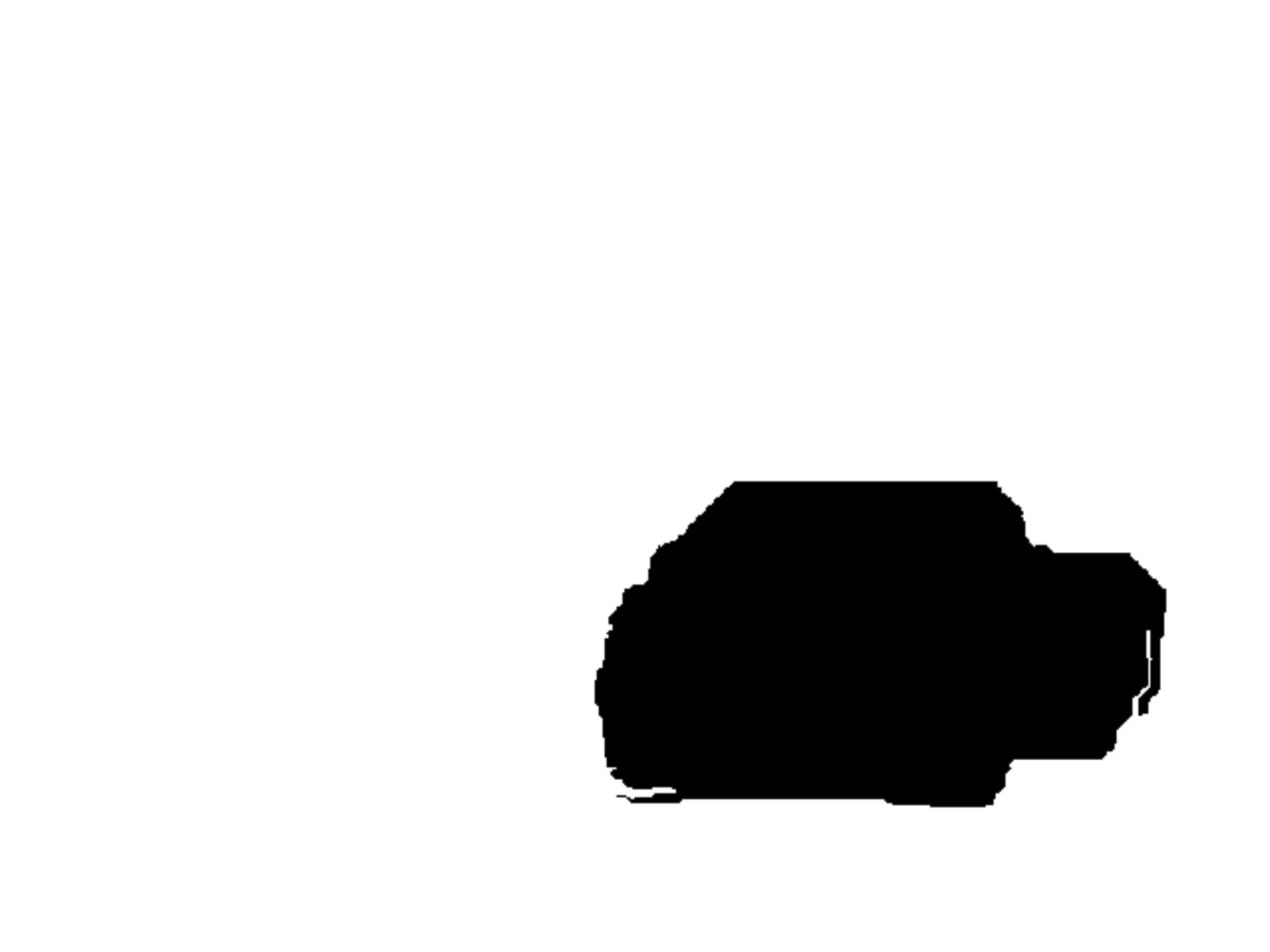}&
\includegraphics[height=5mm]{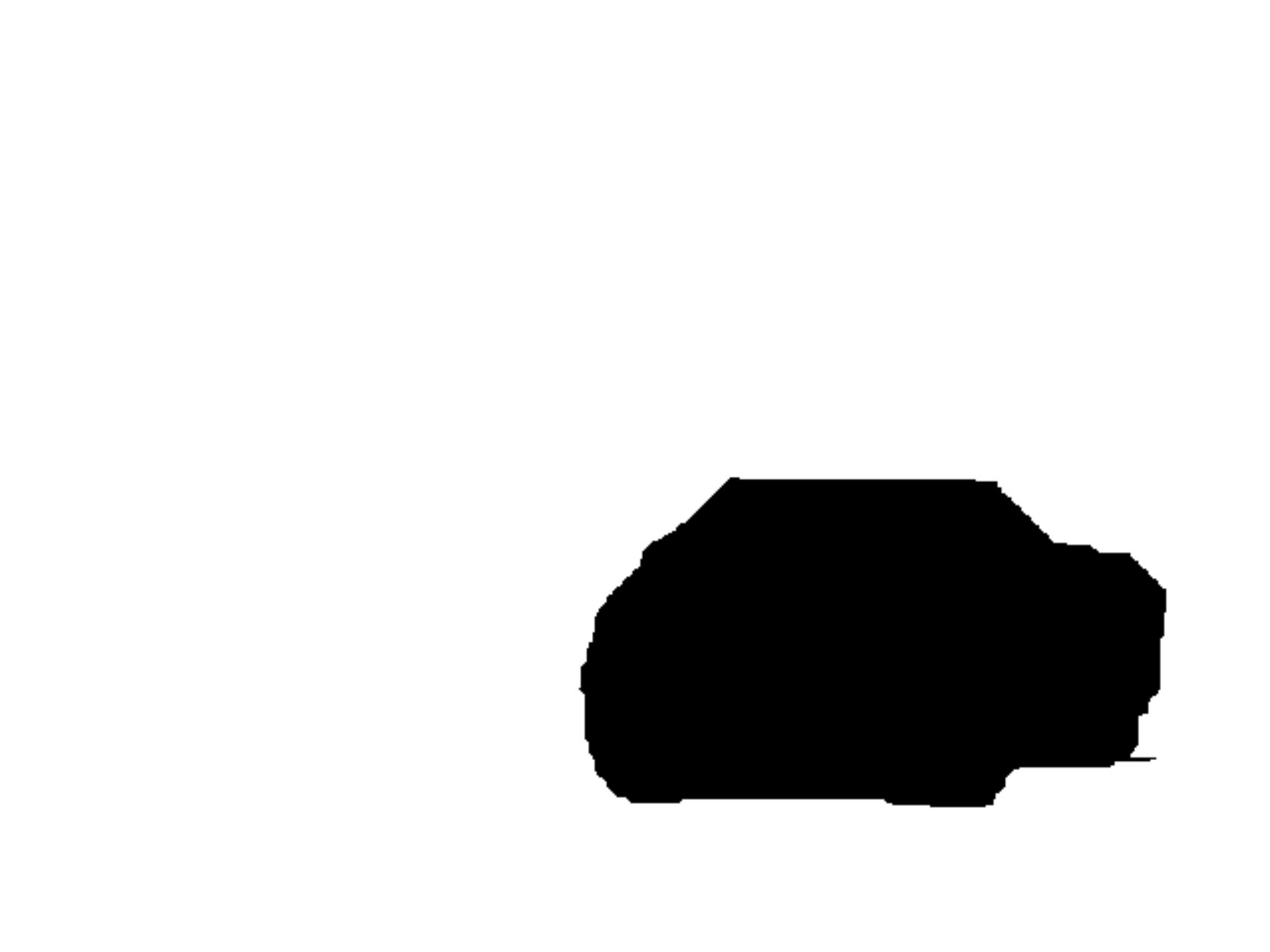}&
\includegraphics[height=5mm]{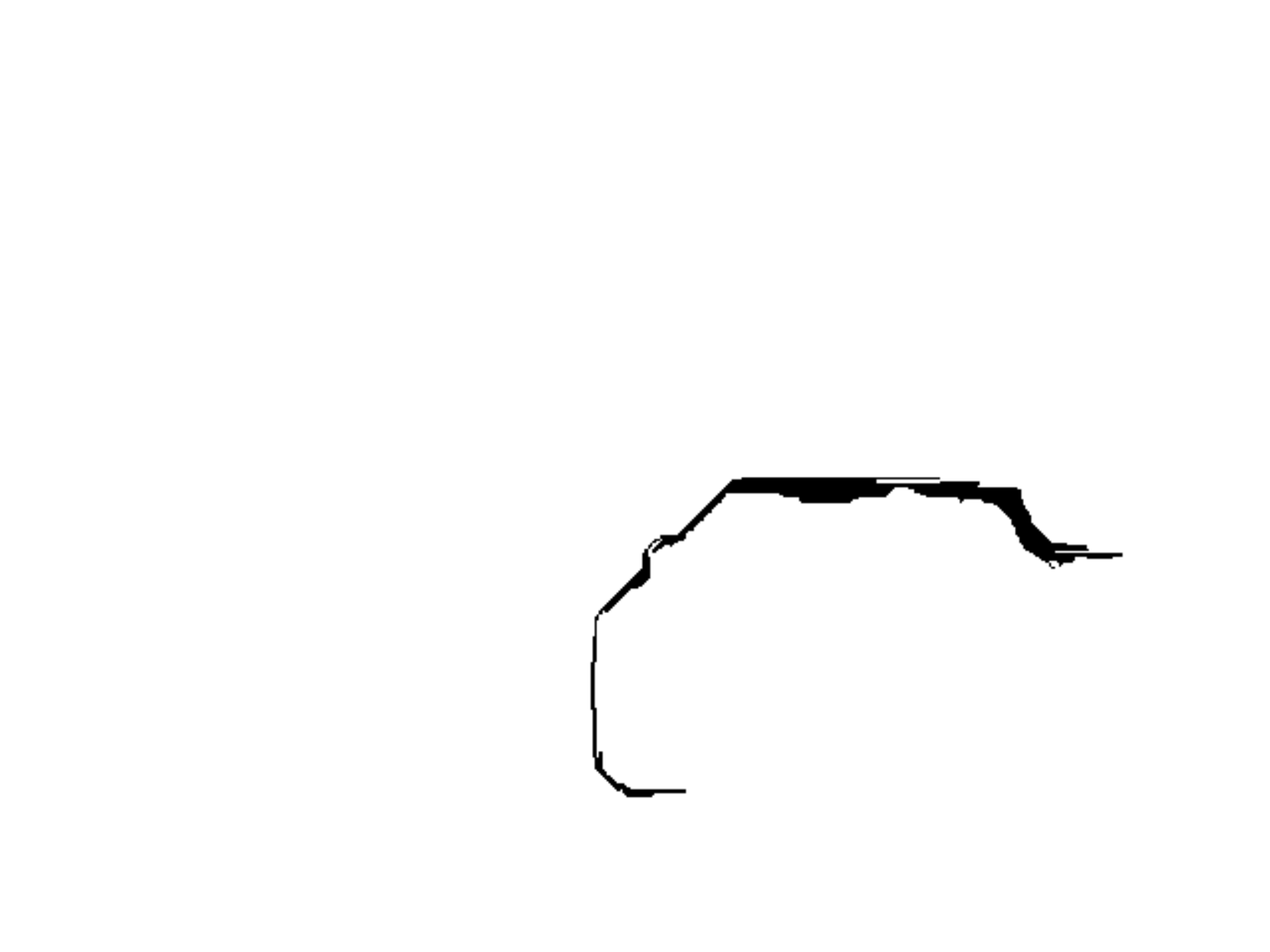}&
\includegraphics[height=5mm]{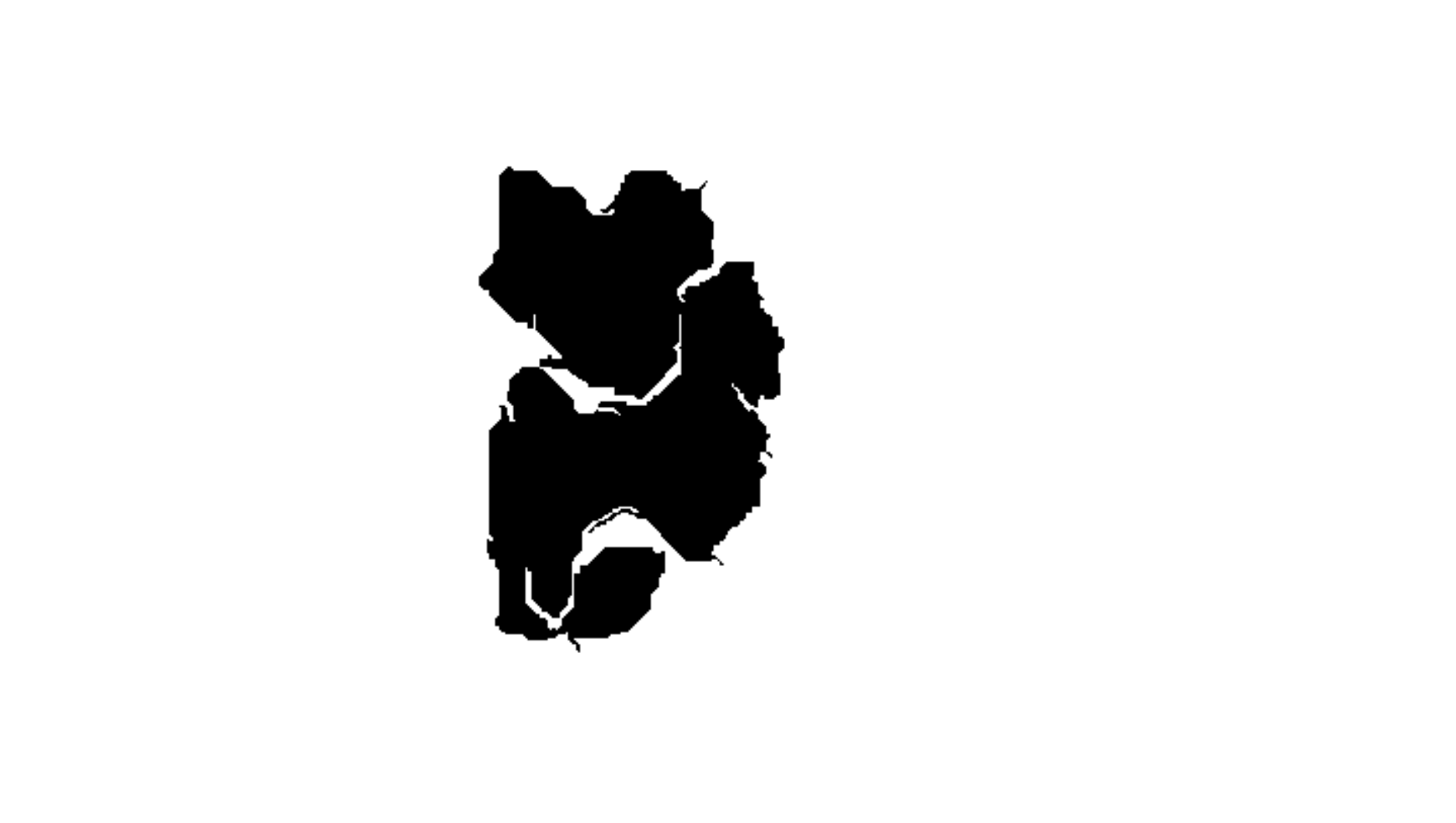}&
\includegraphics[height=5mm]{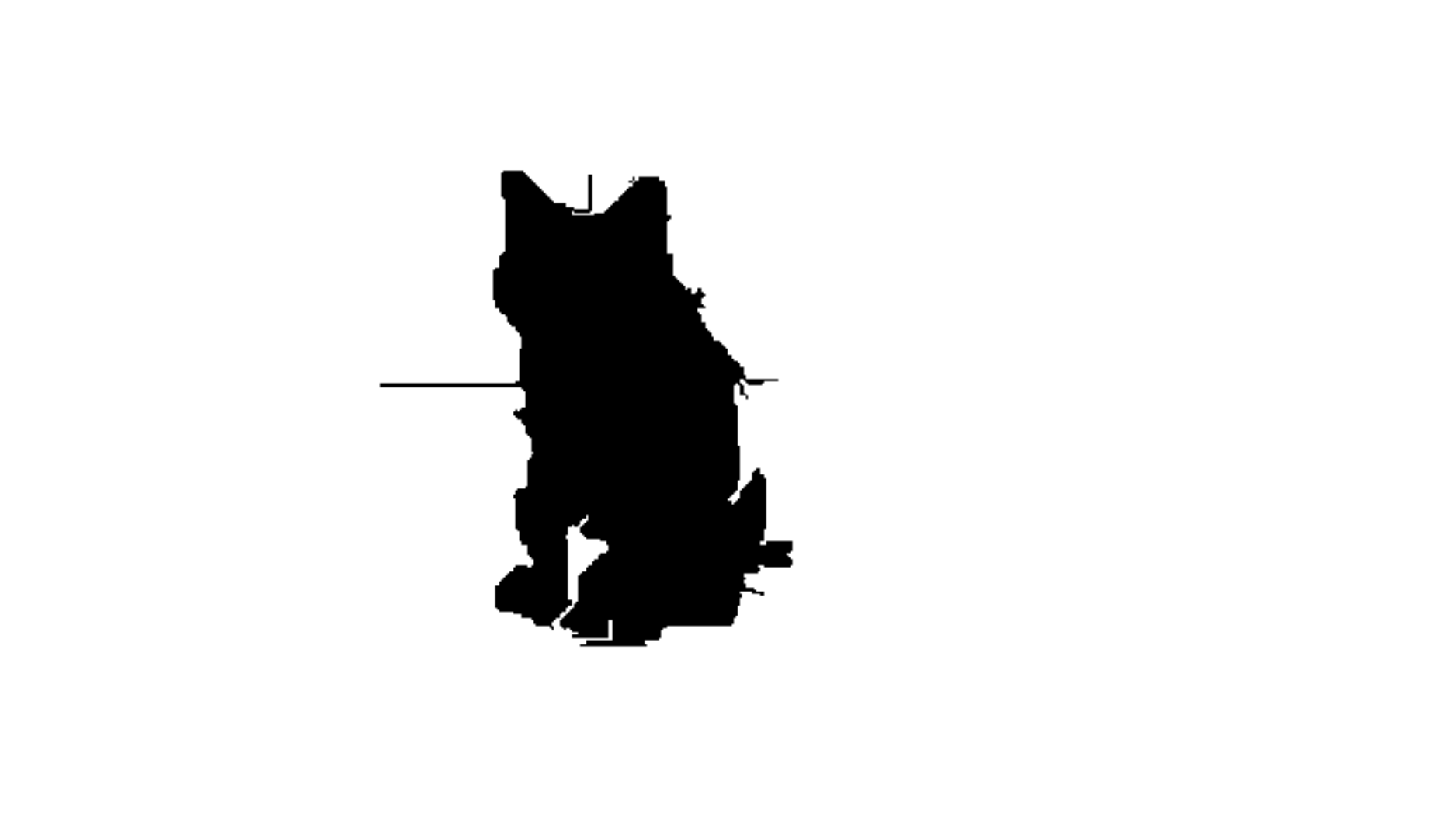}&
\includegraphics[height=5mm]{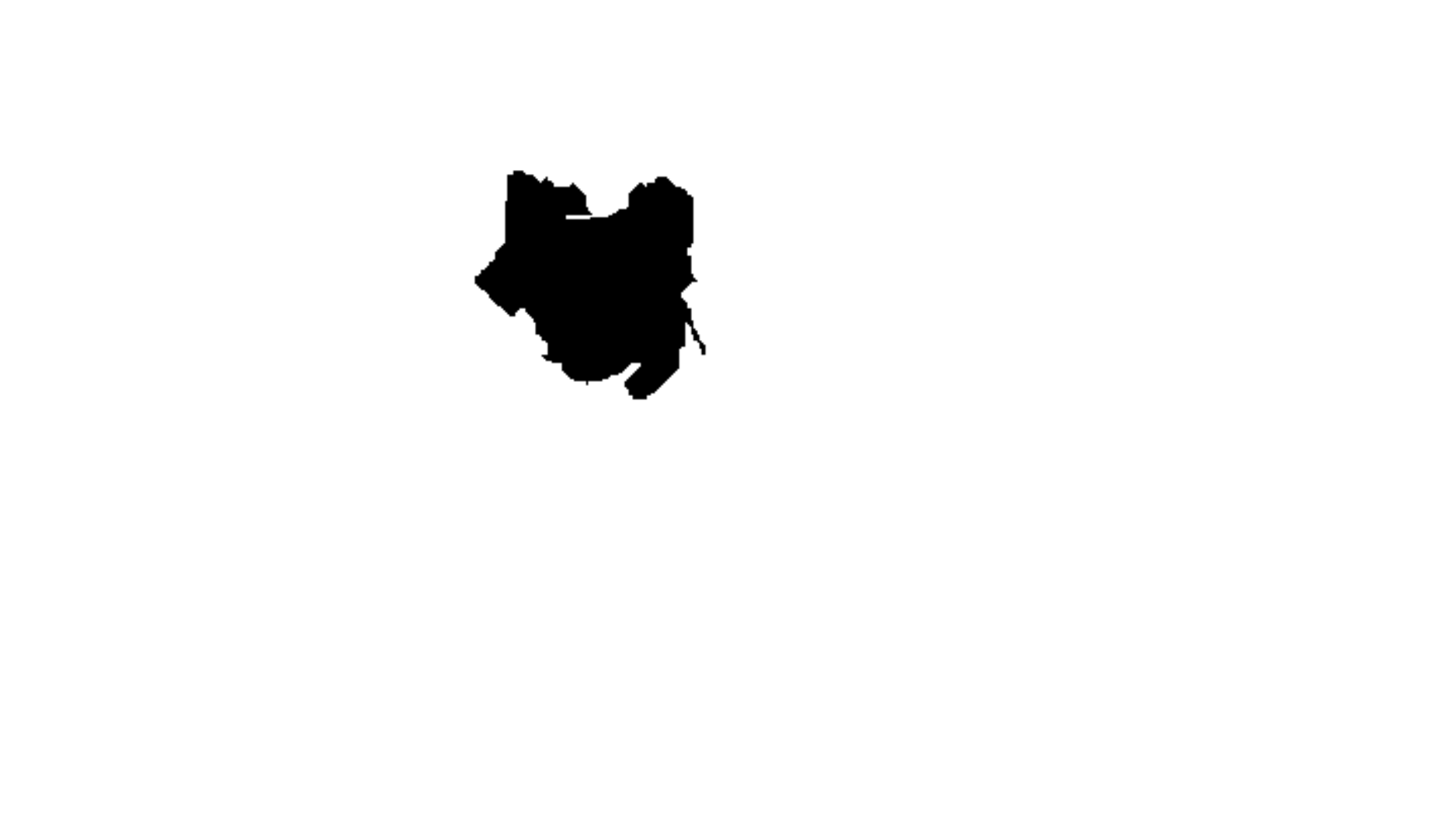}&
\includegraphics[height=5mm]{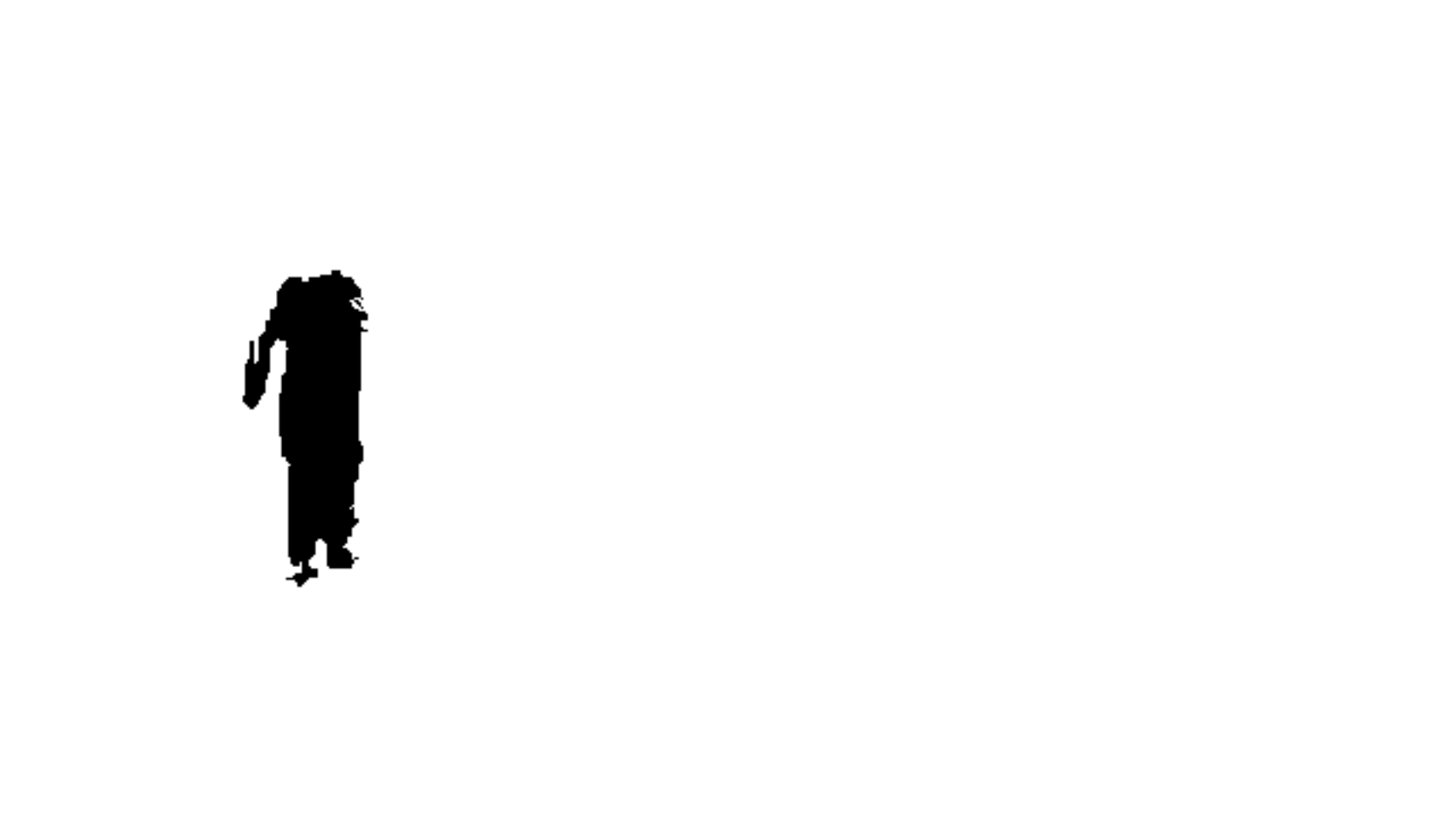}&
\includegraphics[height=5mm]{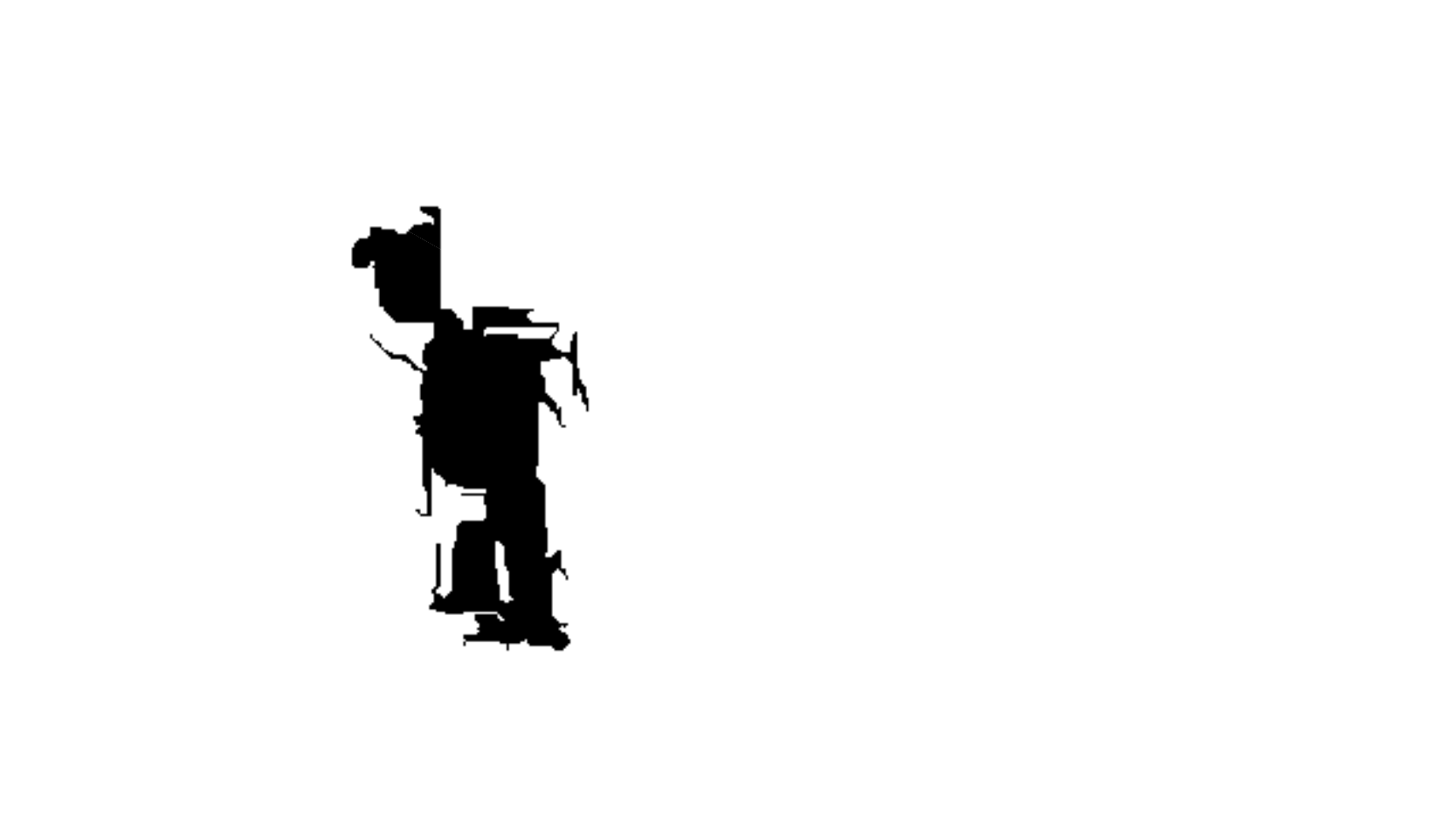}&
\includegraphics[height=5mm]{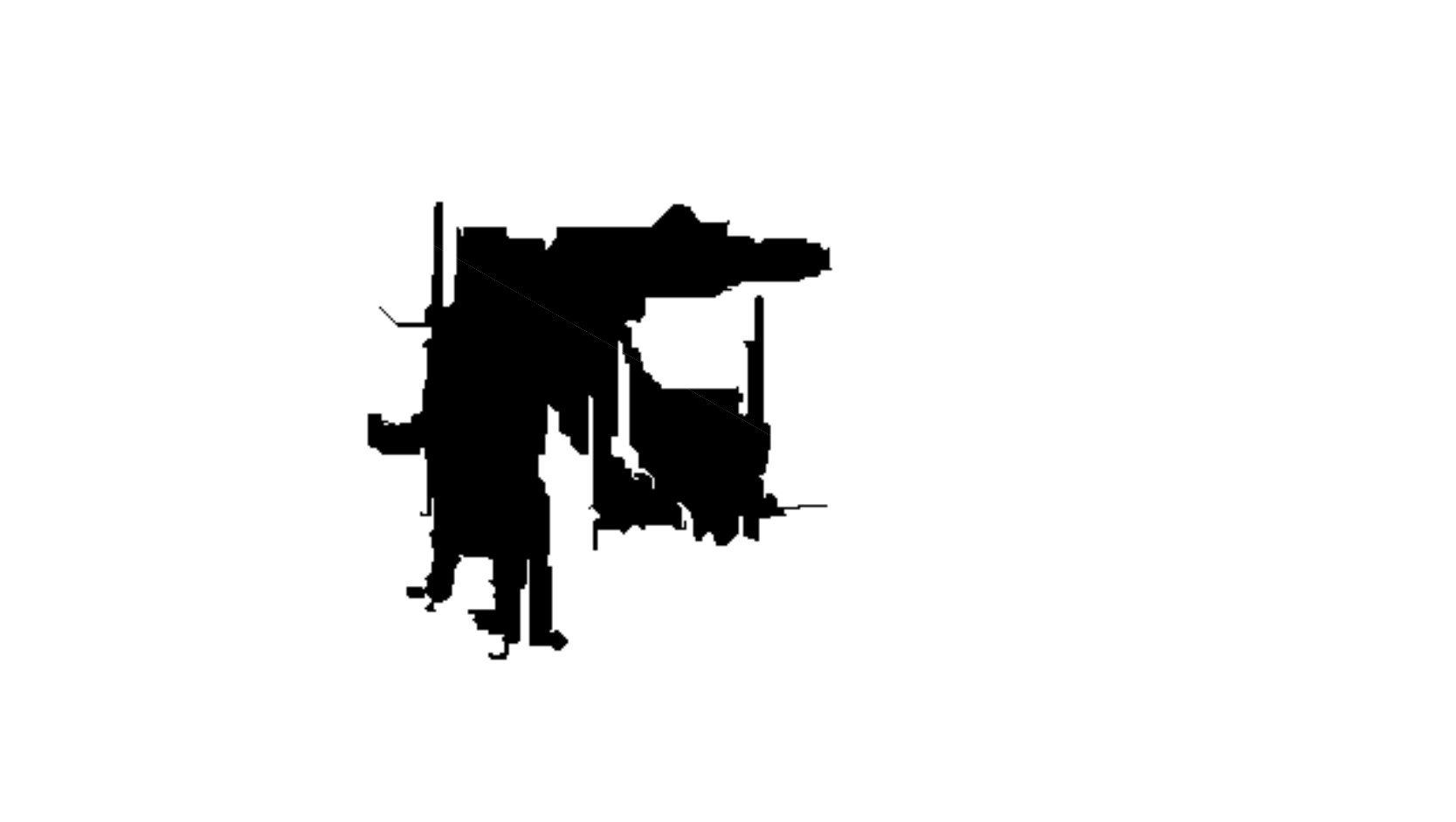}\\
\end{tabular}
\begin{tabular}{@{}c@{}c@{}c@{}c@{}}
\includegraphics[height=15mm]{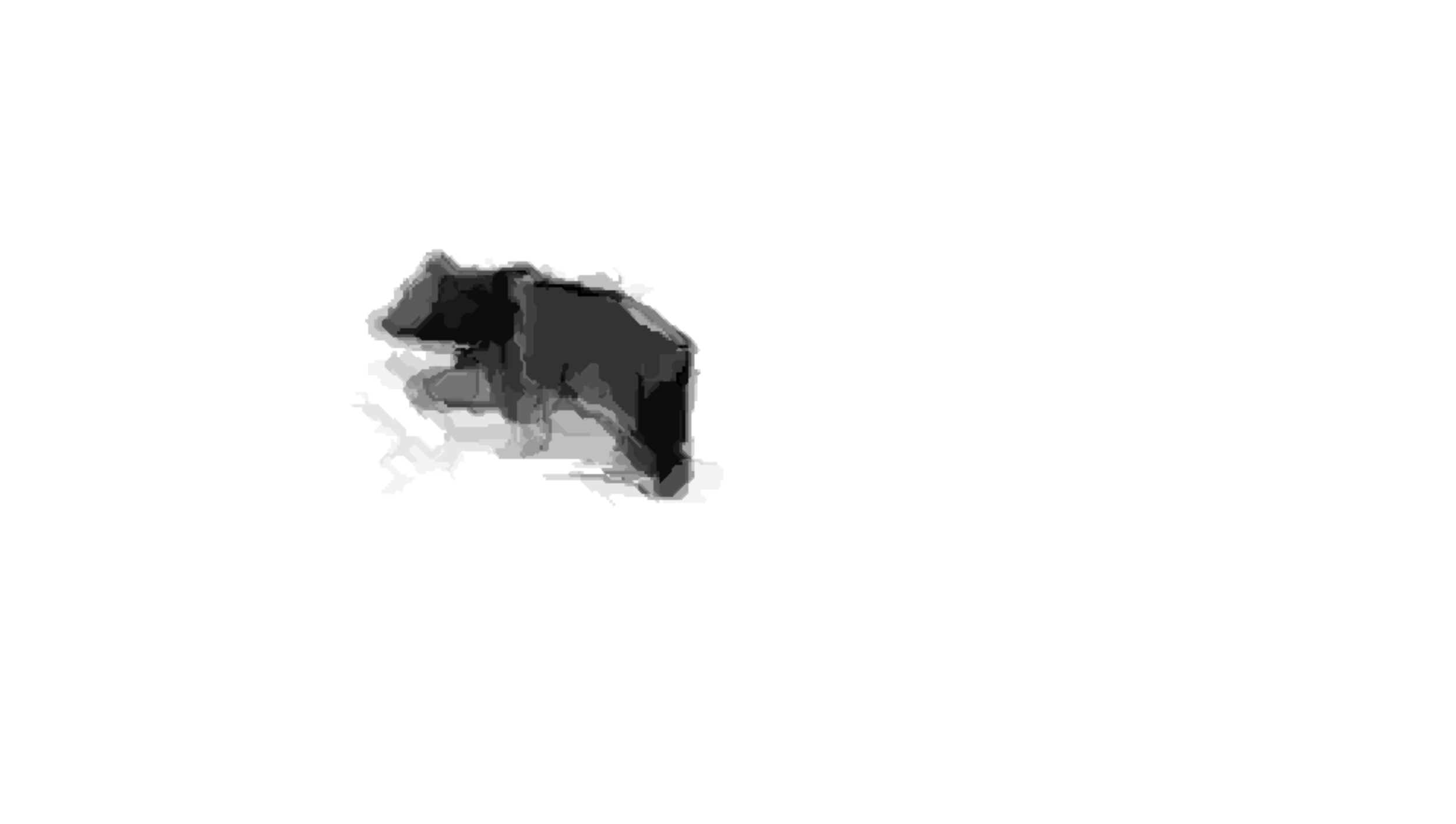}&
\includegraphics[height=15mm]{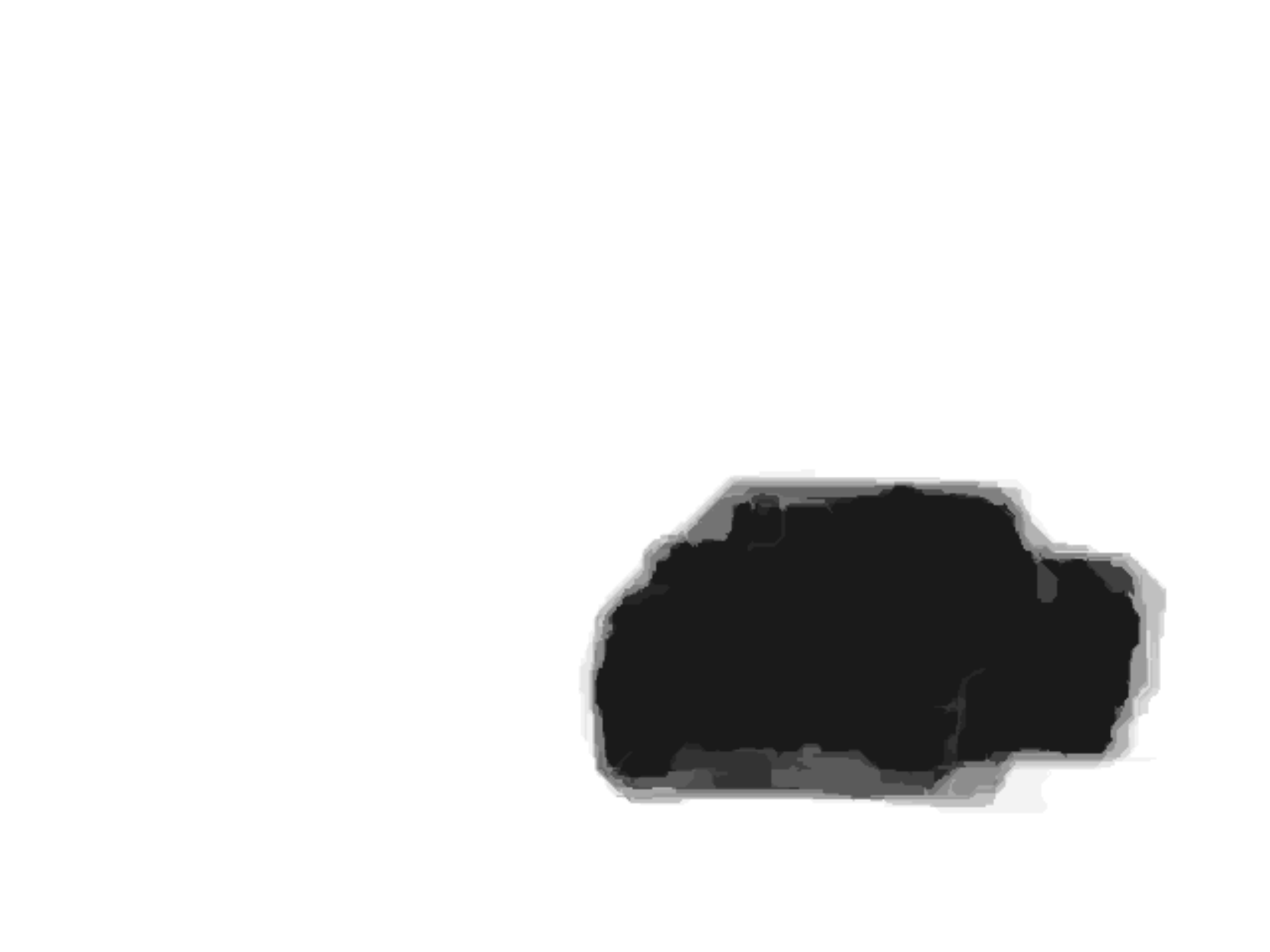}&
\includegraphics[height=15mm]{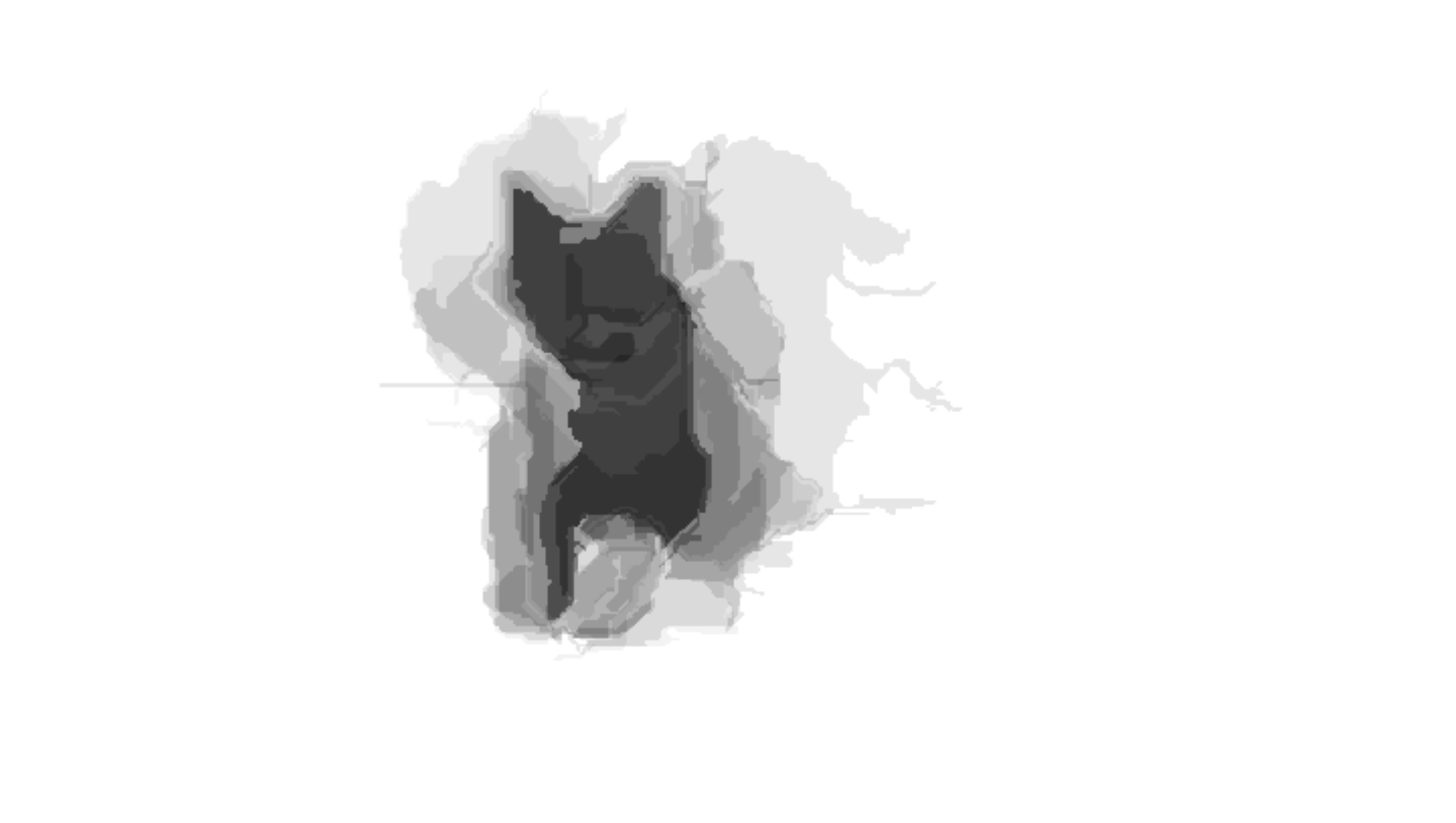}&
\includegraphics[height=15mm]{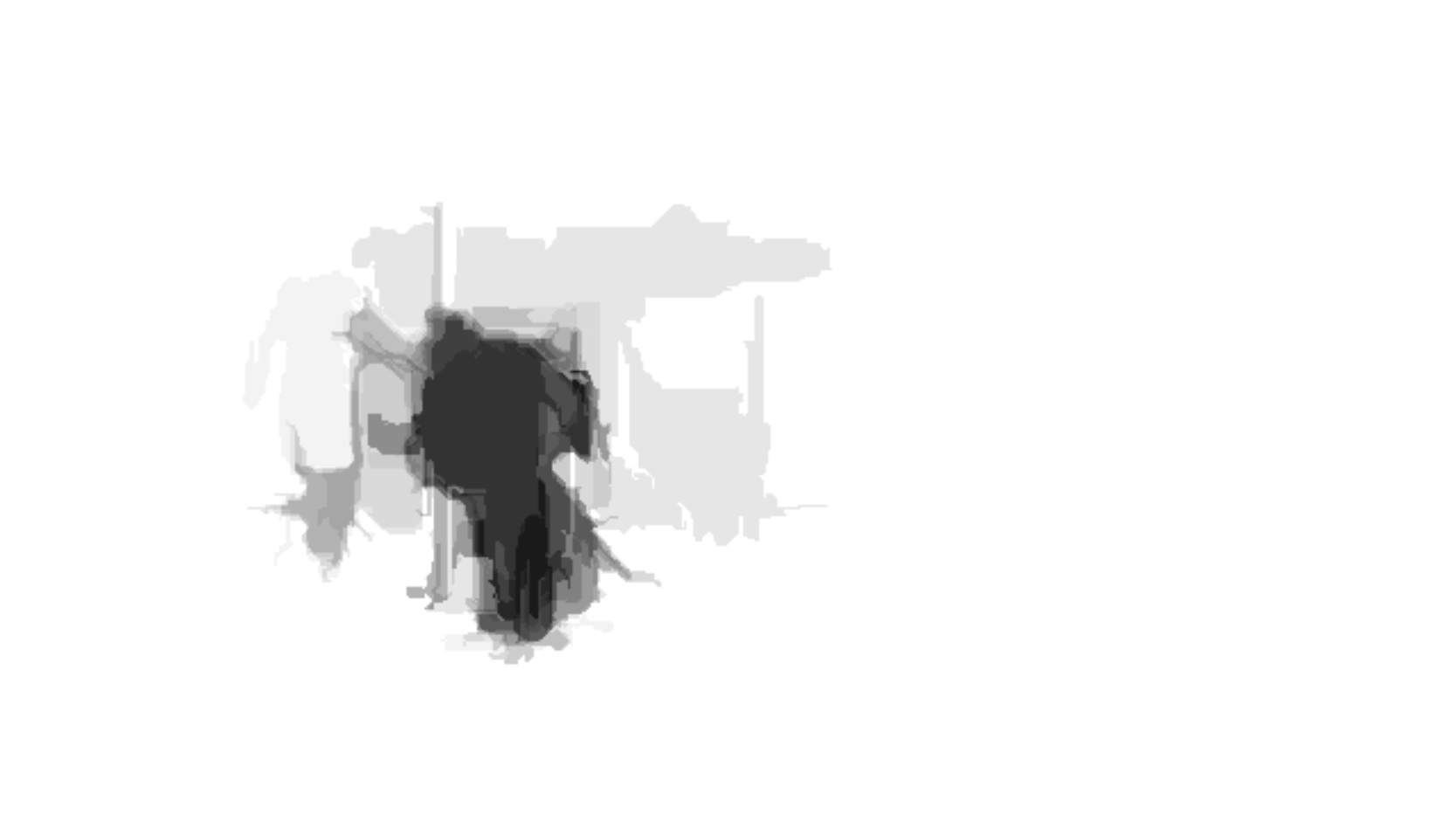}\\
\end{tabular}
\vspace{0.1cm}
\rotatebox{0}{faster R-CNN + deepLab}
\begin{tabular}{@{}c@{}c@{}c@{}c@{}}
\includegraphics[height=15mm]{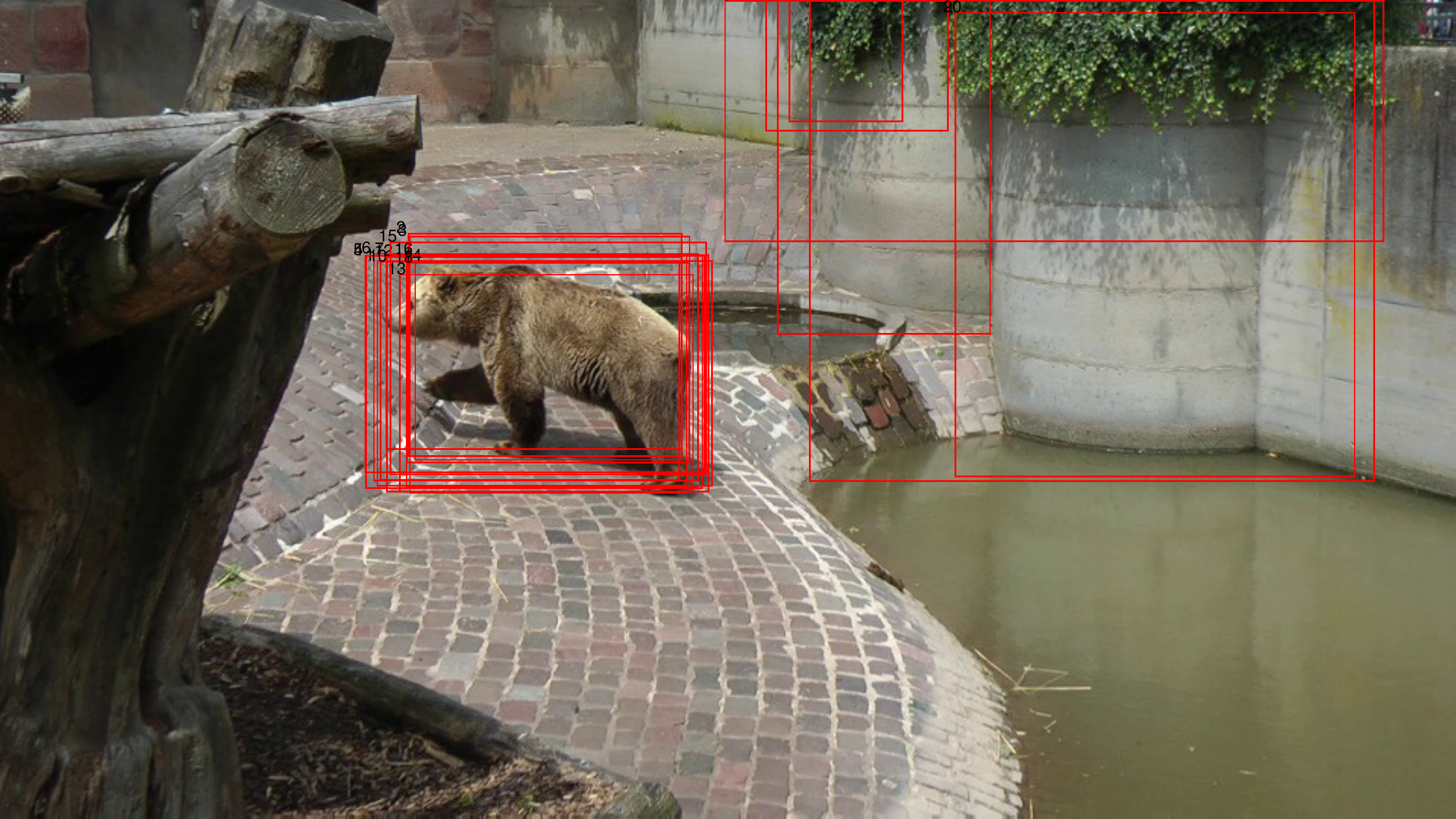}&
\includegraphics[height=15mm]{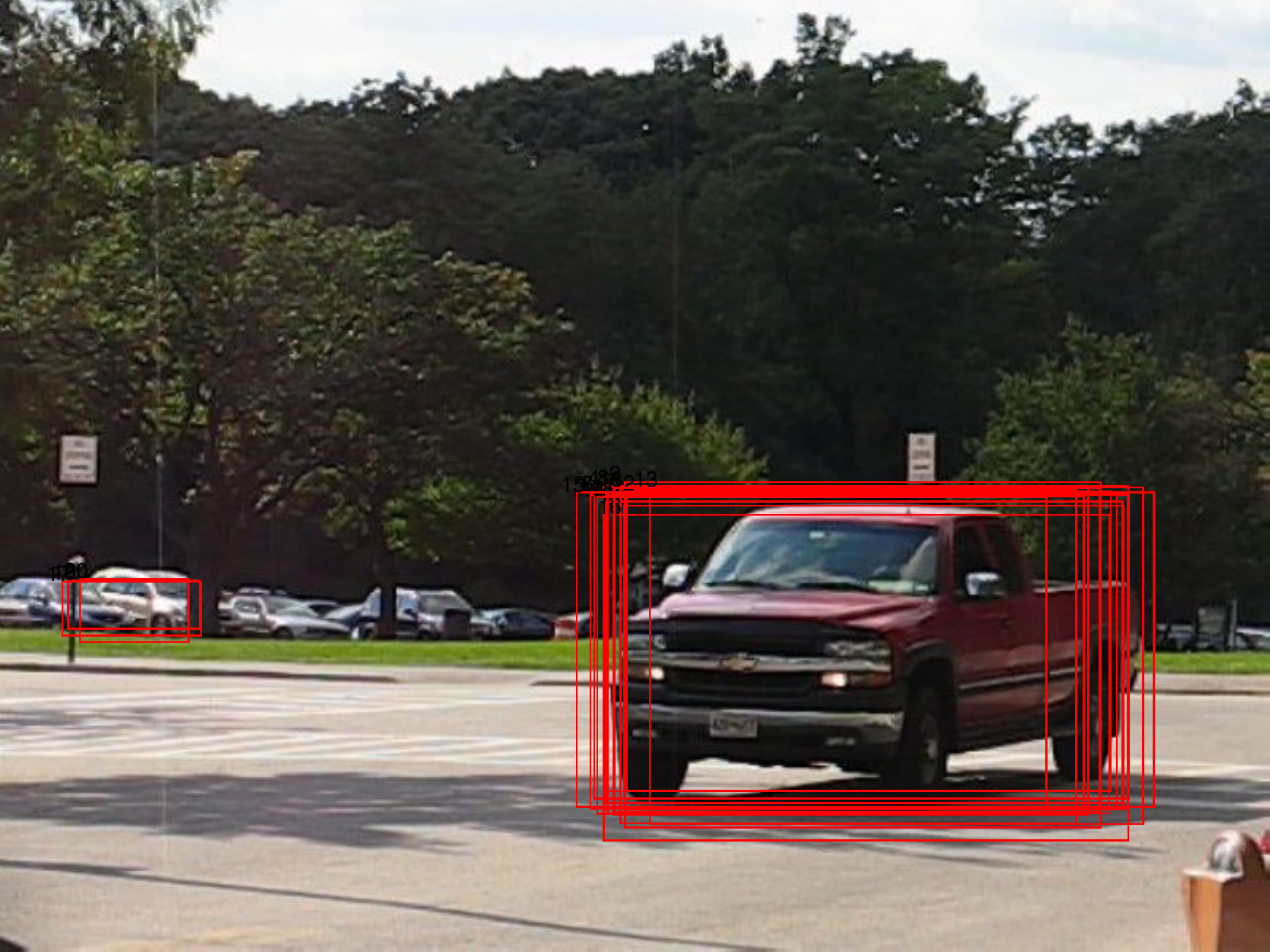}&
\includegraphics[height=15mm]{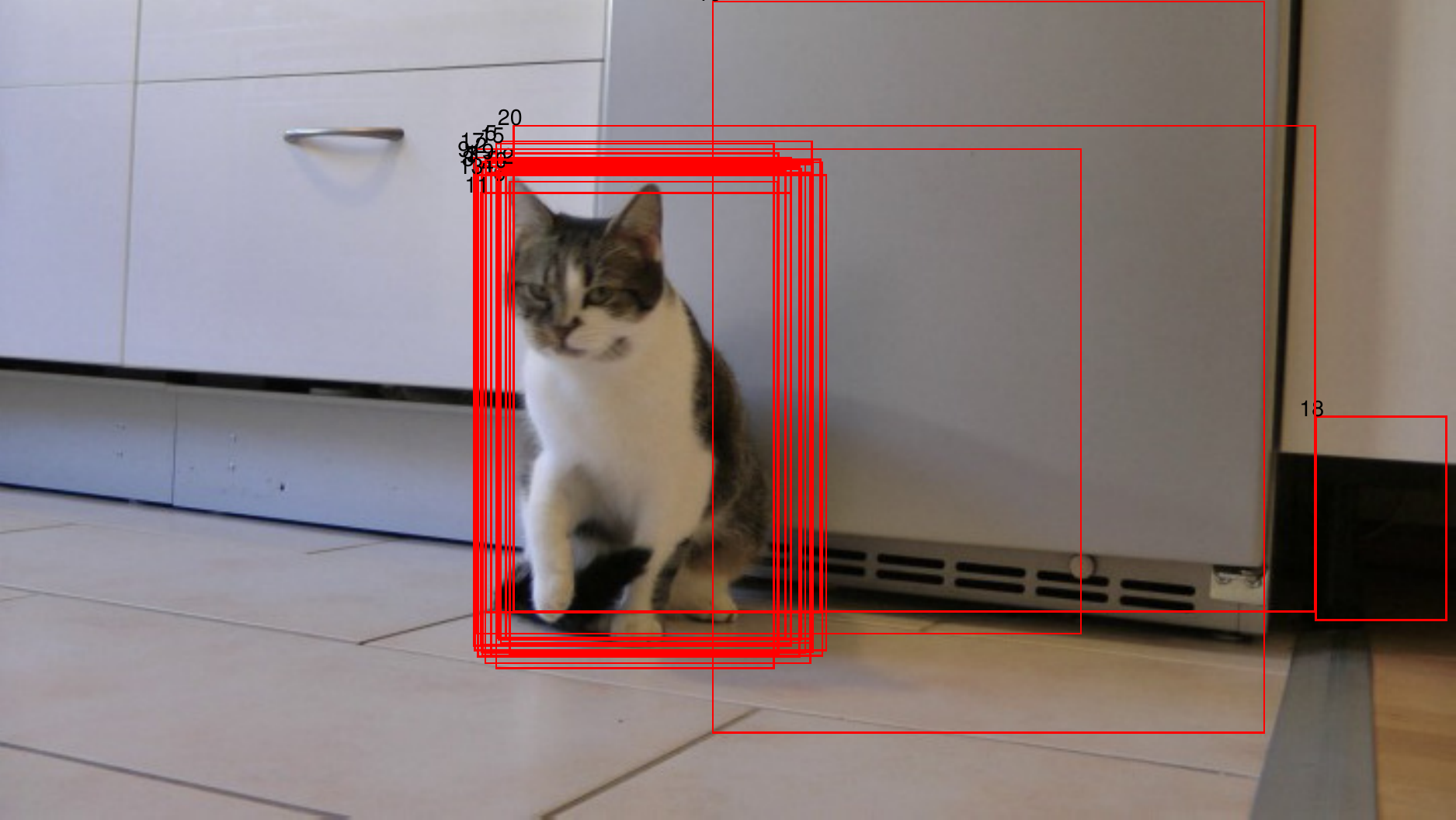}&
\includegraphics[height=15mm]{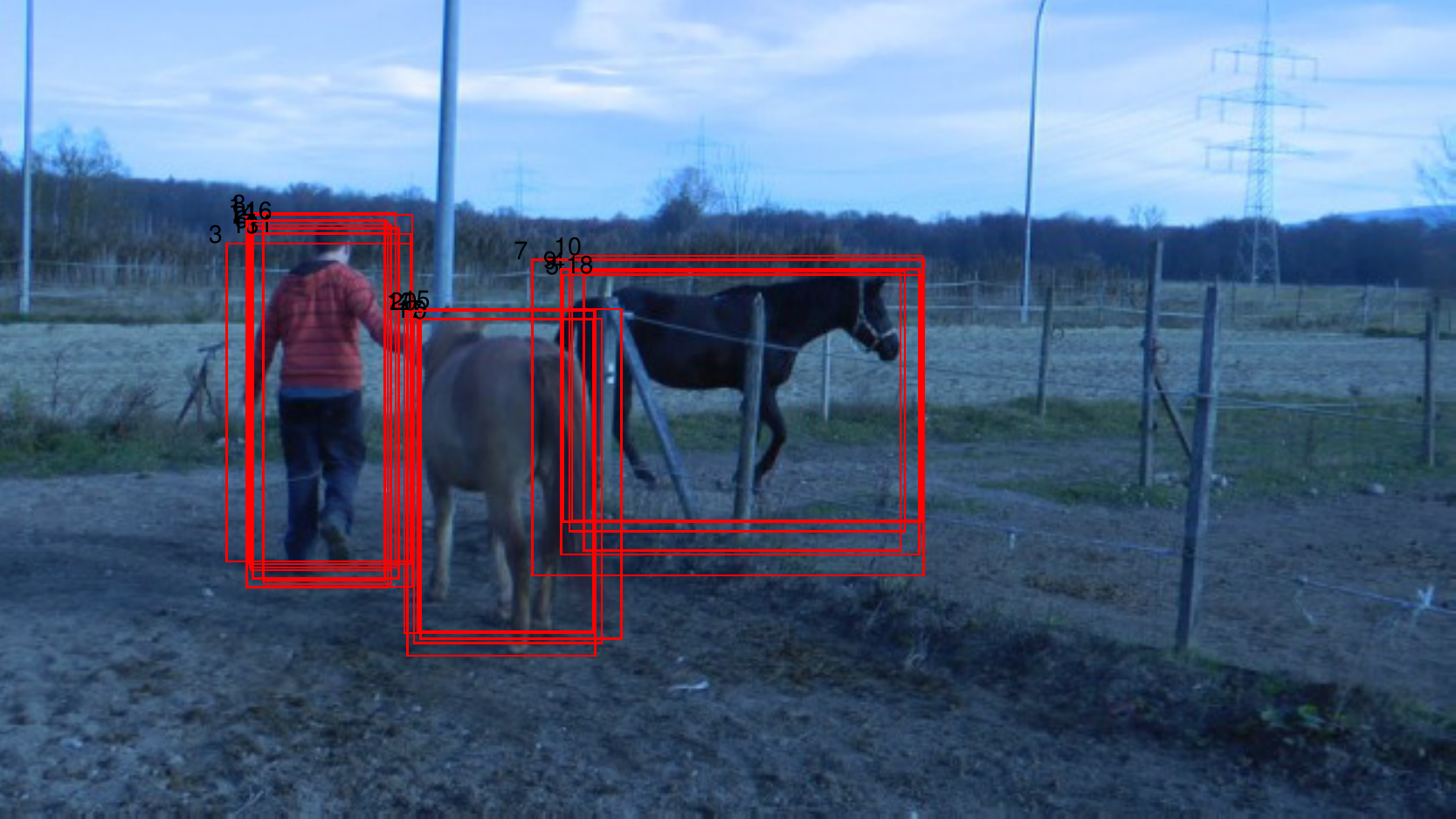}\\
\end{tabular}
\begin{tabular}{@{}c@{}c@{}c@{}c@{}c@{}c@{}c@{}c@{}c@{}c@{}c@{}c@{}c@{}c@{}c@{}}
\includegraphics[height=5mm]{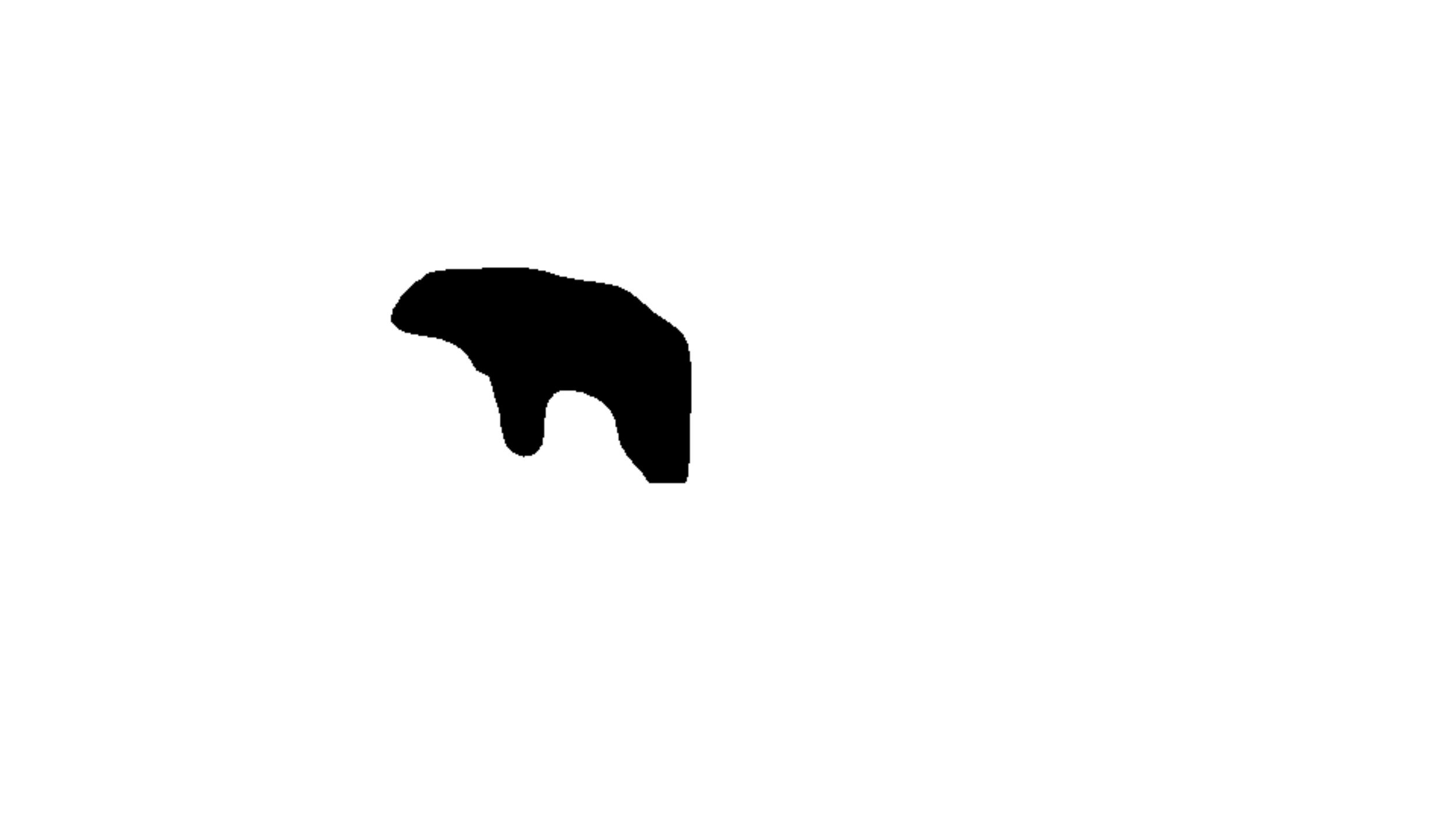}&
\includegraphics[height=5mm]{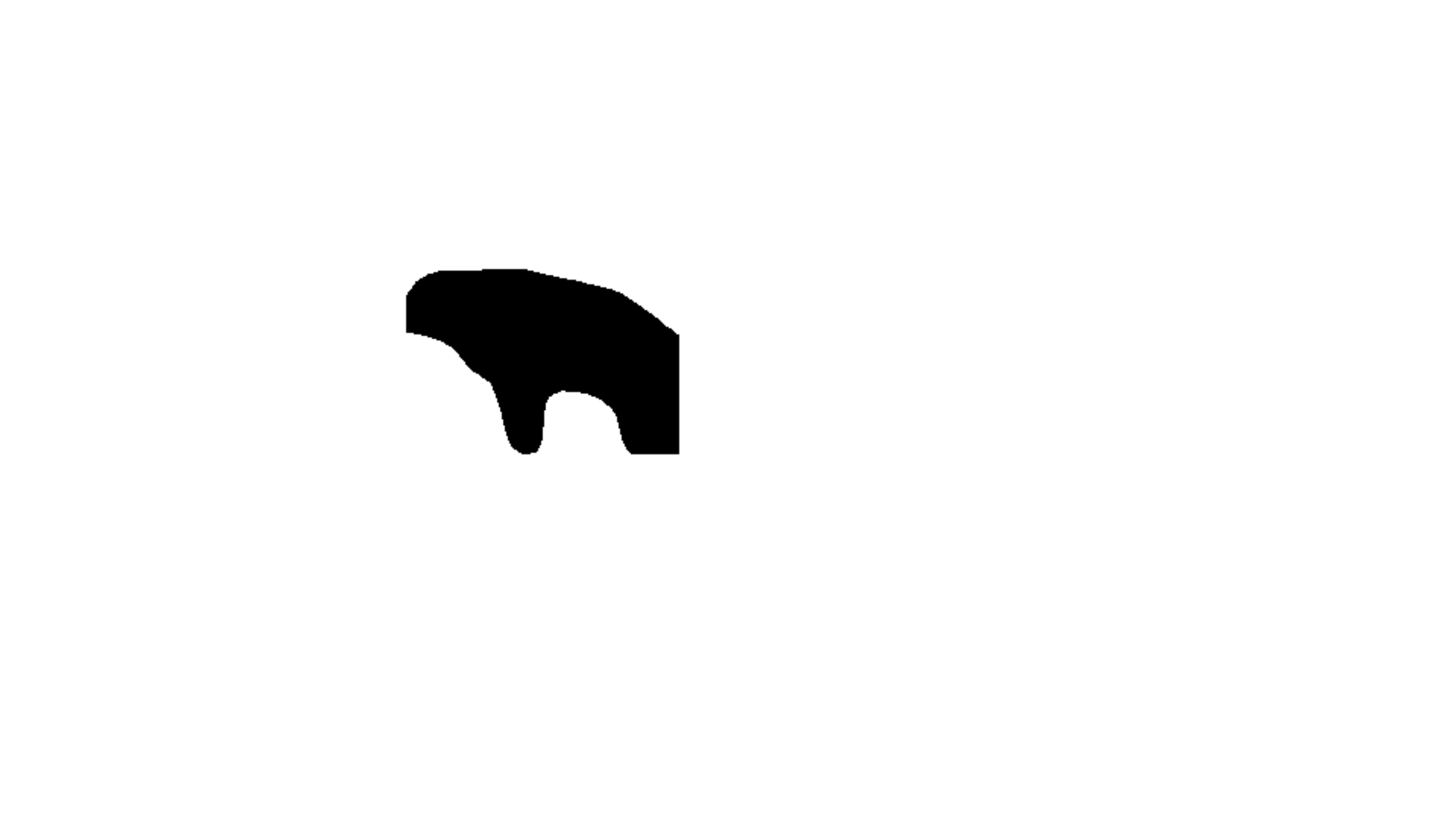}&
\includegraphics[height=5mm]{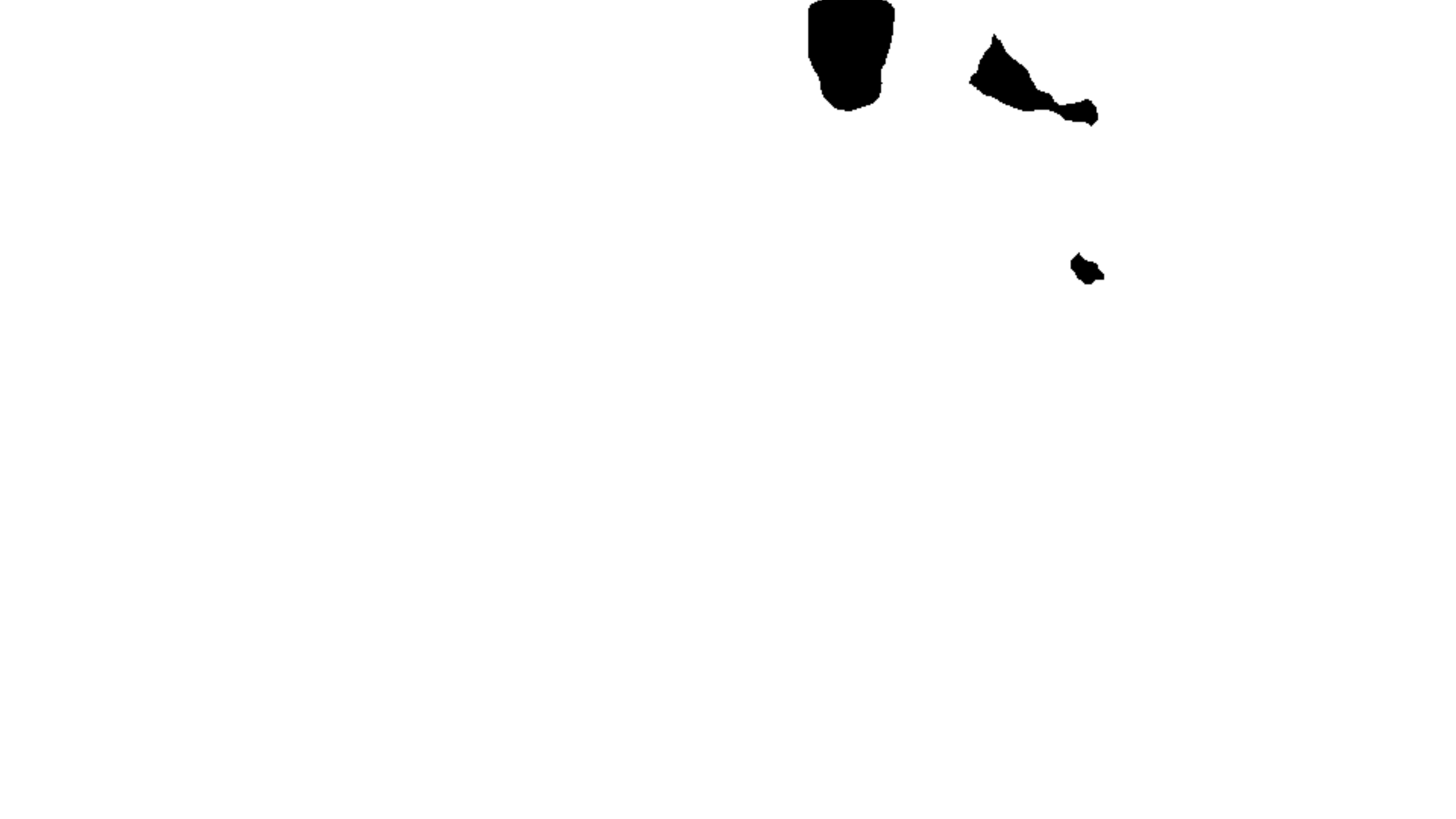}&
\includegraphics[height=5mm]{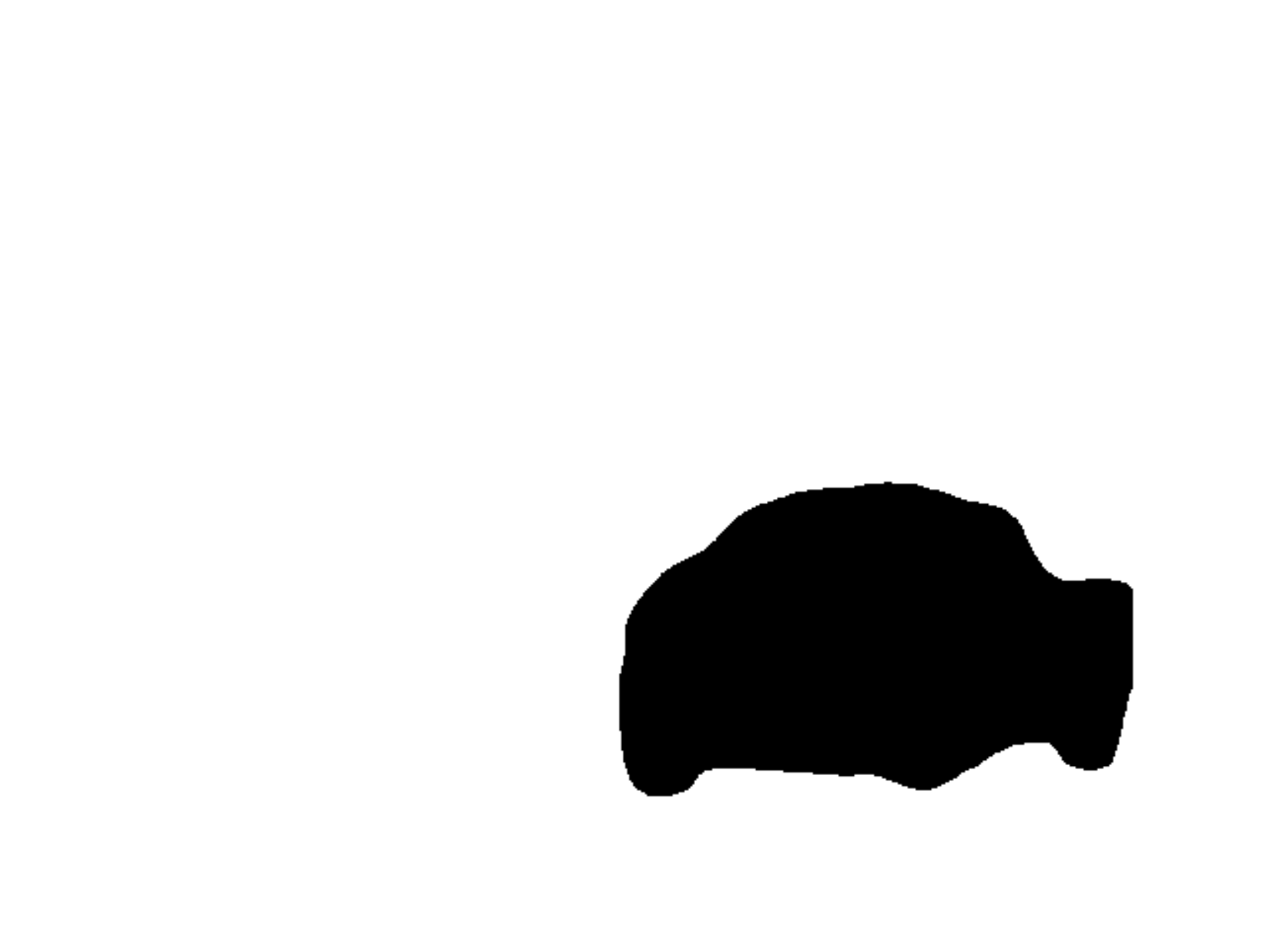}&
\includegraphics[height=5mm]{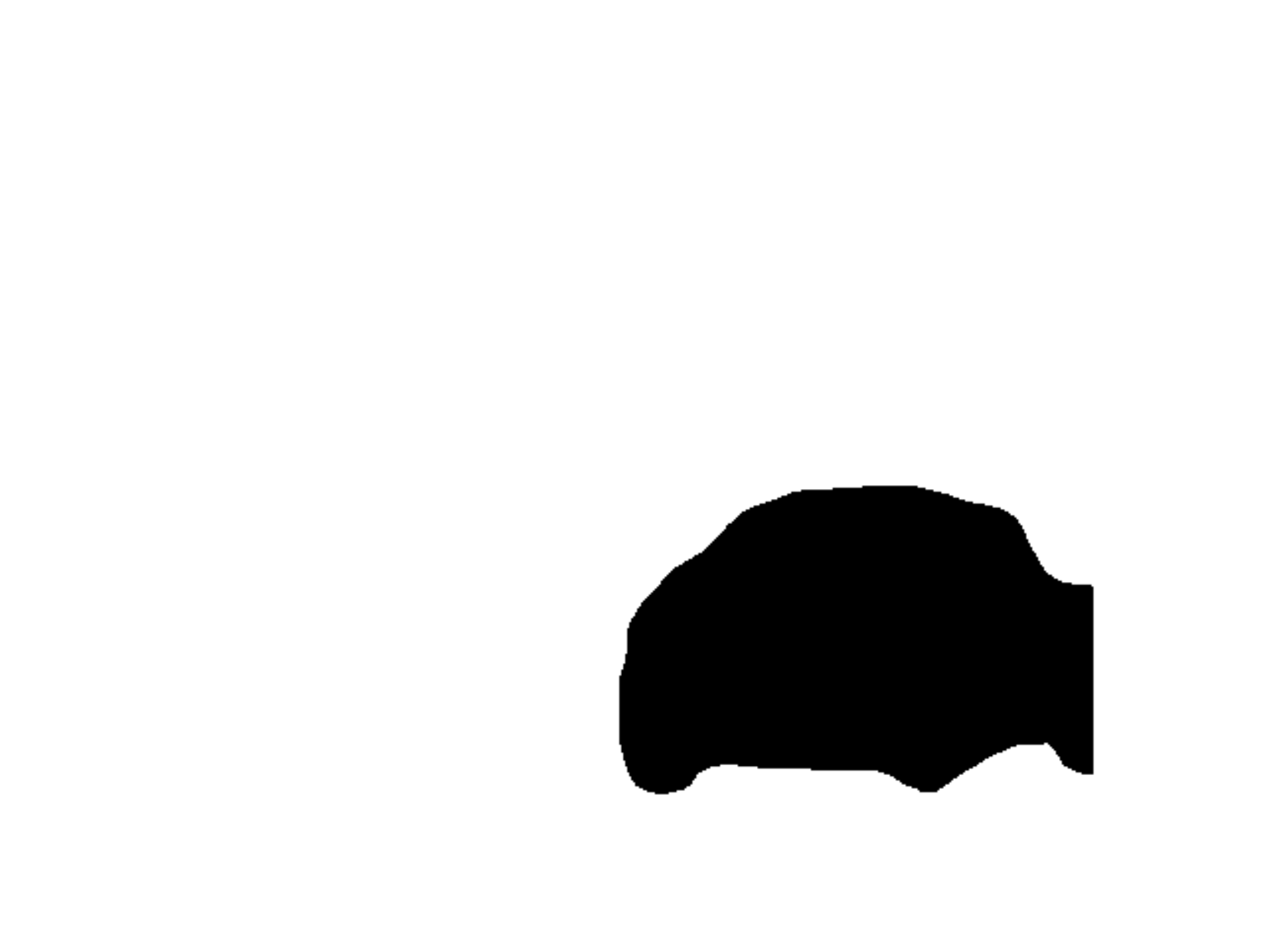}&
\includegraphics[height=5mm]{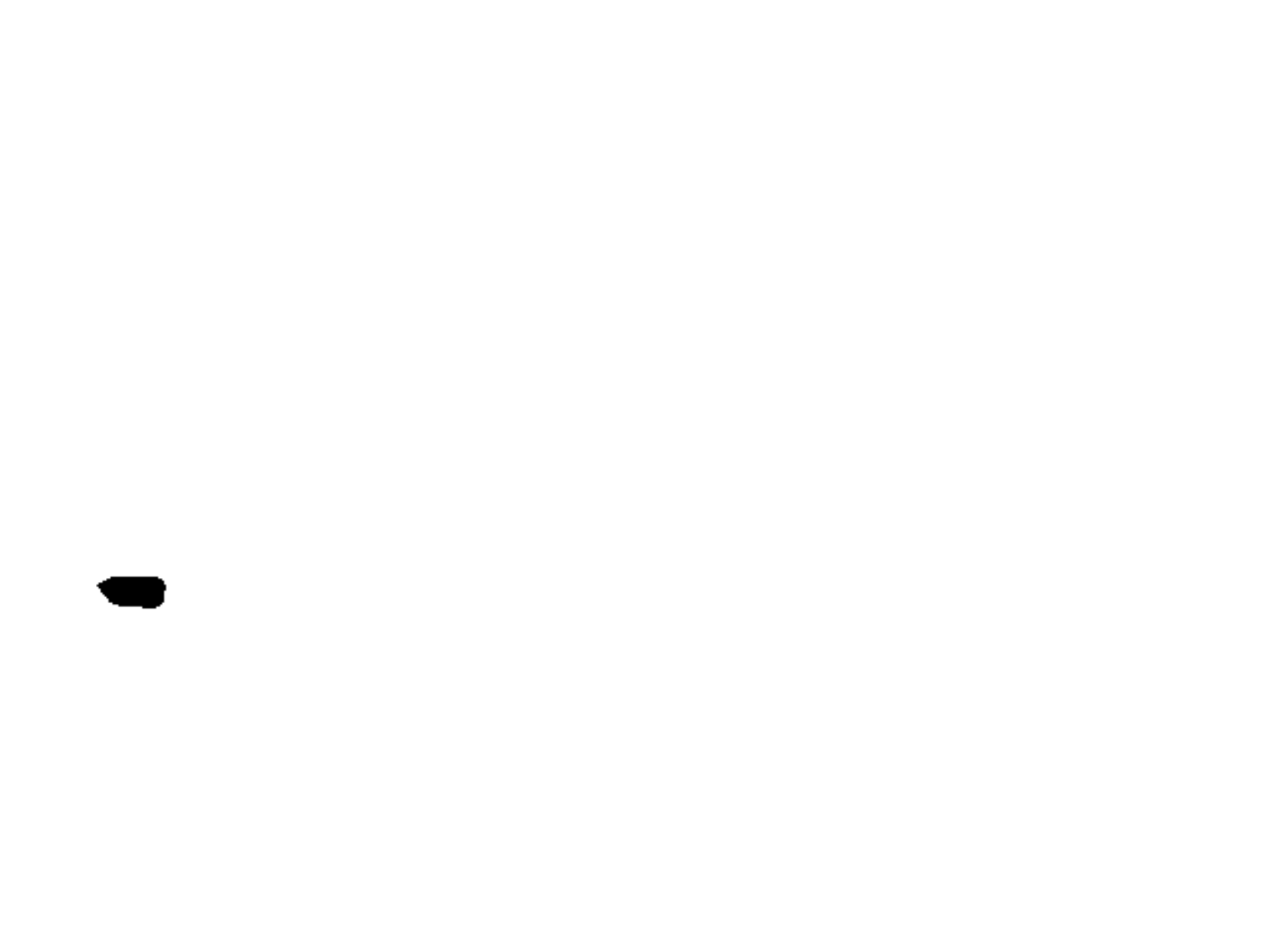}&
\includegraphics[height=5mm]{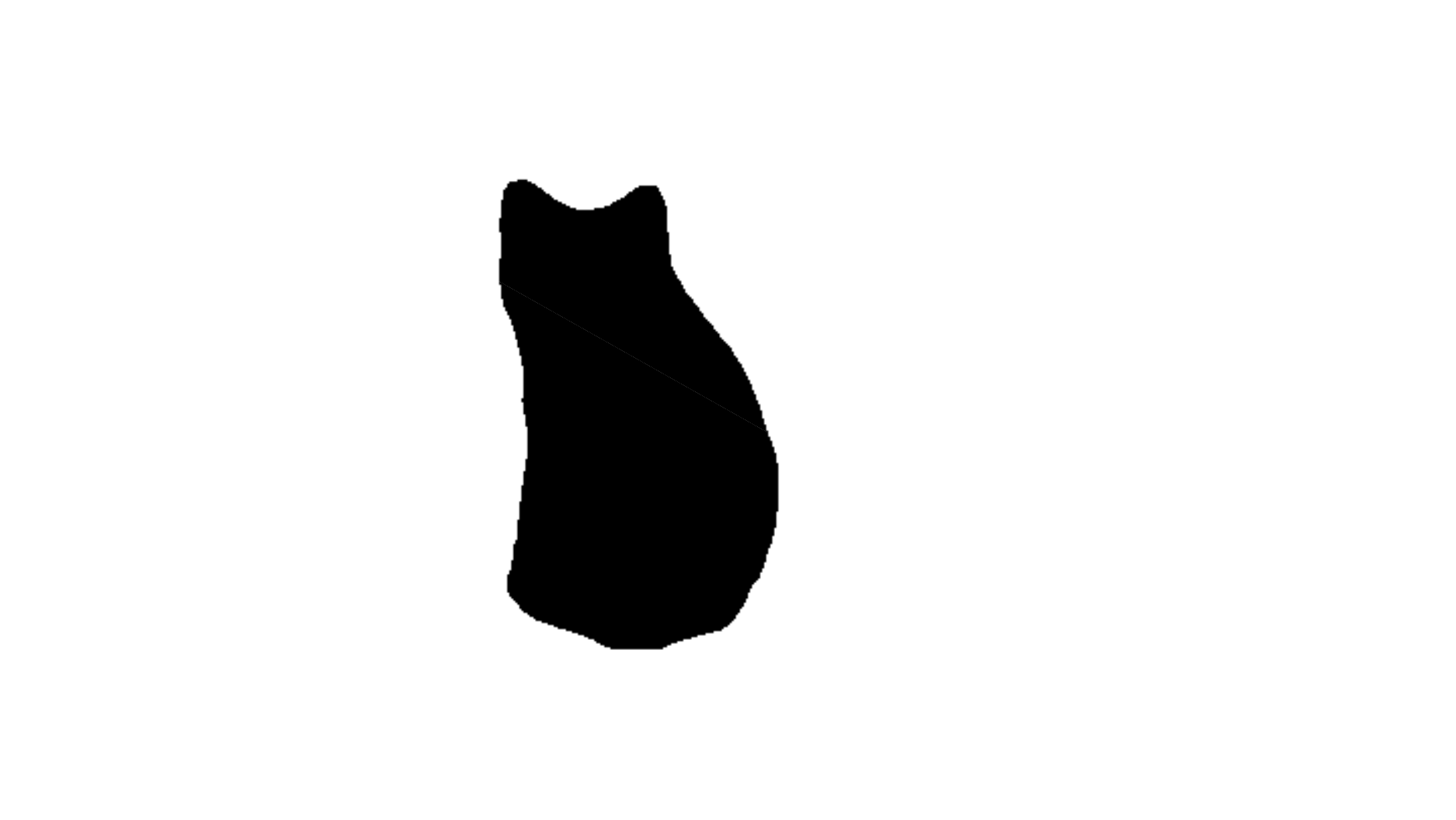}&
\includegraphics[height=5mm]{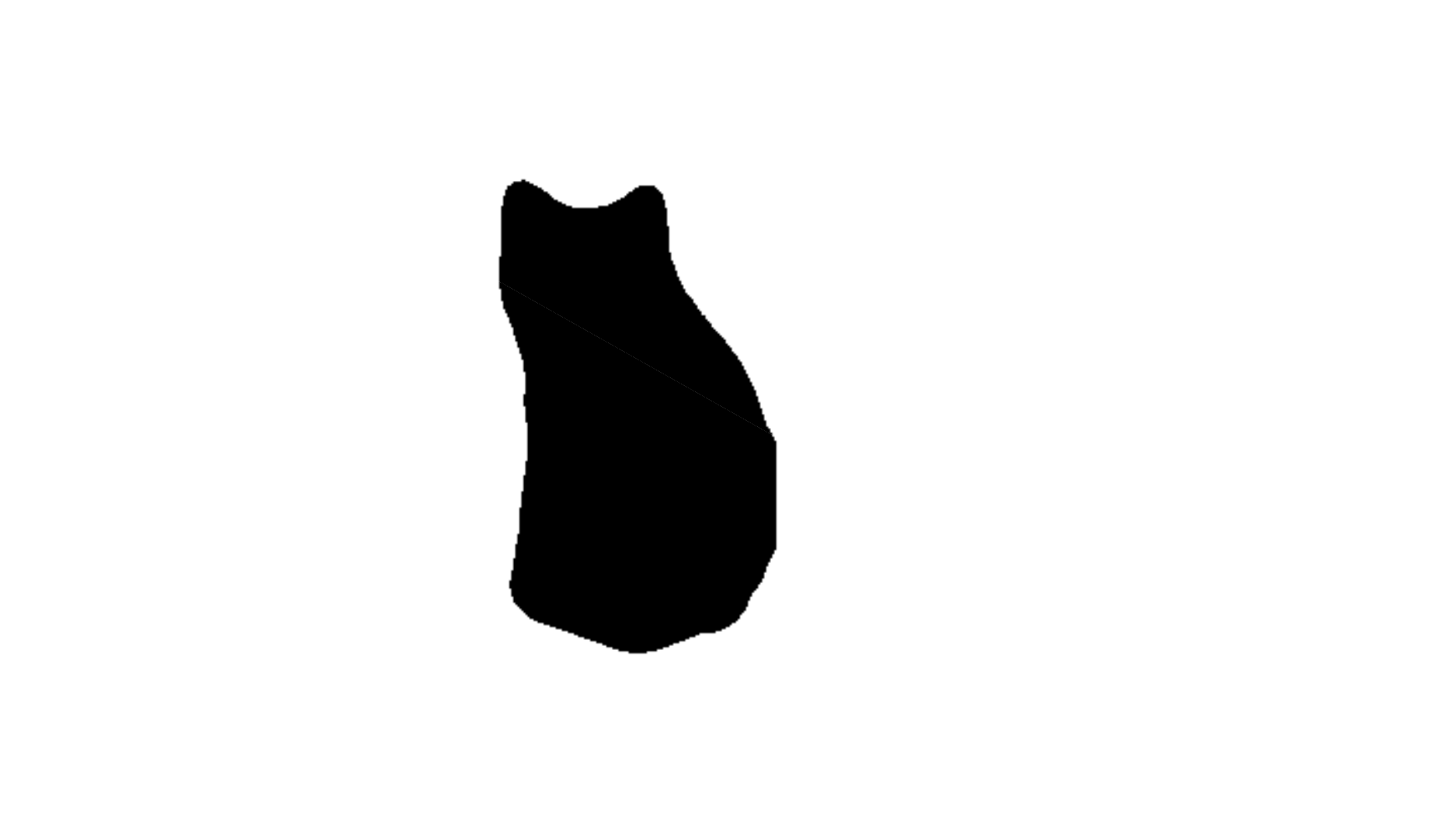}&
\includegraphics[height=5mm]{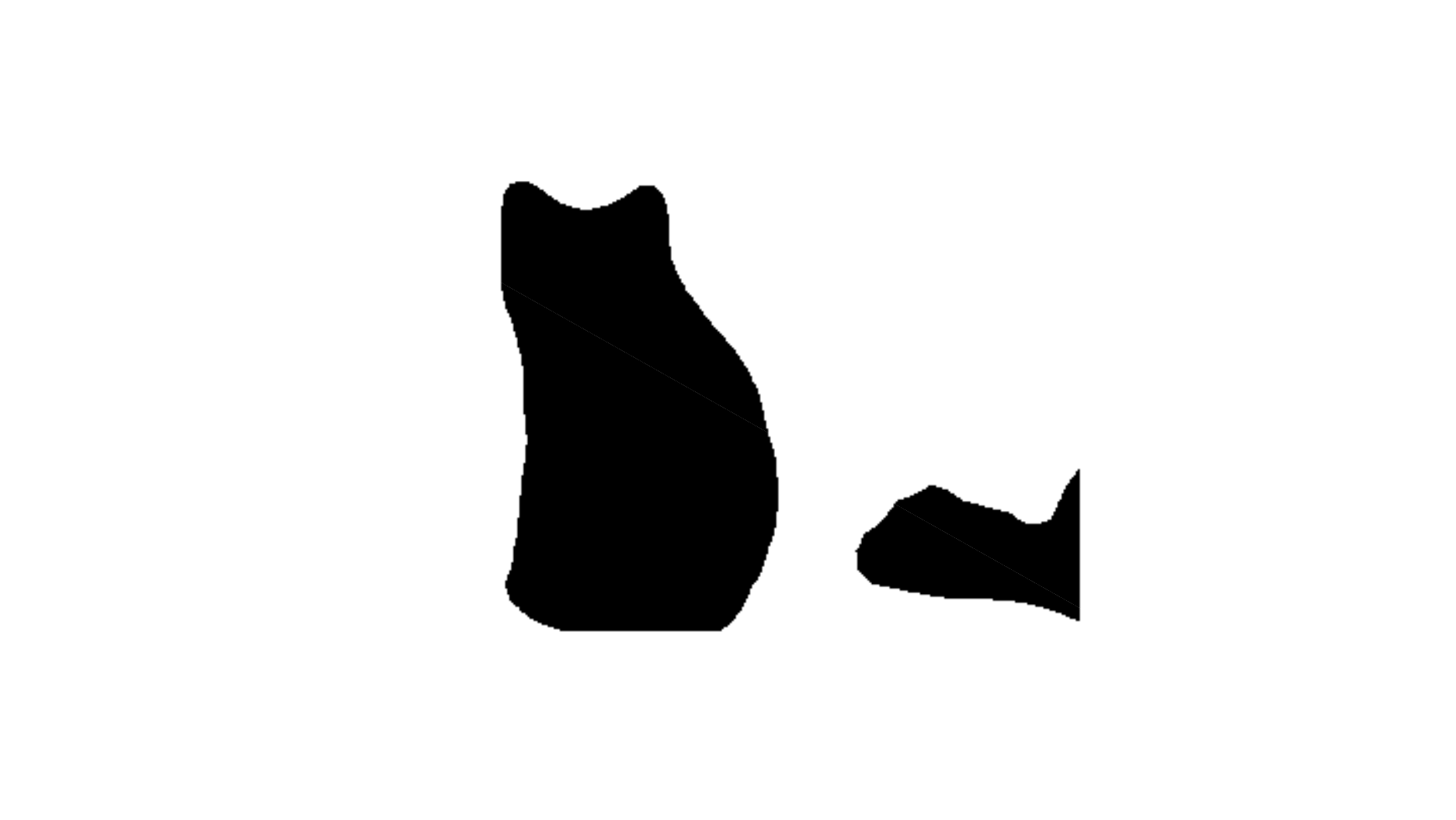}&
\includegraphics[height=5mm]{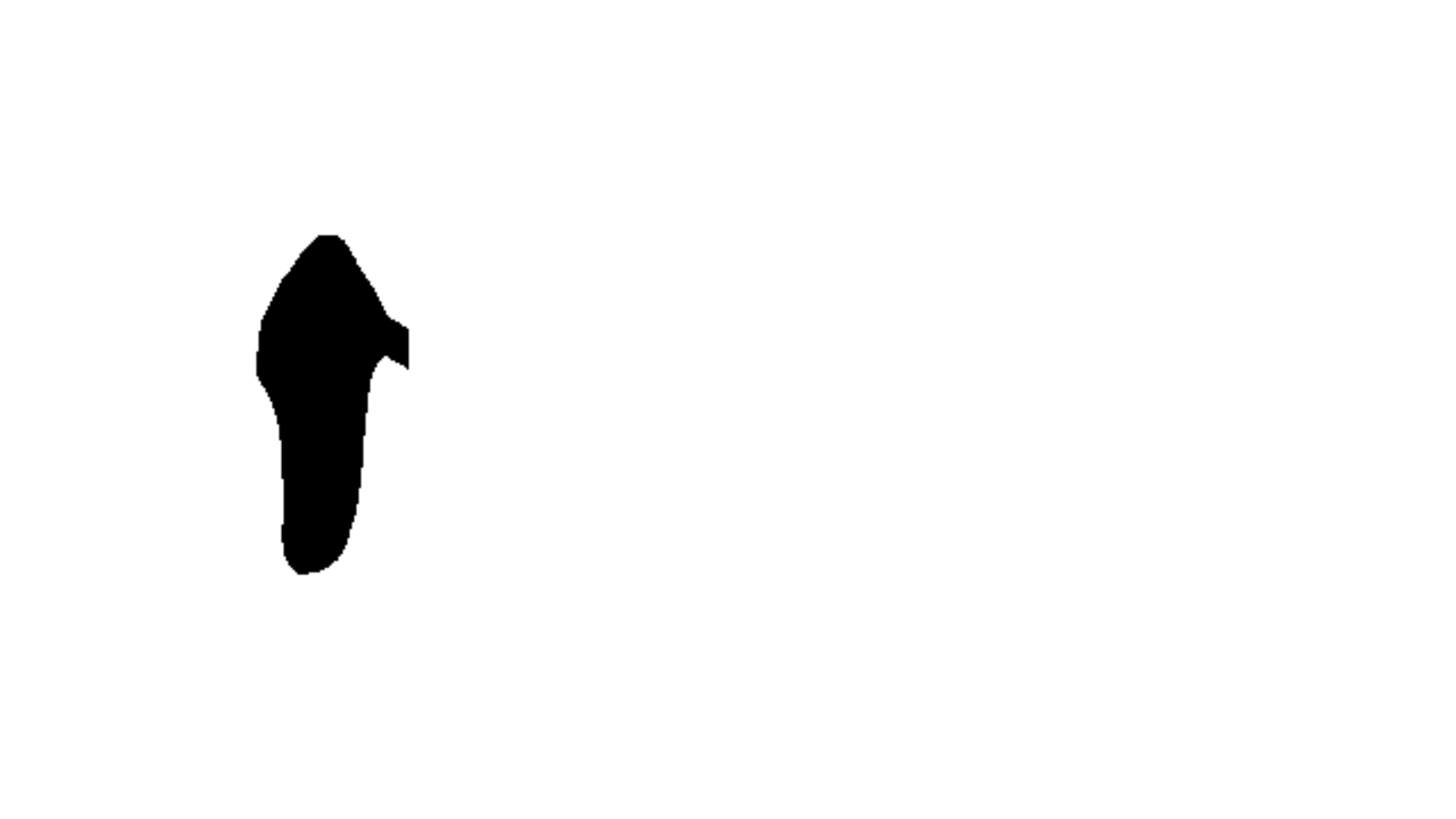}&
\includegraphics[height=5mm]{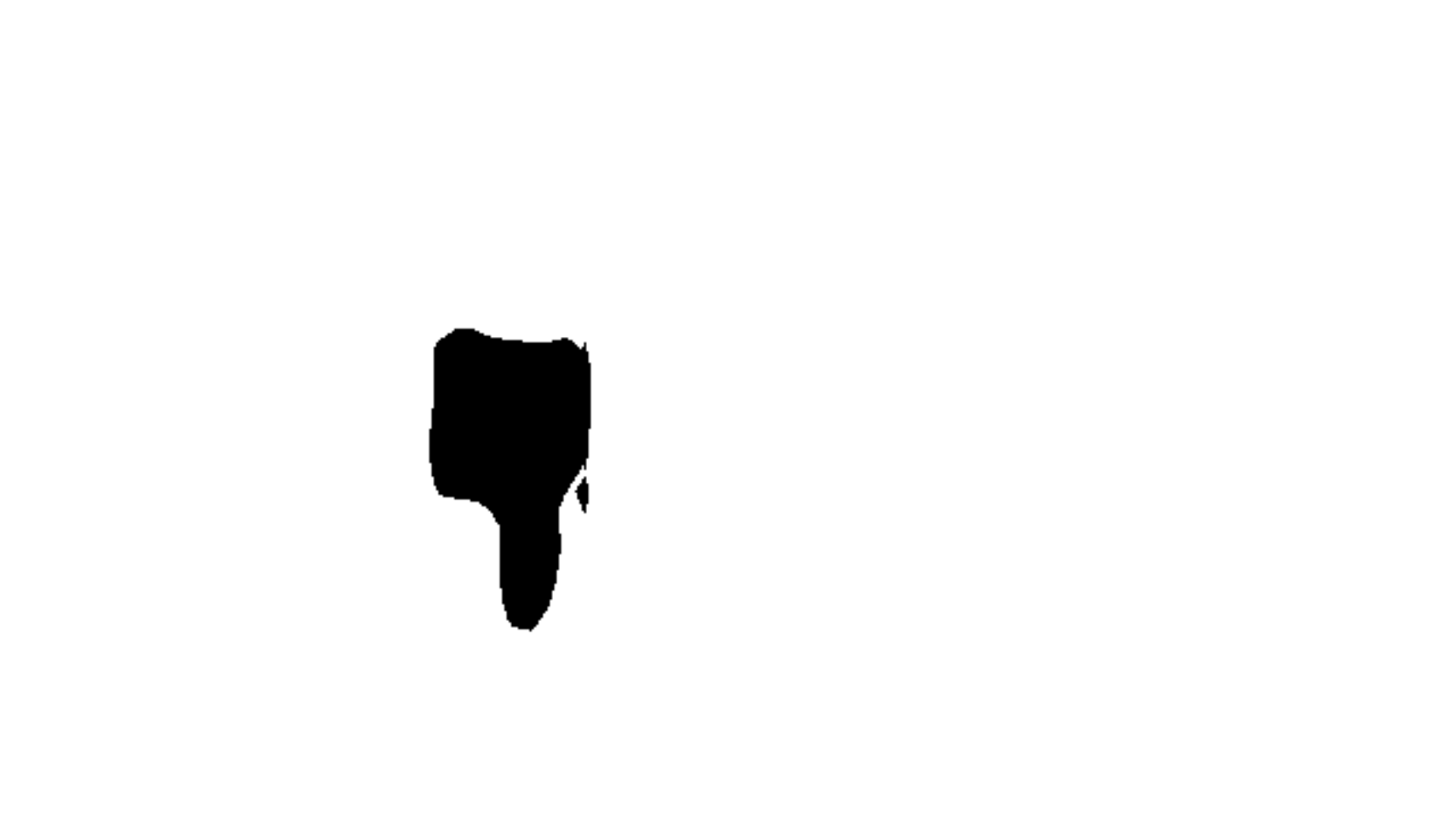}&
\includegraphics[height=5mm]{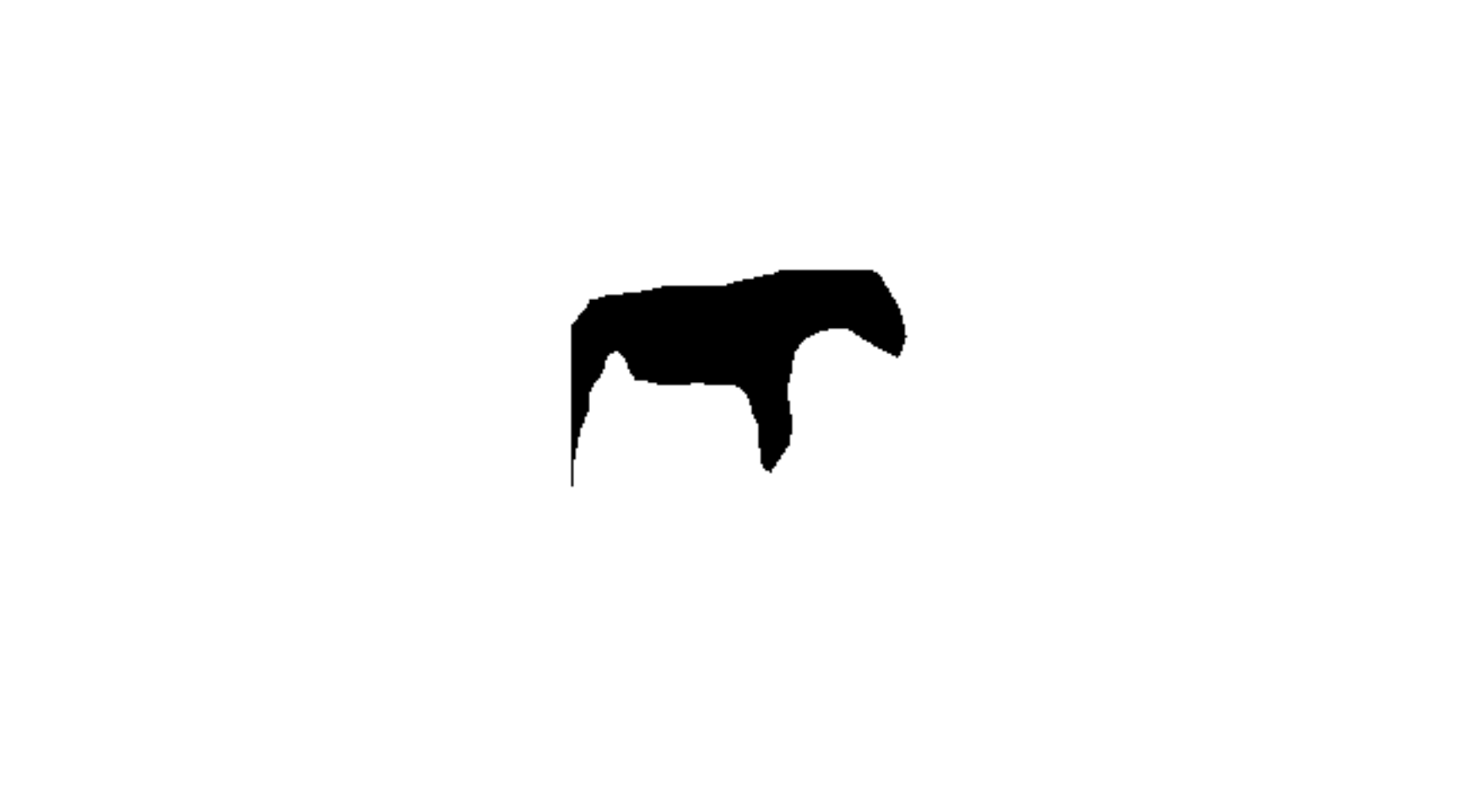}\\
\end{tabular}
\begin{tabular}{@{}c@{}c@{}c@{}c@{}}
\includegraphics[height=15mm]{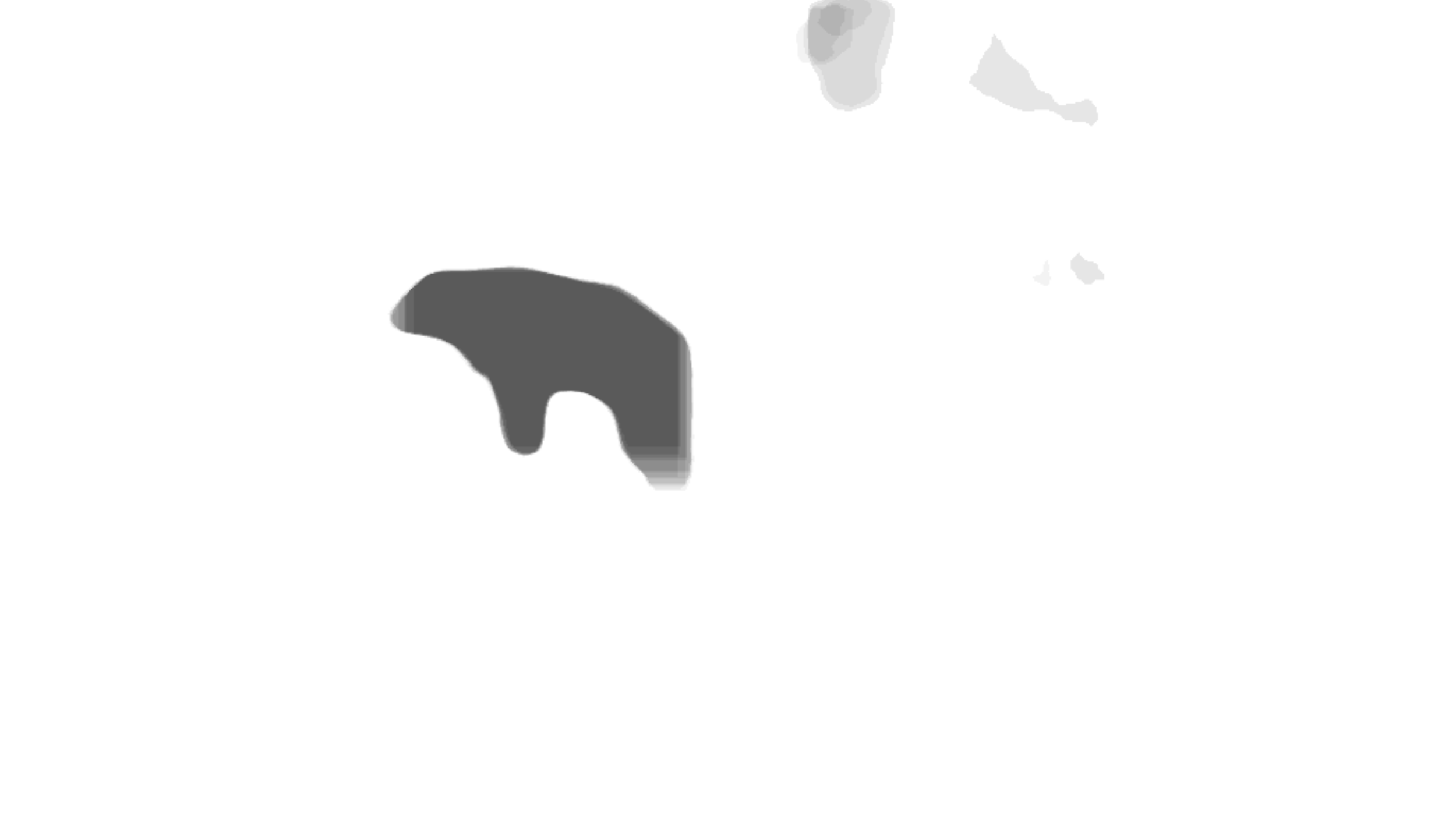}&
\includegraphics[height=15mm]{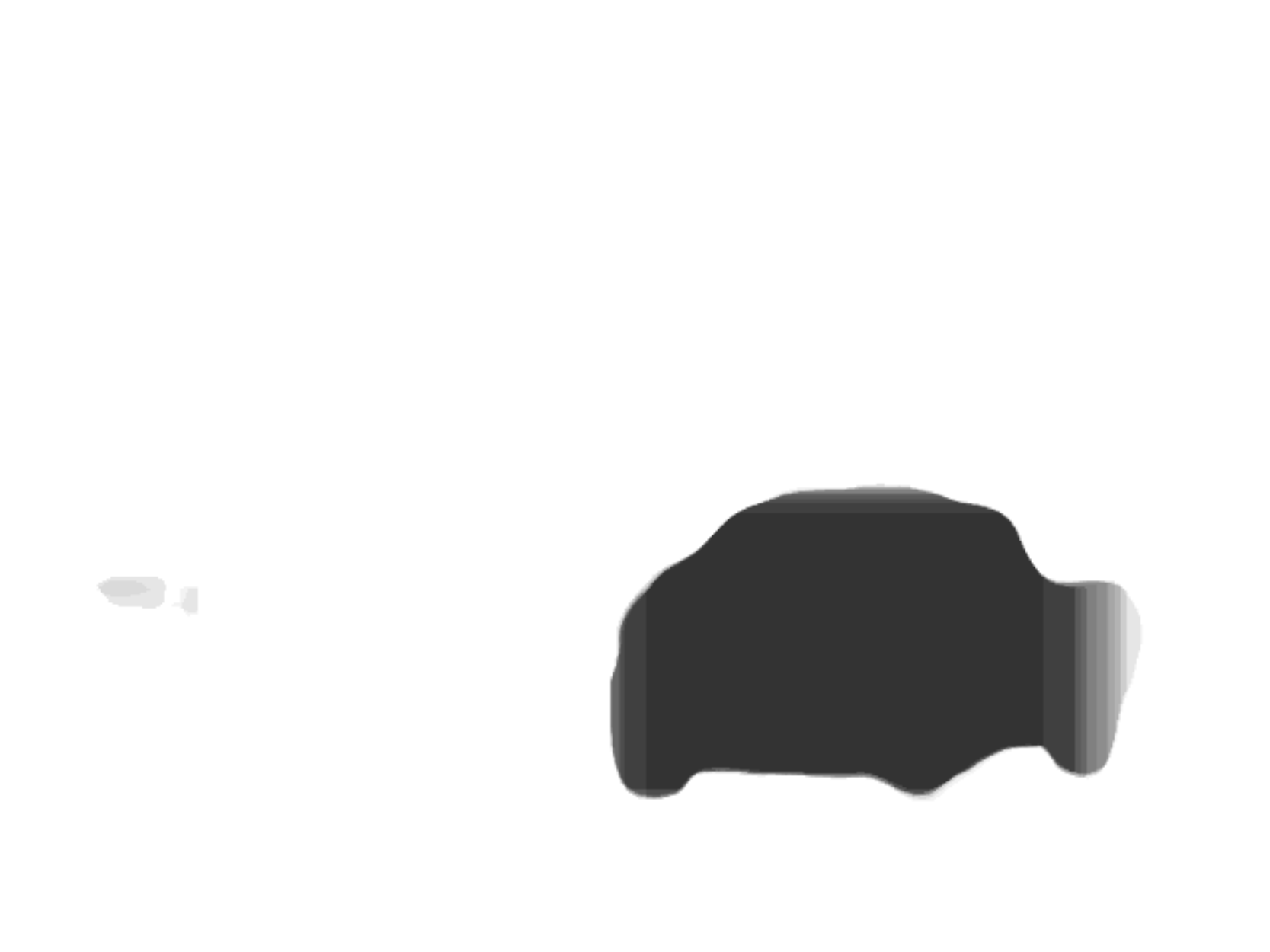}&
\includegraphics[height=15mm]{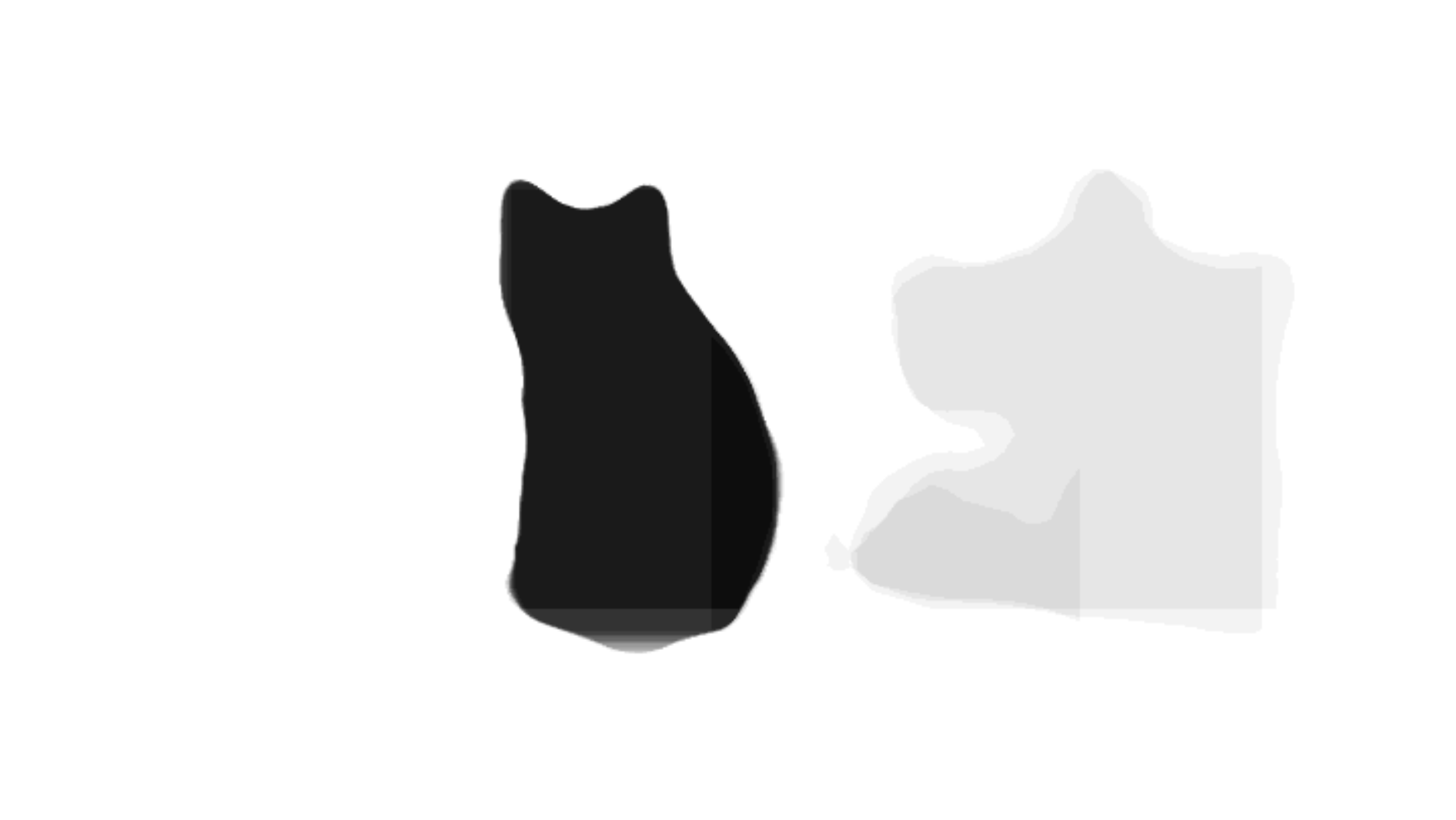}&
\includegraphics[height=15mm]{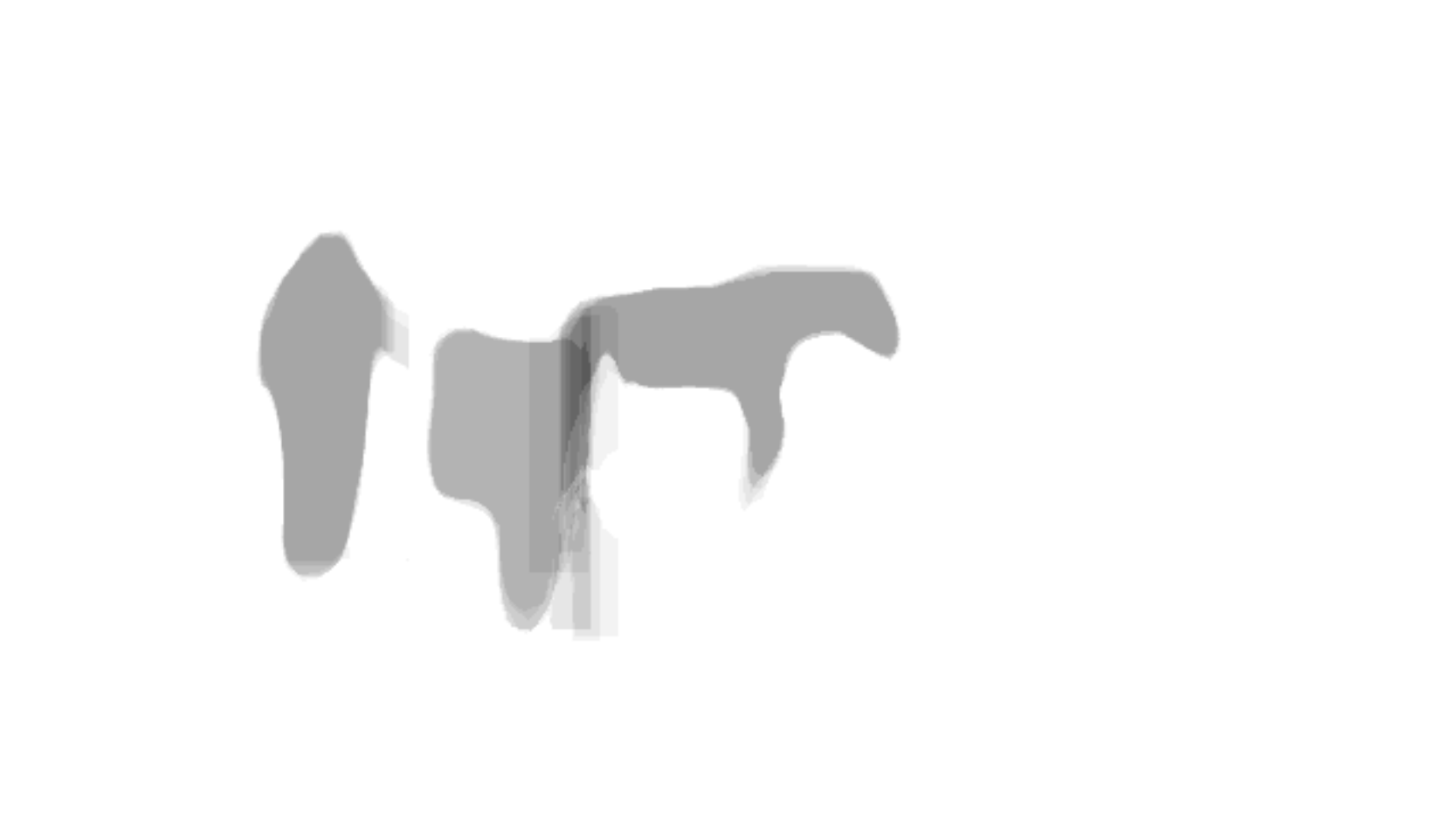}\\
\end{tabular}\vspace{-0.3cm}
\caption{\label{fig:LSDAexamples}Examples of the object detections and according segmentations. Top: LSDA detections on images from FBMS59 sequences \cite{Ochs14}. The first row shows the best 20 detections. The second row shows three exemplary selective search proposals and third row visualizes the average segmentation of all proposals. Bottom: The corresponding faster R-CNN detections. The first row shows the best 20 detections with a minimum detection score of 0.2. The second row shows three exemplary segmentations from deepLab \cite{chen14semantic,papandreou15weak} on these detections and third row visualizes the average segmentation.}
\vspace{-0.2cm}
\end{figure}

\myparagraph{Large Scale Detection through Adaptation}
The LSDA is a general object detector, trained to detect 7602 object categories \cite{Hoffman14Lsda}. In our experiments, we directly use the code and model deployed with their paper.
It operates on a set of object proposals, which is produced by selective search \cite{UijlingsIJCV2013}.
The selective search method operates on hierarchical segmentations, which means that we obtain a segmentation mask for each detection bounding box.
This segmentation provides a rough spatial and appearance estimation of the object of interest.

To better capture the moving objects in the video, we additionally generate selective search proposals from optical flow images and pass them to the LSDA framework. Example results for the detections and according frame-wise segmentations are given in Fig. \ref{fig:LSDAexamples}~(top).

\myparagraph{Faster R-CNN}
Faster R-CNN is an object detector that integrates a region proposal network with the Fast R-CNN \cite{girshickICCV15fastrcnn} network. It achieves state-of-the-art object detection accuracy on several benchmark datasets including PASCAL VOC 2012 and MS COCO with only 300 proposals per image \cite{renNIPS15fasterrcnn}. In our experiments, we directly used the code and model deployed with their paper. 

On the detections, we generate segmentation proposals using DeepLab \cite{chen14semantic,papandreou15weak}, again by directly using their implementation. Example results for the detections and according frame-wise segmentations are given in Fig. \ref{fig:LSDAexamples}~(bottom).
\begin{figure} [t]
\centering
\scalebox{0.85}{
\begin{tabular}{c@{\hspace{0.8cm}}c}
\includegraphics[height=0.34\linewidth]{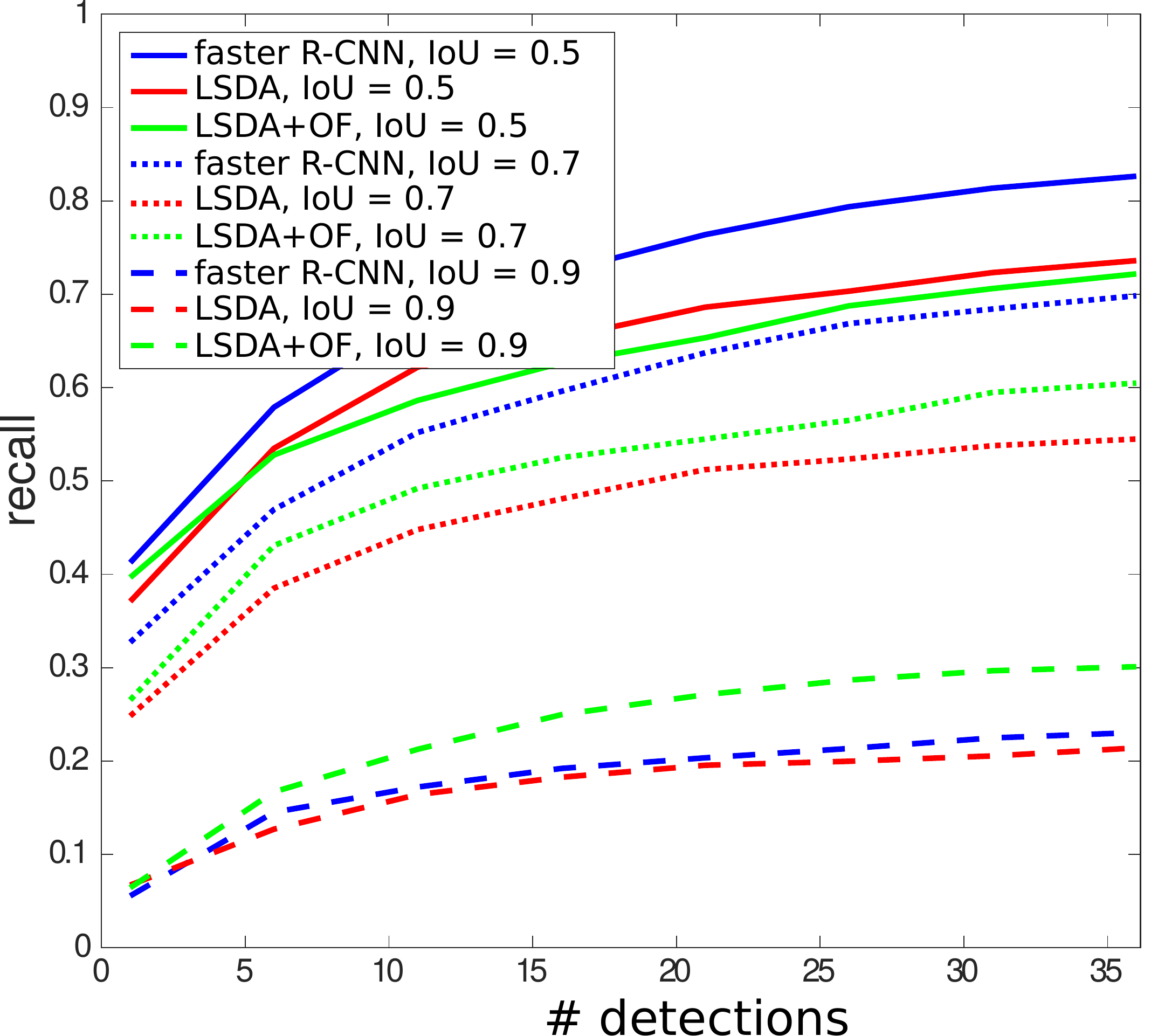}&
\includegraphics[height=0.34\linewidth]{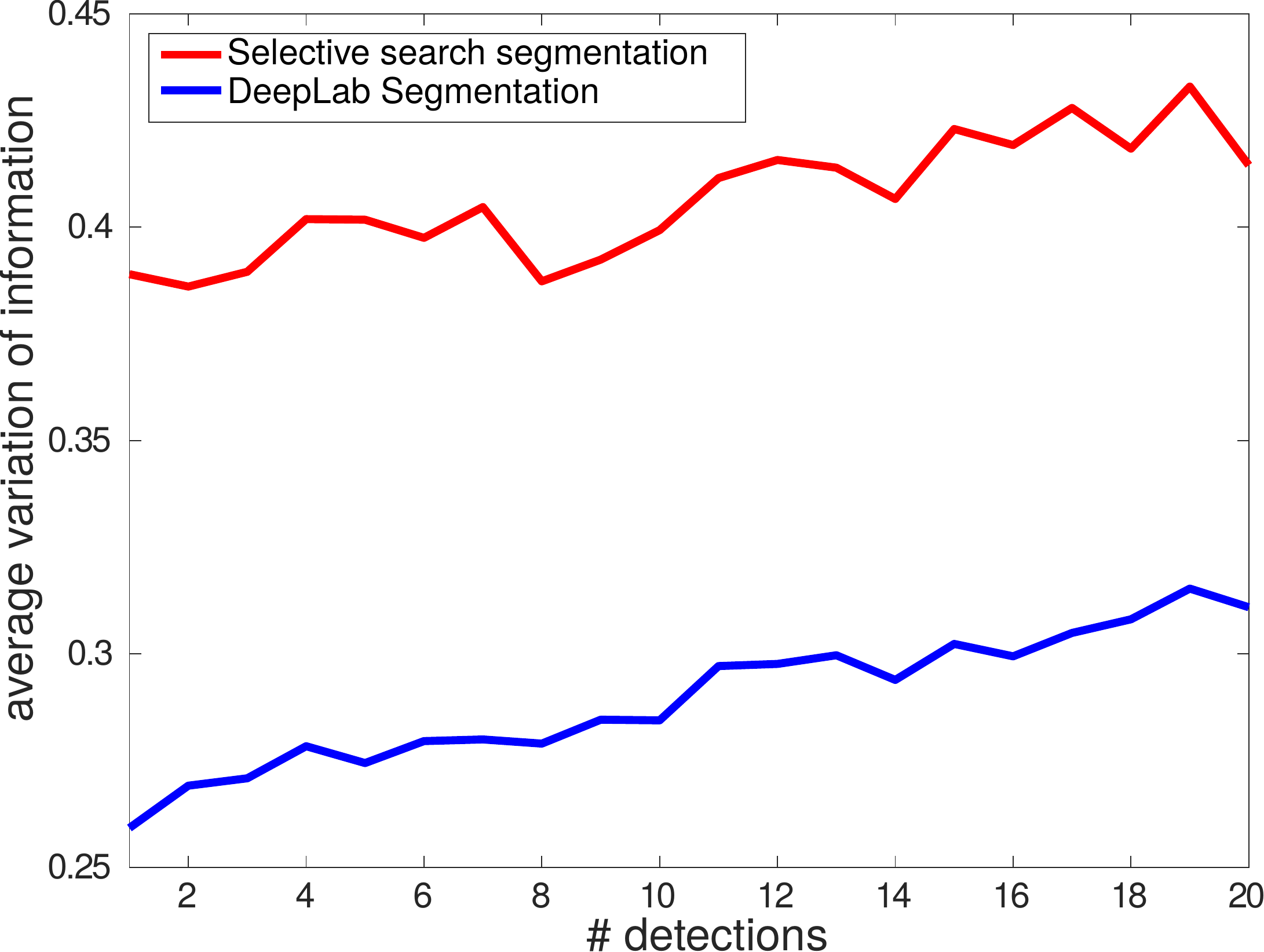}\\
\begin{minipage}{0.45\linewidth}(a) The detection performance of LSDA \cite{Hoffman14Lsda} and faster R-CNN \cite{renNIPS15fasterrcnn}. We compare the recall for three different IoU thresholds 0.9, 0.7, and 0.5.\end{minipage}&\begin{minipage}{0.45\linewidth}(b) The Variation of Information for the proposed object masks over the number of detections (lower is better). \phantom{............ ......... .......... .........}\end{minipage}
\end{tabular}}
 \caption{\label{fig:detectionEvaluation} Evaluation of the detection and segment proposals on the annotated frames of the FBMS59 \cite{Ochs14} training set.}
\end{figure}

\myparagraph{Evaluation}
Fig. \ref{fig:detectionEvaluation}(a) shows the achieved recall over the number of detections for LSDA \cite{Hoffman14Lsda} and faster R-CNN \cite{renNIPS15fasterrcnn} for different thresholds on the intersection over union (IoU) on the FBMS59 \cite{Ochs14} training set. For the higher thresholds, the performance of LSDA is improved when proposals from optical flow images are used (LSDA+OF) and for IoU $\geq0.9$, this setup yields best recall. However, for smaller IoU thresholds, faster R-CNN yields highest recall even without considering optical flow. The comparison of the segment mask proposals from selective search (for LSDA) and deepLab (for faster R-CNN) (Fig. \ref{fig:detectionEvaluation}~b(b)) shows the potential benefit of DeepLab. 
The visual comparison on the examples given in Fig. \ref{fig:LSDAexamples} shows that the selective search segmentation proposals selected by LSDA are more diverse than the DeepLab segmentations on the faster R-CNN detection. However, the overall localization quality is worse. We further evaluate detections from both methods in the Joint Multicut model. 

\myparagraph{Implementation Details}
In our graphical model, high-level nodes represent detections from either of the above described methods. For both detectors, we use the same setup. First, we select the most confident detections \footnote{above a threshold of 0.47 for LSDA and 0.97 for faster R-CNN - on a scale between 0 and 1.}. From those, we discard some detections according to the statistics of their respective segmentations. Especially masks from the selective search proposals sometimes only cover object outlines or leak to the image boundaries. Thus, if such a mask covers less than 20\% of its bounding box or more than 60\% of the whole image area, the respective detections are not used as nodes in our graph.

The pairwise terms between detections are computed from the IoU and the normalized distances $d^{\text{sp}}$ of their bounding boxes
\begin{align}
d^{\text{sp}}(v_i^{\text{high}},v_j^{\text{high}})=2\left\|\begin{pmatrix}
\frac{x_{v_i^{\text{high}}} - x_{v_j^{\text{high}}}}{w_{v_i^{\text{high}}} + w_{v_j^{\text{high}}}}\\\frac{y_{v_i^{\text{high}}} - y_{v_j^{\text{high}}}}{h_{v_i^{\text{high}}} + h_{v_j^{\text{high}}}}\end{pmatrix}\right\|\nonumber,
\end{align}
where $\text{pos}_{v^{\text{high}}_i}$, $w_{v^{\text{high}}_i}$, and $h_{v^{\text{high}}_i}$ are defined as in Eq.~\eqref{eq:disthl}. For all pairs of detections within one frame and in neighboring frames, the pseudo cut probability is computed as
\begin{align}
p_{e^{\text{high}}_{ij}}=
    \begin{cases}
      1-\frac{1}{1+\text{exp}(20*(0.7-\text{IoU}(v_i^{\text{high}},v_j^{\text{high}})))}, & \text{if}\ \text{IoU}(v_i^{\text{high}},v_j^{\text{high}})>0.7\\
      \frac{1}{1+\text{exp}(5*(1.2-d^{\text{sp}}(v_i^{\text{high}},v_j^{\text{high}})))}, & \text{if}\ d^{\text{sp}}(v_i^{\text{high}},v_j^{\text{high}})>1.2\\
      0.5, & \text{otherwise}
    \end{cases}
\label{eq:p_hl} 
\end{align} 
The parameters have been set such as to produce reasonable results on the FBMS59 training set. Admittedly, parameter optimization on the training set might further improve our results.

The pairwise terms $c_{e^{\text{hl}}}$ are computed from $p_{e^{\text{hl}}}$ as defined in Eq.~\eqref{eq:p_hl} with $\sigma=2$. This large threshold accounts for the uncertainty in the bounding box localizations.

\newcommand{\mMoSegS}{SC\cite{Ochs14} sparse\xspace}
\begin{table*}[tb]
\centering
\small
\newcommand{\hs}{\hspace{1mm}}
\scalebox{0.88}{
\begin{tabular}{@{} l | @{}   c    c   c    c | @{\hspace{0.1cm}} c  c  c  c @{}}
\multicolumn{1}{c}{} & \multicolumn{4}{c}{\textbf{Training set} (29 sequences)} & \multicolumn{4}{c}{\textbf{Test set} (30 sequences)} \\[0.5mm]
\hline
                      &  \small{P}  & \small{R}  &\small{F}  & \small{O}
                                                   &\small{P} & \small{R} & \small{F} & \small{O} \\
\hline\hline
 \small{SC \cite{Ochs14}}& \small{85.10\%} & \small{62.40\%} & \small{72.0\%}  & \small{17/65} &   \small{79.61\%} & \small{60.91\%} & \small{69.02\%} & \small{24/69}  \\
\small{SC + Higher Order \cite{Ochs12}}& \small{81.55\%} & \small{59.33\%} & \small{68.68\%}  & \small{16/65} &   \small{82.11\%} & \small{64.67\%} & \small{72.35\%} & \small{27/69}
\\
\small{MCe \cite{keuper15a}} &  \small{\bf 86.73\%} & \small{73.08\%} & \small{79.32\%}  & \small{\bf 31/65} & \small{\bf 87.88\%} & \small{67.7\%} & \small{76.48\%} & \small{25/69} \\
\hline
\small{MCe + det. (LSDA)} &  \small{86.43\%} & \small{75.79\%} & \small{80.7617\%}  & \small{31/65} &   \small{-} & \small{-} & \small{-} & \small{-} \\
\small{JointMulticut (LSDA)} &  \small{86.43\%} & \small{75.79\%} & \small{80.7634\%}  & \small{31/65} &  \small{  87.46\%} & \small{70.80\%} & \small{78.25\%} & \small{ 29/69} \\
\hline
\small{MCe + det. (f. R-CNN)} &  \small{83.46\%} & \small{79.46\%} & \small{ 81.41\%}  & \small{35/65} &    \small{-} & \small{-} & \small{-} & \small{-} \\
\small{JointMulticut (f. R-CNN)} &\small{84.85\%} & \small{\bf 80.17\%} & \small{\bf 82.44\%}  & \small{\bf 35/65}  &   \small{84.52\%} & \small{\bf 77.36\%} & \small{\bf 80.78\%} & \small{\bf 35/69} \\
\hline
\end{tabular}
}
\caption{\label{tab:msevaluation} Results on the FBMS-59 dataset on training (left) and test set (right). We report \textbf{P}: average precision, \textbf{R}: average recall, \textbf{F}: F-measure and \textbf{O}: extracted objects with $\text{F}\geq 75\%$. All results are computed for sparse trajectory sampling at 8 pixel distance.}
\end{table*}
\begin{figure*}[ht!]
\centering
\scalebox{1}{
\centering
\begin{tabular}{@{}l@{\hspace{0.01cm}}l@{\hspace{0.01cm}}l@{\hspace{0.01cm}}l@{\hspace{0.15cm}}l@{\hspace{0.01cm}}l@{\hspace{0.01cm}}l@{}}
&frame 80&100&120\hspace{0.8cm}\dots&540&560&580 \\
&\includegraphics[width=0.16\linewidth]{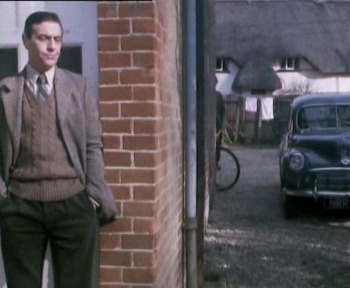}&
\includegraphics[width=0.16\linewidth]{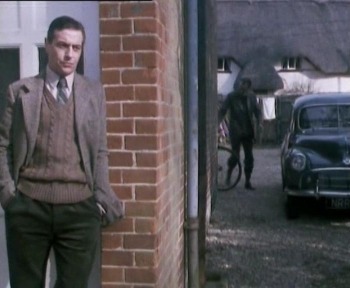}&
\includegraphics[width=0.16\linewidth]{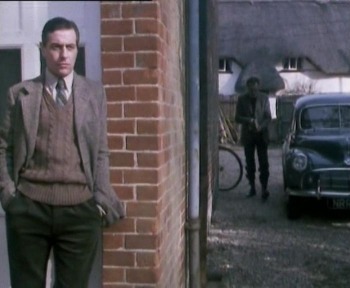}&
\includegraphics[width=0.16\linewidth]{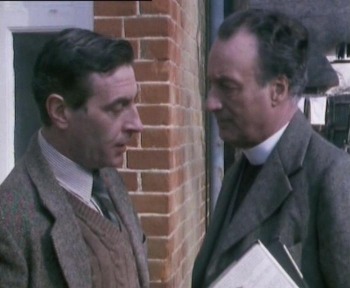}&
\includegraphics[width=0.16\linewidth]{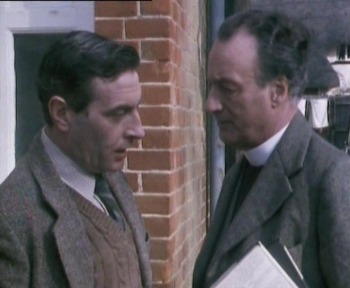}&
\includegraphics[width=0.16\linewidth]{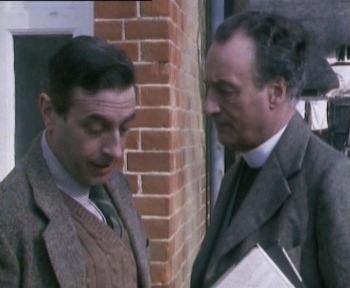}\\
\rotatebox{90}{MCe\cite{keuper15a}}&
\includegraphics[width=0.16\linewidth]{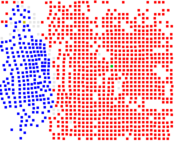}&
\includegraphics[width=0.16\linewidth]{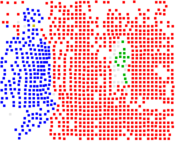}&
\includegraphics[width=0.16\linewidth]{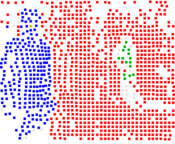}&
\includegraphics[width=0.16\linewidth]{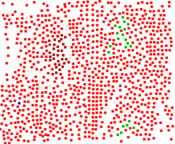}&
\includegraphics[width=0.16\linewidth]{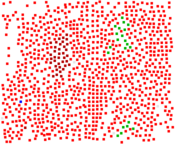}&
\includegraphics[width=0.16\linewidth]{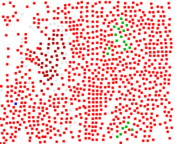}\\
&
\includegraphics[width=0.16\linewidth]{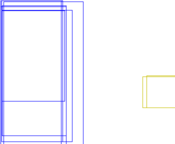}&
\includegraphics[width=0.16\linewidth]{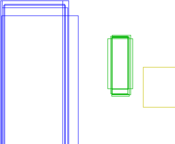}&
\includegraphics[width=0.16\linewidth]{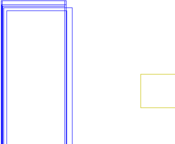}&
\includegraphics[width=0.16\linewidth]{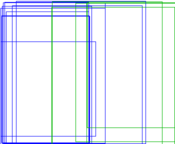}&
\includegraphics[width=0.16\linewidth]{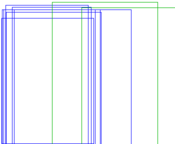}&
\includegraphics[width=0.16\linewidth]{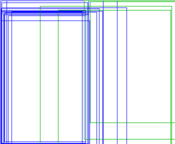}\\
\rotatebox{90}{\hspace{0.2cm}Joint MC}&\includegraphics[width=0.16\linewidth]{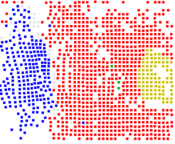}&
\includegraphics[width=0.16\linewidth]{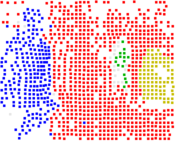}&
\includegraphics[width=0.16\linewidth]{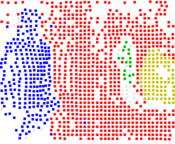}&
\includegraphics[width=0.16\linewidth]{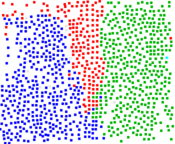}&
\includegraphics[width=0.16\linewidth]{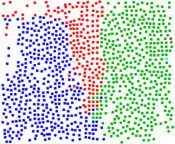}&
\includegraphics[width=0.16\linewidth]{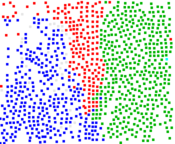}\\
\end{tabular}
}
\caption{Comparison of the proposed Joint Multicut model and the multicut on trajectories (MCe) \cite{keuper15a} on the \emph{marple6} sequence of FBMS59. While with MCe the segmentation breaks between the shown frames, the tracking information from the bounding box subgraph helps our joint model to segment the two men throughout the sequence. Additionally, static, consistently detected objects like the car in the first part of the sequence are segmented as well. As these are not annotated, this causes over-segmentation on the FBMS59 benchmark evaluation.}
\label{fig:marple6}
\end{figure*}
\myparagraph{Results}
Our results are given in Tab. \ref{tab:msevaluation}. The motion segmentation considering only the trajectory information from \cite{keuper15a} performs already well on the FBMS59 benchmark. However, the Joint Multicut model improves over the previous state of the art for both types of object detectors. Note that not only the baseline method of \cite{keuper15a} is outperformed with quite a margin on the test set - also the motion segmentation based on higher-order potentials \cite{Ochs12} can not compete with the proposed joint model.

To assess the impact of the joint model components, we evaluate not only the full model but also its performance if pairwise terms between detection nodes are omitted (denoted by MCe + detections). For LSDA detections, this result is pretty close to the Joint Multicut model, implying that the pairwise information we currently employ between the bounding boxes is quite weak. However, for the better localized faster R-CNN detections, the high-level pairwise terms contribute significantly to the overall performance of the joint model. 

\begin{figure*}[t!]
\centering
\begin{tabular}{@{}l@{\hspace{0.01cm}}l@{\hspace{0.01cm}}l@{\hspace{0.01cm}}l@{\hspace{0.01cm}}l@{\hspace{0.01cm}}l@{}}
&frame 20&40&60&80&100 \\
&\includegraphics[width=0.19\linewidth]{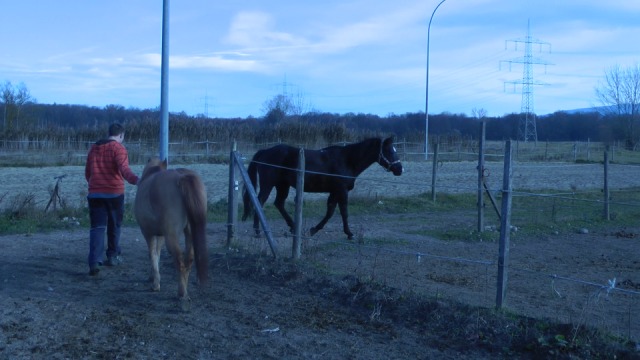}
&\includegraphics[width=0.19\linewidth]{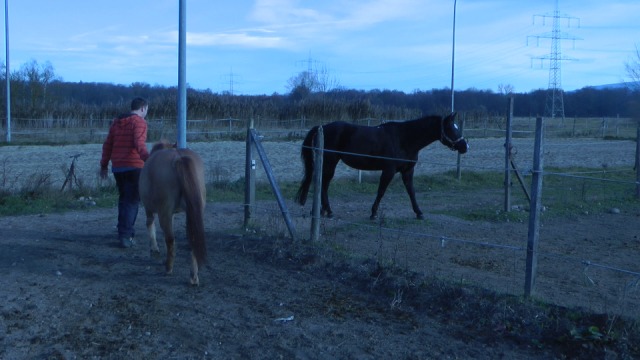}
&\includegraphics[width=0.19\linewidth]{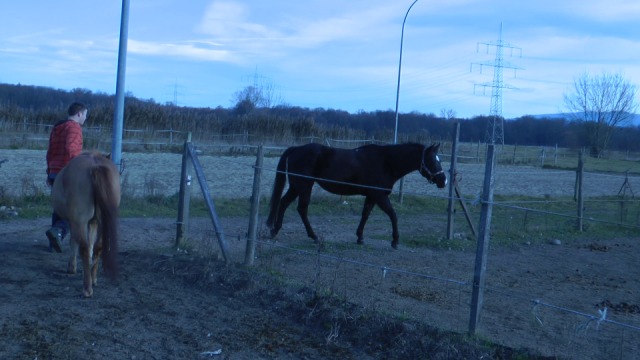}
&\includegraphics[width=0.19\linewidth]{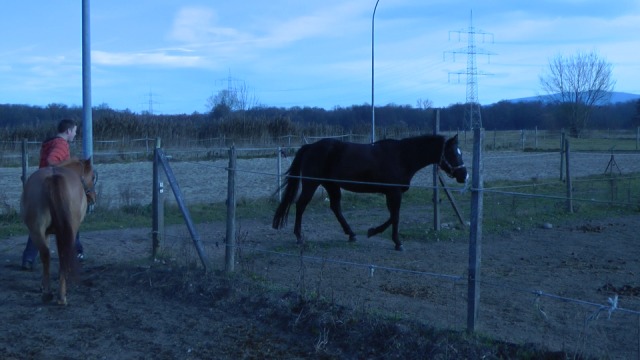}
&\includegraphics[width=0.19\linewidth]{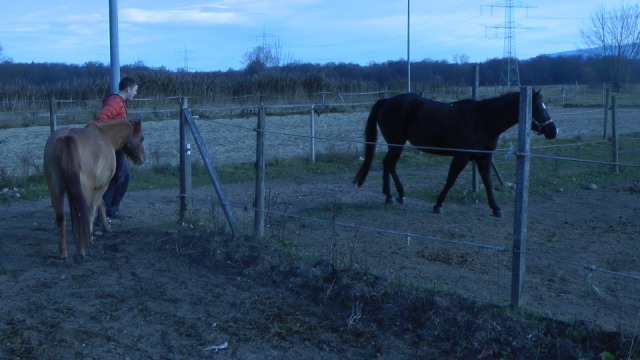}\\
\rotatebox{90}{MCe\cite{keuper15a}}&
\includegraphics[width=0.19\linewidth]{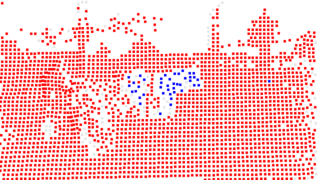}
&\includegraphics[width=0.19\linewidth]{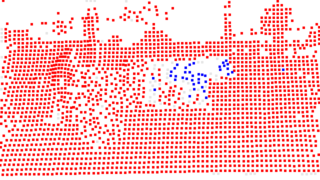}
&\includegraphics[width=0.19\linewidth]{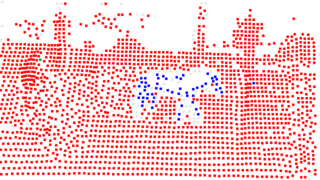}
&\includegraphics[width=0.19\linewidth]{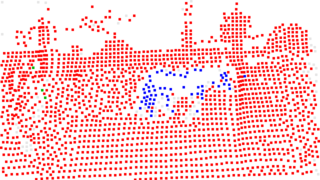}
&\includegraphics[width=0.19\linewidth]{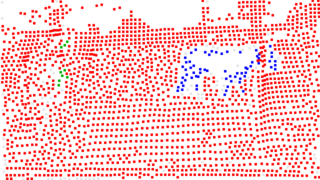}
\\
&
\includegraphics[width=0.19\linewidth]{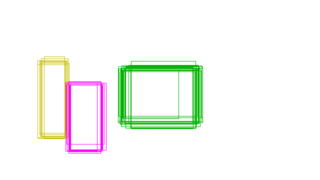}&
\includegraphics[width=0.19\linewidth]{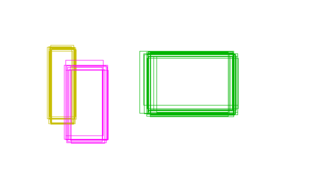}&
\includegraphics[width=0.19\linewidth]{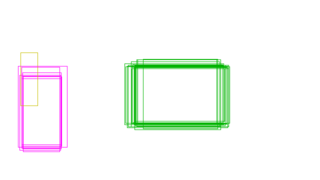}&
\includegraphics[width=0.19\linewidth]{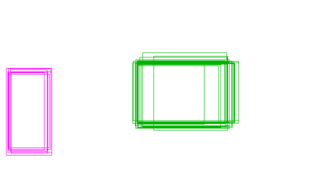}&
\includegraphics[width=0.19\linewidth]{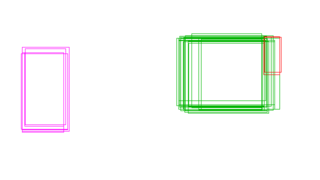}\\
\rotatebox{90}{Joint MC}&\includegraphics[width=0.19\linewidth]{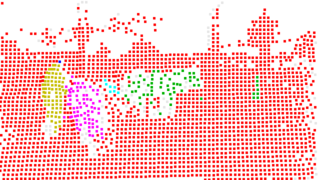}
&\includegraphics[width=0.19\linewidth]{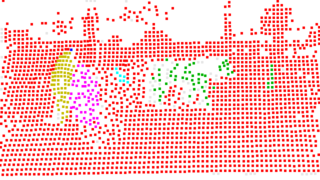}
&\includegraphics[width=0.19\linewidth]{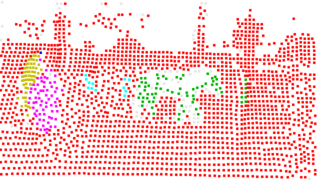}
&\includegraphics[width=0.19\linewidth]{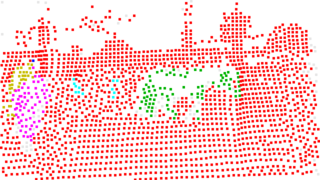}
&\includegraphics[width=0.19\linewidth]{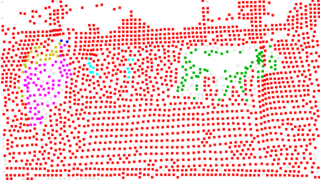}\\
\end{tabular}
\caption{Comparison of the proposed Joint Multicut model and the multicut on trajectories (MCe) \cite{keuper15a} on the \emph{horses06} sequence of FBMS59.}
\label{fig:horses}
\end{figure*}

Qualitative examples of the motion segmentation and object tracking results using the faster R-CNN detections are given in Fig. \ref{fig:marple6} and \ref{fig:horses}. Due to the detection information and the repulsive terms between those object detections and point trajectories not passing through them, static objects like the car in the \emph{marple6} sequence (yellow cluster) can be segmented. The man approaching the camera in the same sequence can be tracked and segmented (green cluster) throughout the sequence despite the scaling motion. Similarly, in the \emph{horses} sequence, all three moving objects can be tracked and segmented through strong partial occlusions.

Since the ground truth annotations are sparse and only contain moving objects, this dataset was not used to quantitatively evaluate the multi-target tracking performance.

\subsection{Multi-Target Tracking Data}
First, we evaluate the tracking performance of our Joint Multicut model on the publicly available sequences: TUD-Campus, TUD-Crossing \cite{Andriluka:2008:PTD} and ParkingLot \cite{Zamir:2012:GMC}. 
These sequences have also been used to evaluate the Subgraph Multicut method \cite{tang15} and therefore allow for direct comparison to the only previous multicut approach to multi-target tracking. 

To assess the quality of the motion segmentation of the joint approach, we annotated the sequences TUD-Campus and ParkingLot with ground truth segmentations of all pedestrians in every 20th frame. For the TUD-Crossing sequence, such annotations have been previously published by \cite{leibe11}. 

\myparagraph{Implementation Details}
To allow for direct comparison to \cite{tang15}, we compute all high-level information, i.e.\ the detection nodes $v^{\text{high}}\in V^{\text{high}}$, edges $e^{\text{high}}\in E^{\text{high}}$, and their costs $c_{e^{\text{high}}}$  exactly as reported in \cite{tang15} with only one difference: the Subgraph Multicut models from \cite{tang15} employs not only pairwise but also unary terms which our proposed Joint Multicut model does not require. We omit these terms.

In \cite{tang15}, DPM-based person detections \cite{Felzenszwalb2010PAMI} are used. To add robustness and enable the computation of more specific pairwise terms, these detections are grouped to small, overlapping tracklets of length 5 as in \cite{Andriluka:2008:PTD} without applying any Non-Maximum Suppression. Since tracklets are computed in every frame, the same detections can be part of several (at most 5) tracklets.  

Pairwise terms between the tracklets are computed from temporal distances, normalized scale differences, speed, spatio-temporal locations and dColorSIFT features \cite{zhao2013unsupervised}, combined non-lineraly as in \cite{tang15}.

\begin{figure} [t]
\centering
\begin{tabular}{llrr}
\begin{minipage}{0.25\textwidth}(a) Mean pedestrian shape template\hfill\end{minipage}
&\begin{minipage}{0.15\textwidth}\includegraphics[width=0.5\linewidth]{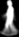}\end{minipage}&
\begin{minipage}{0.35\textwidth}(b) Trajectory-Tracklet edges:\\
For every pair of trajectories and tracklets, an edge is inserted if the trajectory either hits a bounding box template or passes sufficiently far outside the bounding box.
\hfill \end{minipage}
\begin{minipage}{0.25\textwidth}\includegraphics[width=0.8\linewidth]{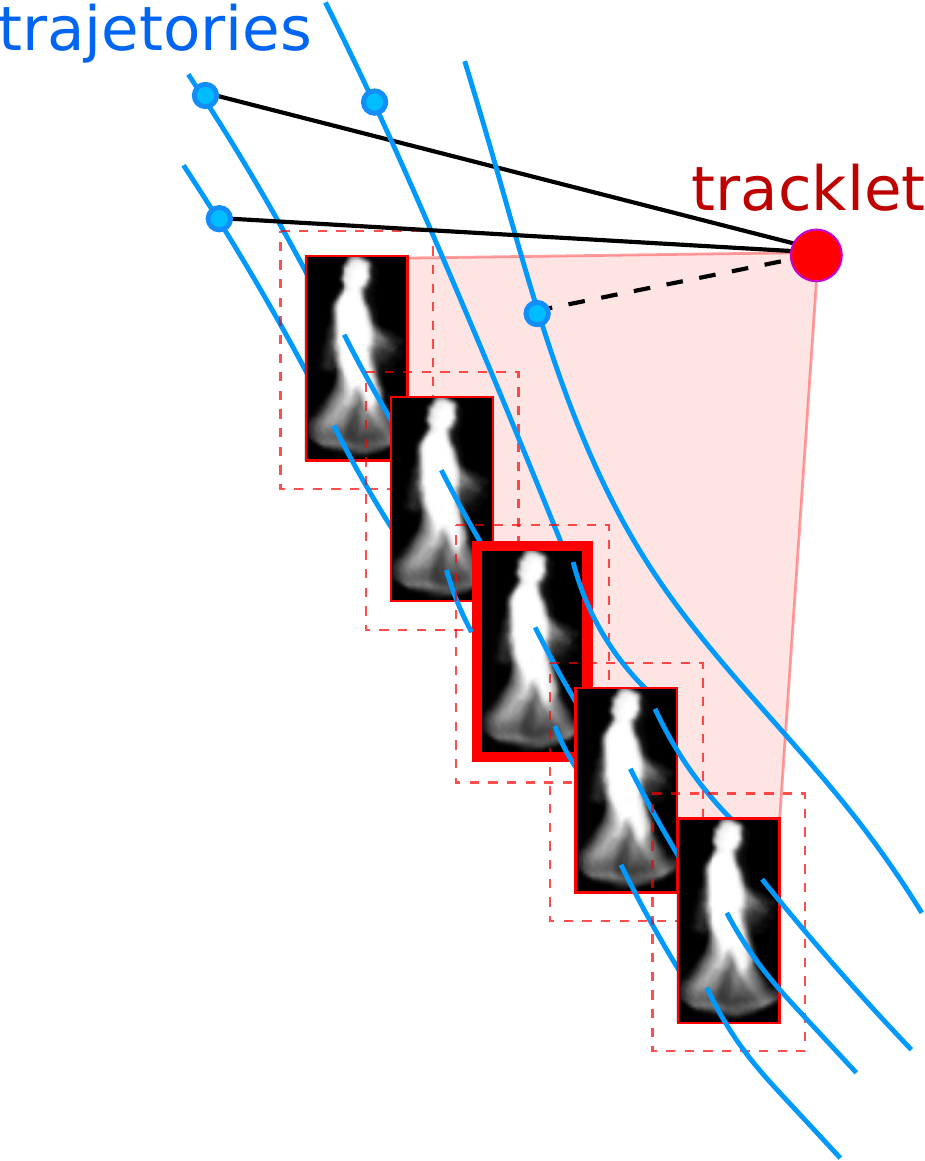}\end{minipage}
\end{tabular}\vspace{-1.2em}
 \caption{\label{fig:pedGraph} The average pedestrian shape template and the trajectory-tracklet edges.}
\vspace{-1.2em}
\end{figure}

To compute pairwise terms $c_{e^{\text{hl}}_{ij}}$ between trajectory and tracklet nodes as described in Sec.~\ref{sec:hlpairwise}, we compute the average pedestrian shape from the shape prior training data provided in \cite{Cremers-et-al-cvpr08} (see Fig.~\ref{fig:pedGraph}~(a)). For every detection $\text{bbx}_k$, $T_{\text{bbx}_k}$ denotes the pedestrian template shifted and scaled to the $k$th bounding box position and size. The tracklet information allows to determine the walking direction of the pedestrian, such that the template can be flipped accordingly. For every detection $\text{bbx}_k$ with $k=\{1,\dots,5\}$ of a tracklet $v^{\text{high}}_i$, the cut propability $p(\text{bbx}_k,v^{\text{low}}_j)$ to a trajectory node $v^{\text{low}}_j$ is computed according to Eq.~\eqref{eq:p_hl} with $\sigma=1.2$. 

A trajectory node $v^{\text{low}}_j$ is linked to a tracklet node $v^{\text{high}}_i$ coexisting in a common frame with an edge cost 
\begin{align}
\label{pedtrbbx}
c_{e_{ij}^{hl}} =  \sum_{k=1}^{5}\text{logit}(p(\text{bbx}_k,v^{\text{low}}_j)).
\end{align} 
Fig.~\ref{fig:pedGraph}~(b) visualizes the edges between tracklets and point trajectories.

\myparagraph{Evaluation Metrics}
The pedestrian motion segmentation is evaluated with the metrics precision (P), recall (R), f-measure(F) and number of retrieved objects (O) as proposed for the FBMS59 motion segmentation benchmark \cite{Ochs14}.
To evaluate the tracking performance, we use standard CLEAR MOT as evaluation metrics. Additionally, we report mostly tracked (MT), partly tracked (PT), mostly lost (ML) and fragmentation (FM).
\begin{table*}[tb]
\centering
\small
\newcommand{\hs}{\hspace{1mm}}
\scalebox{0.80}{
\begin{tabular}{@{} l  @{\hspace{0.12cm}} c    c    c   c    c }
                     &  \small{D} &  \small{P}  & \small{R}  &\small{F}  & \small{O}\\
\hline\hline
TUD-Campus &&&&&\\
\hline
\small{SC \cite{Ochs14}}     &  \small{0.80\%} & \small{71.67\%} & \small{47.28\%} & \small{56.97\%}   & \small{2/8} \\
\small{MCe \cite{keuper15a}} &  \small{0.80\%} & \small{58.94\%} & \small{59.68\%} & \small{59.31\%} & \small{3/8} \\
\hline
\small{MCe + det.} &  \small{0.80\%} & \small{ 73.93\%} & \small{ 58.67\%} & \small{65.43\%}  & \small{ 3/8} \\
\small{Tracklet MC + traj.} &  \small{0.80\%} & \small{\bf 85.15\%} & \small{\bf 66.40\%} & \small{\bf 74.61\%}  & \small{\bf 5/8}  \\
\small{JointMulticut} &  \small{0.80\%} & \small{75.67\%} & \small{61.97\%} & \small{68.13\%}  & \small{ 3/8}  \\
\hline\hline
TUD-Crossing&&&&&\\
\hline
\small{SC \cite{Ochs14}}&  \small{0.85\%} & \small{\bf 67.92\%} & \small{20.16\%} & \small{31.09\%}  & \small{0/15} \\
\small{MCe \cite{keuper15a}} &  \small{0.85\%} &  \small{43.78\%} & \small{38.53\%}  &\small{40.99\%} & \small{1/15}\\
\hline
\small{MCe + det.} &  \small{0.85\%} & \small{67.34\%} & \small{57.33\%} & \small{61.93\%}  & \small{3/15} \\
\small{Tracklet MC + traj.} &  \small{0.85\%} & \small{63.10\%} & \small{57.45\%} & \small{60.14\%}  & \small{ 5/15}  \\
\small{JointMulticut} &  \small{0.85\%} & \small{62.10\%} & \small{\bf 64.65\%} & \small{\bf 63.35\%}  & \small{\bf 8/15}  \\
\hline\hline
ParkingLot&&&&&\\
\hline
\small{SC \cite{Ochs14}}&  \small{1.01\%} & \small{ 77.06\%} & \small{14.79\%} & \small{24.81\%}  & \small{0/15} \\
\small{MCe \cite{keuper15a}} &  \small{1.01\%} & \small{58.97\%} & \small{18.14\%} & \small{27.75\%}  & \small{ 0/15}\\
\hline
\small{MCe + det.} &  \small{1.01\%} & \small{74.51\%} & \small{63.52\%} & \small{ 68.47\%}  & \small{ 5/15} \\
\small{Tracklet MC + traj.} &  \small{1.01\%} & \small{\bf 77.54\%} & \small{53.16\%} & \small{63.01\%}  & \small{4/15}  \\
\small{JointMulticut} &  \small{1.01\%} & \small{72.62\%} & \small{\bf 66.93\%} & \small{\bf 69.66\%}  & \small{\bf 7/15}  \\
\hline
\end{tabular}
}
\caption{\label{tab:msevaluationTracking} Motion Segmentation on the Multi-Target Tracking sequences. We report \textbf{D}: average region density, \textbf{P}: average precision, \textbf{R}: average recall, \textbf{F}: F-measure and \textbf{O}: extracted objects with $\text{F}\geq 75\%$. All results are computed for sparse trajectory sampling at 8 pixel distance.}
\end{table*}

\myparagraph{Results}
The evaluation of the motion segmentations on these three pedestrian tracking sequences produced by the Joint Multicut model is given in Tab. \ref{tab:msevaluationTracking}. To assess the importance of the model parts, we not only evaluate the full Joint Model but also the performance of the Multicut formulation when not considering pairwise terms between trajectories (Tracklet MC + traj.) as well as the performance when omitting the pairwise terms between tracklet nodes (MCe + det.).
On the important f-measure and the number of segmented object, the proposed Joint Multicut model improves over the previous state-of-the-art in motion segmentation on the pedestrian sequences.

Quantitative results on the pedestrian tracking task are given in Tab. \ref{tab:tud-campus-result}. Again, we evaluate the importance of the model parts (denoted by MCe + det. and Tracklet MC + traj.). The comparison confirms that the full, joint model performs better than any of its parts.

Compared to the state of the art, the proposed method improves the recall on all three datasets. The general tendency is a decrease in the number of false negatives, while the number of false positives is higher than in \cite{tang15}.
\begin{table*}
\centering
\scalebox{0.80}{
\begin{tabular}{@{}l@{\hspace{0.05cm}}
c@{\hspace{0.12cm}} c@{\hspace{0.12cm}} c@{\hspace{0.12cm}} c@{\hspace{0.12cm}} c@{\hspace{0.12cm}} c@{\hspace{0.12cm}}
c@{\hspace{0.12cm}} c@{\hspace{0.12cm}} c@{\hspace{0.12cm}} c@{\hspace{0.12cm}} c@{\hspace{0.12cm}} c@{\hspace{0.12cm}}
c@{\hspace{0.12cm}} c@{\hspace{0.05cm}}
}
\!\!\!  & Rcll&  Prcsn & FAR & GT & MT & PT & ML & FP  & FN &IDs & FM& MOTA & MOTP &MOTAL\\
\hline
\hline
\!\!\! TUD-Campus \\
\hline
\!\!\! Frakiadaki et al.\cite{FragkiadakiECCV12}&50.4&57.5&1.89&8 &3 &2 &3 &134&178&  3&11&12.3&70.1&12.9\\
\!\!\! Milan et al. \cite{milan15}&32.6&82.4&0.35&8 &1 &3 &4 & 25&242&\bf   0&1 &25.6&72.9&25.6\\
\!\!\! Subgraph MC \cite{tang15} & 83.8  &\bf 99.3&\bf 0.03 &  8 & 5  &  2 & 1 & \bf 2 & 58  &\bf 0 & 1 &83.3  & 76.9 & 83.3 \\
\!\!\! Tracklet MC + traj. & \bf  90.3& 94.5&    0.27&    8&    6&    1&    1&   19&\bf  35&  \bf   0& \bf 0&   85.0&  \bf 77.1&   85.0\\
\!\!\! MCe \cite{keuper15a} + det.&82.5&93.7&0.28&8&5&2&1&20&63&3&2&76.0 &77.4 &76.7\\
\!\!\! JointMulticut &  87.5 & 98.4 &  0.07  & 8 &  5 &  2 & 1 & 5  &  45 & 1 & \bf 0 &\bf 85.8   & 77.0  &{\bf 86.0}\\
\hline
\hline
\!\!\! TUD-Crossing \\
\hline
\!\!\! Frakiadaki et al.\cite{FragkiadakiECCV12}&75.8&82.3&0.90&13&7 &5 &1 &180&267& 13&17&58.3&73.1&59.3\\
\!\!\! Milan et al. \cite{milan15}&  - &  - &0.2 &13&3 &7 &3 & 37&456&15&16&53.9&72.8& - \\
\!\!\! Subgraph MC \cite{tang15} & 82.0  &\bf 98.8&\bf  0.05 &  13 & 8  &  3 & 2 & \bf  11 & 198  &\bf 1 &\bf  1 & 80.9  &\bf  78.0 & 81.0\\
\!\!\! Tracklet MC + traj. & 85.4 &97.7& 0.11&   13& 9&  4&0  &22&161&5&11& 82.9&  76.9&   83.3\\
\!\!\! MCe \cite{keuper15a} + det.&\bf 92.5&83.3&1.01&13&12&1 &0  &204&\bf 83&14&5&72.7 &77.2 &73.8\\
\!\!\! JointMulticut &85.5  &  97.7 &0.11  & 13& 9  & 4   &0  & 22  & 160 & 2 & 9 &\bf 83.3  &77.3 &\bf 83.4 \\
\hline
\hline
\!\!\! ParkingLot \\
\hline
\!\!\! Subgraph MC \cite{tang15} & 96.1 &\bf 95.4&\bf 0.45 &  14 &13   & 1  &  0&\bf  113 & 95  &\bf  5& 18 &\bf 91.4  &\bf 77.4 &{\bf 91.5}\\
\!\!\! Tracklet MC + traj. & 96.6& 93.6&  0.66& 14&   13&    1&  0&  164&  85&9&\bf 13&89.5&  76.9&89.9 \\
\!\!\! MCe \cite{keuper15a} + det.&\bf 96.8& 88.6& 1.23&  14& 13&    1&  0 & 307&\bf  79&    6& 15&   84.1&   77.0&   84.3\\
\!\!\! JointMulticut & 96.6  & 94.9&  0.52&   14 & 13  &1  & 0  & 129 & 85 & 6 &  15  & 91.1 &77.2&91.3\\
\hline
\end{tabular}}
\caption{Tracking result on multi-target tracking sequences.}
\label{tab:tud-campus-result}
\vspace{-0.4cm}
\end{table*}
\begin{figure*}[t!]
\centering
\begin{tabular}{@{}l@{}l@{}l@{}l@{}l@{}}
\includegraphics[width=0.2\linewidth]{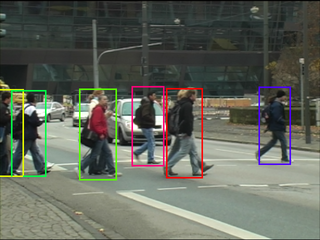}&
\includegraphics[width=0.2\linewidth]{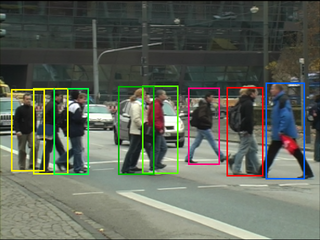}&
\includegraphics[width=0.2\linewidth]{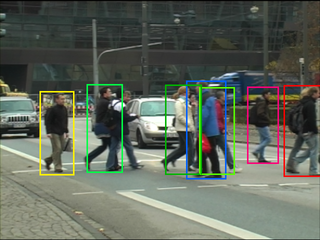}&
\includegraphics[width=0.2\linewidth]{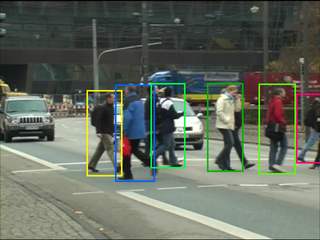}&
\includegraphics[width=0.2\linewidth]{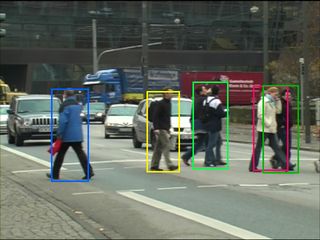}\\
\includegraphics[width=0.2\linewidth]{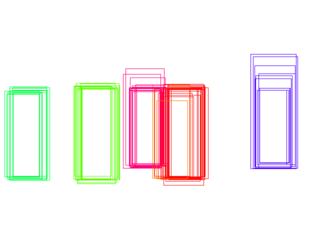}&
\includegraphics[width=0.2\linewidth]{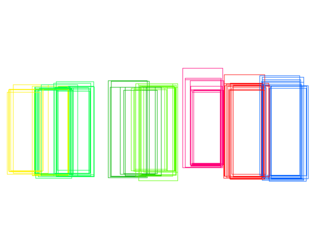}&
\includegraphics[width=0.2\linewidth]{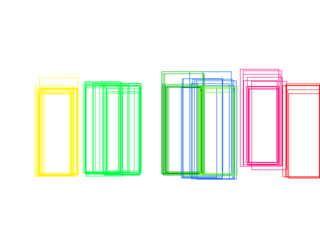}&
\includegraphics[width=0.2\linewidth]{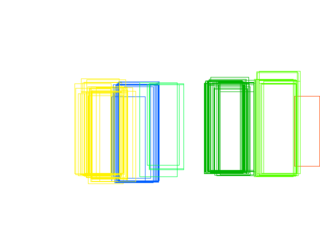}&
\includegraphics[width=0.2\linewidth]{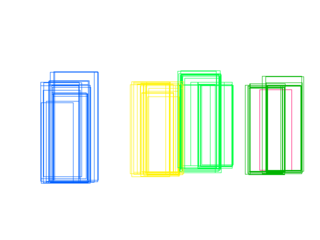}\\
\includegraphics[width=0.2\linewidth]{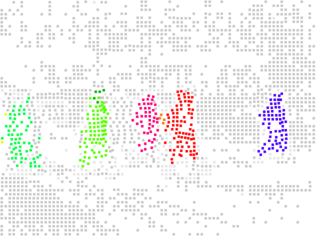}&
\includegraphics[width=0.2\linewidth]{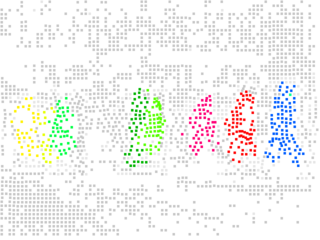}&
\includegraphics[width=0.2\linewidth]{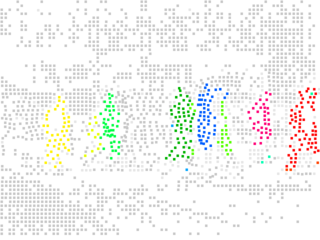}&
\includegraphics[width=0.2\linewidth]{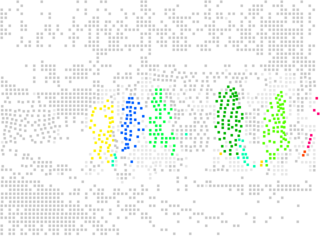}&
\includegraphics[width=0.2\linewidth]{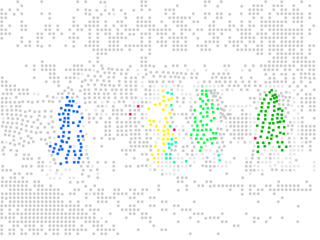}\\
frame 20&40&60&80&100
\end{tabular}
\caption{Results of the proposed Joint Multicut model on the TUD-crossing sequence.}
\label{fig:crossing}
\end{figure*}

A qualitative result is given in Fig. \ref{fig:crossing} for the TUD-crossing sequence. The bounding boxes overlayed on the image sequence are, for every frame and cluster, the ones with the highest detection score. These were also used for the tracking evaluation. The second row shows all clustered bounding boxes and the third row visualizes the trajectory segmentation. Both detection and trajectory clusters look very reasonable. Most persons can be tracked and segmented through partial and even complete occlusions. Segmentations provide better localizations for the tracked pedestrians.  

\subsection{2D MOT 2015} 
To allow for a comparison to other state-of-the-art multi-target tracking methods, we evaluate our joint multicut approach on the Multiple Object Tracking Benchmark 2D MOT 2015 \cite{MOT15}. In this benchmark, detections for all sequences are provided after Non-Maximum suppression. Thus, the provided detections are already too sparse for the pregrouping into tracklets that has been employed in \cite{tang15}. As a consequence, we use the detections directly as nodes as it is done for the motion segmentation. The computation of the pairwise terms between detections and between detections and trajectories need adapted as well.

\myparagraph{Implementation Details}
We compute the cut probabilities between detection nodes using Deep Matching \cite{weinzaepfelhal00873592}. Deep Matching is based on a deep, multi-layer convolutional architecture and performs dense image patch matching. It works particularly well when the displacement between two images is small. 

More concretely, each detection $d$ has the following properties: its spatio-temporal location $(t_d, x_d, y_d)$,
scale $h_d$, detection confidence $\textit{conf}_d$ and a set of matched keypoints $M_d$ inside the detecion $d$.
Given two detection bounding boxes $d_1$ and $d_2$, we define  $MU = |M_{d_{1}} \cup M_{d_{2}}|$, and $MI = |M_{d_{1}} \cap M_{d_{2}}|$
between the set $M_{d_{1}}$ and $M_{d_{2}}$.
Then the pairwise feature $f_e$ between the two detections no more that 3 frames appart is defined as  $(f_1, \textit{minConf}, f_1 \cdot \textit{minConf}, f_1^2, \textit{minConf}^2)$, where $f_1 = MI/MU$ and $\textit{minConf}$ is the minimum detection score between  $\textit{conf}_{d_1}$ and $\textit{conf}_{d_2}$.

The computation of pairwise terms between detections and trajectories is performed according to eq. \eqref{pedtrbbx} with an undirected template computed as the average of \ref{fig:pedGraph}~(a) and its horizontally flipped analogon. The sparseness of the detections also alters the statistics of the graph. Assuming that about 20 bounding boxes have been suppressed for every true detection, we weight the links between trajectory and detection nodes by factor 20. We are aware that this is a crude heuristic. Better options would be to learn this factor per sequence type or (better) to use the detections before Non-Maximum suppression which are unfortunately not provided. 

\myparagraph{Results} 
Our final results on the 2D MOT 2015 benchmark are given in table \ref{tab:MOT-result}. Compared to the state-of-the-art multi-target tracking method \cite{choi15}, we have an overall improvement in MOTA. Again, we observe a decrease in the number of false negatives while false positives increase.

\begin{table*}
\centering
\scalebox{0.80}{
\begin{tabular}{@{}l@{\hspace{0.05cm}}
c@{\hspace{0.12cm}} c@{\hspace{0.12cm}} c@{\hspace{0.12cm}} c@{\hspace{0.12cm}} c@{\hspace{0.12cm}} c@{\hspace{0.12cm}}
c@{\hspace{0.12cm}} c@{\hspace{0.12cm}} c@{\hspace{0.12cm}} c@{\hspace{0.12cm}} c@{\hspace{0.12cm}} c@{\hspace{0.12cm}}
c@{\hspace{0.12cm}} c@{\hspace{0.05cm}}
}
\!\!\!               & & & FAR & & MT &  & ML & FP & FN &IDs & FM& MOTA & MOTP &\\
\hline
\!\!\! Choi \cite{choi15}&&         &{\bf 1.4}&&12.2\%&        &44\%  &{\bf 7,762}&32,547&{\bf 442}&823&33.7&{\bf 71.9}&\\
\!\!\! Milan et al. \cite{milan15}&&&{\bf 1.4}&&5.8\% &        &63.9\%&7,890&39,020&697&{\bf 737}&22.5&71.7&\\
\!\!\! JointMulticut & & & 1.8 & &{\bf 23.2\%}       &  &{\bf 39.3\%}&10,580&{\bf 28,508}&457&969&{\bf 35.6}&{\bf 71.9}&\\
\hline
\end{tabular}}
\caption{Tracking results on the 2D MOT 2015 benchmark.}
\label{tab:MOT-result}
\vspace{-0.4cm}
\end{table*}

\section{Conclusion}
This paper proposes a Multicut Model that jointly addresses multi-target tracking and motion segmentation so as to leverage the advantages of both. Motion segmentation allows for precise local motion cues and correspondences that support robust multi-target tracking results with high recall. Object detection and tracking allows a more reliable grouping of
motion trajectories on the same physical object. Promising experimental results are obtained in both domains with a strong improvement over the state of the art in motion segmentation.

\section*{Acknowledgments}
M.K. and T.B. acknowledge funding by the ERC Starting Grant VideoLearn.
\bibliographystyle{splncs}
\bibliography{jointSegTrack.bib}

\end{document}